\PassOptionsToPackage{nopatch=footnote}{microtype}
\documentclass[10pt, a4paper]{article}
\usepackage{todonotes}
\usepackage[final]{lrec2026}
\usepackage{subcaption}
\usepackage{placeins}
\usepackage{xspace}
\newcommand{\npn}{\textsc{npn}\xspace}
\newcommand{\prep}{\texttt{PREP}\xspace}
\newcommand{\unk}{\texttt{[UNK]}\xspace}
\usepackage{tabularx}

\usepackage[most]{tcolorbox}
\usepackage{langsci-gb4e}
\usepackage{array}

\title{`Layer su Layer': Identifying and Disambiguating the Italian \npn Construction in BERT's family}

\name{Greta Gorzoni, Ludovica Pannitto, Francesca Masini} 

\address{Alma Mater Studiorum - University of Bologna\\
         greta.gorzoni@studio.unibo.it, ludovica.pannitto@unibo.it, francesca.masini@unibo.it\\}

\abstract{
Interpretability research has highlighted the importance of evaluating Pretrained Language Models (PLMs) and in particular contextual embeddings against explicit linguistic theories to determine what linguistic information they encode. This study focuses on the Italian \npn (noun$_i$-preposition-noun$_i$) constructional family, challenging some of the theoretical and methodological assumptions underlying previous experimental designs and extending this type of research to a lesser-investigated language.
Contextual vector representations are extracted from BERT and used as input to layer-wise probing classifiers, systematically evaluating information encoded across the model’s internal layers. The results shed light on the extent to which constructional form and meaning are reflected in contextual embeddings, contributing empirical evidence to the dialogue between constructionist theory and neural language modelling.
 \\ \newline  
 }

\begin{document}

\maketitleabstract

\section{Introduction}
\label{sec:intro}

The remarkable empirical performance obtained by Pretrained Language Models (PLMs) across a wide range of tasks has fueled enthusiasm in both computational approaches and theoretical debates about language \citep{brown2020languagemodelsfewshotlearners}. Despite these successes, PLMs remain largely opaque \citep{rogers-etal-2020-primer}. High predictive accuracy does not automatically entail theoretical understanding: it remains unclear what kinds of linguistic information these models encode, how such information is internally structured, and to what extent their representations align with linguistically motivated categories.

Assessing a model’s linguistic competence therefore requires an explicit theoretical characterization of the phenomenon under investigation.
Construction Grammar (CxG, \citealt{fillmore1988mechanisms,goldberg_constructions_1995,goldberg_jackendoff_1996}) offers a particularly suitable framework in this respect: within CxG, language is conceived as a structured network of conventionalized form–meaning pairings, namely Constructions (Cxns), which may vary in both complexity and schematicity. Since phenomena at the syntax–semantics interface remain comparatively underexplored in interpretability research \cite{GraichenBoledaDeDiosFlores2026}, a constructionist perspective provides a principled way to investigate whether contextual embeddings encode structured form–meaning associations rather than surface-level distributional regularities \cite{pannitto-herbelot-2023-calamo,weissweiler_better_2022, Rambelli_2025}.

This study focuses on the Italian \npn (noun$_i$-preposition-noun$_i$) constructional family (e.g., \textit{layer su layer} `layer upon layer', \citealt{masini2024npn}) and builds on previous research on the English \npn Cxn \cite{scivetti_Construction_2025}, which showed that BERT encodes information relevant to both Cxn identification and semantic disambiguation. By extending the investigation to Italian, where (as in English) \npn patterns form a network of horizontally related Cxns exhibiting semantic specialization and competition, this work tests whether constructional knowledge is testable across these different languages and across the \npn family of sister Cxns.
Our work also addresses a theoretical need within CxG. While constructionist approaches provide a rich descriptive and explanatory framework, they increasingly call for empirical validation. Neural language models, whose representations emerge from large-scale distributional exposure, offer a unique opportunity to test usage-based constructionist hypotheses in a controlled and quantifiable way. By probing contextual embeddings for evidence of constructional organization, this work contributes to the dialogue between linguistic theory and neural language modelling, asking not just whether models perform well, but whether they encode linguistically meaningful structure.

The paper is structured as follows. Section \ref{sec:italiannpn} introduces the \npn construction, while Section \ref{sec:related} situates the study within the relevant literature. Section \ref{sec:Research Questions} formulates the research questions and details the methodological design. Section \ref{sec:methods} introduces the dataset and experimental setup. Sections \ref{sec:identification} and \ref{sec:disambiguation} present and discuss the results of the identification and semantic disambiguation experiments, respectively.\footnote{All code developed for this project is available on GitHub: \url{https://github.com/GretaGorzoni00/NPN}}

\section{The \npn Construction}
\label{sec:italiannpn}

\npn expressions challenge traditional grammatical categories and motivate a model capable of capturing phenomena along the lexicon–syntax continuum. Formally, the pattern consists of nominal reduplication interrupted by a preposition.

  \begin{tcolorbox}[colback=white, colframe=black, arc=4mm, boxrule=0.8pt, center title, title=Construction Schema]
    \[
      Noun_i \; \text{Preposition} \; Noun_i
    \]
  \end{tcolorbox}

Treating \npn expressions as semi-specified Cxns accounts for both their productivity and their idiosyncratic properties: nominal identity, determiner absence, restricted prepositional choice, as well as characteristic meanings such as succession, distributivity, or contact, which are encoded within the constructional schema itself rather than derived from syntactic operations.

Despite his complex positioning in the constructionist framework \cite{goldberg_jackendoff_1996}, in his influential analysis of the English \npn pattern, \citet{jackendoff_Construction_2008} identifies a limited set of productive prepositions (\emph{by, for, to, after, upon}) and associates them with a restricted inventory of core meanings (e.g., \textit{succession}, \textit{juxtaposition}, \textit{comparison}), alongside numerous idiomatic instantiations. Within this account, \npn is treated as a unified abstract constructional schema pairing a formal template with a family of related meanings.
This view has been challenged by \citet{sommerer_absent_2021}, who argue that the \npn family does not instantiate a single highly abstract Cxn. Instead, they propose a network of horizontally linked, semi-schematic Cxns, emphasizing usage patterns and semantic specialization over maximal abstraction.

The Italian \npn family has been investigated by \citet{masini2024npn}, who adopts a similar horizontal-network perspective. Based on corpus evidence from CORIS\footnote{\textit{Corpus di Riferimento per l’Italiano Scritto}; \citealt{favretti_coriscodis_nodate}.}, Masini identifies 8 semi-schematic Cxns, summarized in Table~\ref{tab:npn_constructions}. Her dataset amounts to 1,298 types with varying token frequencies \citep{masini2024amsacta}. This study focuses only on Cxns with the prepositions \textit{a} `at/to' and \textit{su} `on'.

\begin{table}
\centering
\scriptsize
\begin{tabularx}{\columnwidth}{m{0.009\columnwidth}m{0.38\columnwidth}m{0.18\columnwidth}m{0.06\columnwidth}m{0.09\columnwidth}}
\hline
& \textbf{Form} & \textbf{Meaning} & \textbf{Types} & \textbf{Tokens} \\
\hline
\textbf{1} & $\,[[x]_{N_i}\ a\ [x]_{N_i}]\,_{\mathrm{MOD}_j}$ &
Succession, iteration, distributivity & 21  & 399 \\
\hline
\textbf{2} & $\,[[x]_{N_i}\ a\ [x]_{N_i}]\,_{\mathrm{MOD}_j}$ &
Juxtaposition, contact  & 27  & 1108 \\
\hline
\textbf{3} & $\,[[x]_{N_i^{sg}}\ su\ [x]_{N_i^{sg}}]\,_{\mathrm{MOD}_j}$ &
Succession, iteration, distributivity & 71  & 178 \\
\hline
\textbf{4} & $\,[[x]_{N_i^{pl}}\ su\ [x]_{N_i^{pl}}]\,_{\mathrm{MOD}_j}$ &
Greater plurality, accumulation & 252  & 403 \\
\hline
\textbf{5} & $\,[[x]_{N_i}\ per\ [x]_{N_i}]\,_{\mathrm{MOD}_j}$ &
Succession, iteration, distributivity & 385  & 3324 \\
\hline
\textbf{6} & $\,[[x]_{N_i}\ per\ [x]_{N_i}]\,_{\mathrm{SUB.CL}_j}$ &
Inescapable presupposition & 21  & 21 \\
\hline
\textbf{7} & $\,[[x]_{N_i}\ dopo\ [x]_{N_i}]\,_{\mathrm{MOD}_j}$ &
Succession, iteration, distributivity & 368  & 2387 \\
\hline
\textbf{8} & $\,[[x]_{N_i}\ contro\ [x]_{N_i}]\,_{\mathrm{MOD}_j}$ &
Juxtaposition, contact & 56  & 170\\
\hline

\end{tabularx}
\caption{The 8 Italian \npn Cxns postulated by \cite{masini2024npn}. This work deals with Cxns 1--4.}
\label{tab:npn_constructions}
\end{table}

The Italian \npn family exhibits substantial overlap in the noun lexemes licensed by different patterns \citep{masini2024npn}. Such overlap and semantic competition make the Italian \npn family a particularly informative test case for contextual embeddings, as the coexistence of partially overlapping Cxns with subtly differentiated semantic profiles raises the question of how such distinctions are encoded in distributional representations and whether they are accessible to PLMs.

\section{Related work}
\label{sec:related}

Recent work has investigated whether LLMs encode constructional knowledge using a variety of experimental designs. One line of research \citep{madabushi2020cxgbertbertmeetsconstruction,madabushi2025construction} examines multiple Cxns organized along a gradient of schematicity, testing whether models generalize across instantiations and whether increasingly schematic patterns remain accessible in contextual embeddings.
A complementary approach focuses on individual Cxns and operationalizes constructional knowledge through controlled discrimination tasks, often contrasting target Cxns with carefully designed distractors \citep{scivetti_Construction_2025, weissweiler_better_2022}.
Recently, attention has turned to rare but productive Cxns \citep{weissweiler2025hybrid}, which provide a stringent test for constructional generalization: because they are infrequent, successful modelling cannot rely solely on collostructional frequency effects but must capture construction-specific contributions.

Results converge on a graded pattern: lower-level, lexically anchored Cxns tend to be more accessible to LLMs, while schematic patterns pose greater challenges \citep{bonial2024construction}. 
At the same time, recent work has questioned core assumptions underlying constructional interpretability. \citet{jumelet2024blackbigboxeslanguage} argue that model generalizations should not only be measured in terms of performance but also compared to human generalization patterns. \citet{dunn2025llms} further caution against confirmation bias in Cxn probing, emphasizing the importance of detecting false positives and avoiding uncritical expansion of the Constructicon (viz., the network of Cxns) based on model behaviour alone.

This work directly builds upon \citet{scivetti_Construction_2025}, who investigate the English \npn Constructional pattern using a probing framework. In their study, contextual embeddings extracted from BERT are used as input to a linear classifier to assess whether constructional information is encoded in the model’s representations. Their dataset includes instances of the \npn Cxn (e.g., `I was living \emph{moment to moment}') contrasted with superficially similar distractors (e.g., `In Rome largesse was doled out by \textit{individuals} \emph{to individuals}'), with lemma-level disjoint train–test splits to prevent lexical memorization. Two experiments target Cxn identification and one addresses semantic disambiguation. Results show that BERT-based probes reliably distinguish Cxns from distractors and capture semantic distinctions, with peak performance in middle-to-late layers, outperforming static embedding baselines.

\section{Research Questions and Methodological Design}
\label{sec:Research Questions}

As Cxns are assumed to be inherently language-specific, probing constructional knowledge requires moving beyond the English-centric focus that characterises much of the existing literature. Moreover, the \npn Cxn occupies an intermediate position on the lexicon–syntax continuum, making it a suitable test case for assessing whether models can capture constructional generalizations that are neither fully schematic nor fully lexically specified.

This study addresses two research questions:
\begin{description}
    \item[RQ1] Can BERT distinguish instances of the Italian \npn Cxn from distractors?
    \item[RQ2] Can BERT distinguish between the construction-specific meanings associated with different \npn instantiations within the Italian \npn family?
\end{description}

Following \citet{scivetti_Construction_2025}, we adopt a probing framework in which contextual embeddings extracted from BERT family's models are used as input to a linear classifier. The preposition is operationally treated as the structural head of the Cxn \cite{jackendoff_Construction_2008}, and its embedding is selected as a primary locus for probing.

To address the research questions, two complementary tasks are implemented: (i) Cxn identification, testing whether the model distinguishes actual \npn instances from formally similar sequences (see Section~\ref{sec:identification}), and (ii) semantic disambiguation, 
assessing whether the information encoded in the contextual embedings are adequate
to distinguish between the construction-specific meanings associated with different \npn instantiations.

Our study extends \citet{scivetti_Construction_2025} in several important ways. First, the dataset includes different horizontally related Italian \npn semi-schematic Cxns featuring prepositions \textit{a} `at/to' and \textit{su} `on' (namely Cxns 1, 2, 3 and 4 in Table~\ref{tab:npn_constructions}): this design tests whether embeddings capture constructional generalizations within a network of sister Cxns, with localized competition~\cite{masini2024npn}.

Second, the distractors set is broadened, including distinct Cxns (e.g., \textsc{pnpn}) identified in the literature (see Example~\ref{ex:agenzia}: similar cases are excluded from \citet{scivetti_Construction_2025}'s setup) beyond surface-isomorphic patterns: this refines the identification task moving from syntactic discrimination to identification of true form–meaning pairings.

\ea
\label{ex:agenzia}
[...] \textit {con una successione da \textbf{agenzia ad agenzia} quasi automatica. \\}
`[...] with an almost automatic succession from \textbf{agency to agency}'.
\z

Third, both the embedding of the \texttt{[UNK]} token (used as prepositional substitute) and the embedding of the preposition itself (henceforth, \texttt{PREP}) are probed, allowing us to compare abstracted and lexically grounded representations, and to assess how constructional information interacts with prepositional semantics.

\section{Methods}
\label{sec:methods}

\subsection{Data}
\label{sec:data}

The dataset used in this study \citep{Gorzoni_NPN_dataset_2025} is derived from the Italian \npn dataset presented in \citet{masini2024npn}, extended with full sentential contexts extracted from CORIS\footnote{The manually annotated dataset of Italian \npn Constructions and Distractors \citep{Gorzoni_NPN_dataset_2025}, including semantic annotations and inter-annotator agreement data, is archived on Zenodo: \url{https://zenodo.org/records/18268135}}.

The full dataset contains 3,256 attested instances of the Italian \npn constructional pattern instantiated by the prepositions \textit{a} `at/to' and \textit{su} `on'. Following the annotation schema proposed in \cite{masini2024npn}, each occurrence is manually annotated with one of five semantic labels: \textit{succession/iteration/distributivity}, \textit{greater\_plurality/accumulation}, \textit{juxtaposition/contact}, \textit{connection/transition}, and \textit{idiosyncratic}. The dataset further comprises 1,751 distractor instances spanning eight pattern types, for a detailed description see Appendix~\ref{sec:distractors}.
All data were manually cleaned to remove ill-formed sentences. Since the present study focuses exclusively on Cxns (1), (2), (3) and (4) in Table~\ref{sec:italiannpn}, only Cxn instances annotated with the \textit{succession/iteration/distributivity}, \textit{greater\_plurality/accumulation} and \textit{juxtaposition/contact} label were included in the operative dataset for the analysis. The resulting dataset consists of 1,281 constructional instances and 989 distractors.\footnote{Numerous distractor instances belonging to the same distractor type in the full dataset were excluded in order to ensure heterogeneity.}
All instances in the dataset were annotated for their \textit{meaning}, the \textit{lemma} instantiating the Cxn and its \textit{number}. A subset of 100 Cxns (equally balanced across Cxns (1), (2), (3) and (4) in Table \ref{tab:npn_constructions}) were cross-annotated by a group of 5 annotators. Annotation quality of Cxns' meaning was assessed using Cohen’s $\kappa$ and Krippendorff’s $\alpha$ \cite{artstein_inter-coder_2008}. Pairwise Cohen’s $\kappa$ shows strong to near-perfect reliability across annotator pairs ($\kappa = 0.79-0.91$).
Nominal $\alpha$ is high ($\alpha = 0.858$) and further increases when a reduced penalty is assigned to confusions between semantically adjacent labels (\textit{succession/iteration/distributivity} and \textit{greater\_plurality/accumulation}, $\alpha = 0.892$).
The entire dataset was filtered to only retain longer sentences (> 5 tokens) and at most 30 items with the same lemma per preposition. 

For each experiment, data is then split into 5 training and test partitions using an 80/20 ratio. Unlike \citet{scivetti_Construction_2025}, we do not enforce full lemma-level disjointness across splits. 
Instead, we adopt a modified strategy: lemma–label pairs are never shared between training and test sets, while allowing the same lemma to occur across labels. 
This design preserves lexical separation at the level of individual labels while retaining cross-label lexical overlap. In particular, it enables evaluation on cases where a lemma appears in training only with the opposite label, providing a stricter test of whether constructional distinctions are recoverable from contextual embeddings beyond lexical identity.
Class balancing is applied to both training and test data.
Tables~\ref{tab:simple}--\ref{tab:ex2split} summarize the composition of the data splits used for both experiments (i) and (ii).

\subsection{Models and probing}
\label{sec:models}

Probing methods aim to investigate which types of information are encoded in a model's internal representations by training an auxiliary classifier, commonly referred to as a probing classifier. The representations are first extracted from a pre-trained model and then used as input for the probing classifier, together with labels corresponding to a linguistic property that has been explicitly operationalised. The central assumption underlying probing is that the probing classifier’s performance reflects the extent to which the target property is accessible in the representations. Crucially, to draw meaningful conclusions about the presence of linguistic information in the embeddings, the probing setup must rely on a deliberately weak classifier paired with a sufficiently complex task~\citep{hewitt_designing_2019}. This constraint helps ensure that high performance is attributable to information encoded in the representations themselves, rather than to the probe's capacity to learn the task independently.

To evaluate whether constructional information is linearly accessible in contextual embeddings, we train a separate logistic regression probing classifier on embeddings extracted from each layer of the four considered BERT family's models, and track their performance across layers. 
\citet{scivetti_Construction_2025}, we adopt a BERT-base architecture\footnote{\url{dbmdz/bert-base-italian-cased}} (\citealt{bayerische_staatsbibliothek_2025}, 12 layers, 768 hidden units, 12 attention heads) trained on Italian data. We also include multilingual BERT (mBERT\footnote{\url{google-bert/bert-base-multilingual-cased}}, \citealt{{devlin-etal-2019-bert}}), that enables cross-linguistic comparison and allows partial replication of \citet{scivetti_Construction_2025} under a unified architecture.
To further evaluate the role of monolingual specialization, we test UmBERTo\footnote{\url{Musixmatch/umberto-commoncrawl-cased-v1}} \citep{musixmatch-2020-umberto}, a RoBERTa-based model trained exclusively on Italian corpora: this allows to evaluate the effect of pretraining objectives. Finally, we include XLM-RoBERTa\footnote{\url{FacebookAI/xlm-roberta-base}} \citep{DBLP:journals/corr/abs-1911-02116}, a multilingual model trained on large-scale cross-lingual corpora, as a multilingual companion to UmBERTo. All models are used in inference mode without fine-tuning. Contextual embeddings are extracted layer-wise and evaluated using the same probing protocol across architectures. We consider representations corresponding to the prepositional token and its \unk substitute.

We also train a control classifier \cite{hewitt_designing_2019} with a random label assigned to each lemma, whose performance should be near chance given the absence of spurious correlations between train and test data.
A probing linear classifier is also trained on static embeddings as baseline: differently from \citealt{scivetti_Construction_2025}, we adopt FastText \citep{bojanowski-etal-2017-enriching,joulin-etal-2017-bag} rather than GloVe as a static baseline because its subword-based representations are better suited to morphologically rich languages such as Italian, allowing us to control for lexical and inflectional variation.

\subsection{Experimental setup}
\label{sec:expsetup}

For the identification task, we perform binary classification (\textit{Construction} vs. \textit{Distractor}). For the disambiguation task, we perform multi-class classification across the three semantic labels. 
In order to ensure maximum comparability with the previous study on English \npn constructions, we extended our experimental setup to the dataset introduced by \cite{scivetti_Construction_2025}. Specifically, their data were converted into our format, subjected to our splitting procedure, and processed through the same embedding extraction and probing classification pipeline adopted for the Italian experiments. In addition, we computed a static baseline using English FastText vectors on the English configuration as well.

\section{Identification task}
\label{sec:identification}

The first experiment evaluates whether contextual embeddings extracted from BERT's models encode sufficient information to distinguish \npn constructions from distractors, and analyzes how the nature of the distractor patterns affect the probing classifier’s behaviour.

In \citet{scivetti_Construction_2025}'s implementation, in fact, the identification task contrasts actual \npn instances with surface-isomorphic patterns. Because the distractor class is syntactically homogeneous and derives from a different structural configuration (e.g., the preposition does not function as the head in the distractor cases), the task may largely rely on syntactic structural cues, which could make it less clear to what extent specific constructional information is being actually tested.

Given the heterogeneous composition of our \textit{distractor} portion of the dataset, three different sampling procedures were implemented\footnote{All the the train test split and in particular the constraint needed for each configuration are menaged through the Python package \cite{Pannitto_sample-dataset_2025}}:
\begin{itemize}
    \item[(i)] a fully balanced configuration (henceforth, \texttt{SIMPLE}), where both training and test sets contain all distractor types;
    \item[(ii)] a configuration where training is performed exclusively on structurally distinct constructions (i.e., \textsc{pnpn}, \textsc{verbal} and \textsc{nsungiù}) and testing on the full balanced test set (henceforth, \texttt{OTHER} configuration);
    \item[(iii)] a configuration where, conversely, training is performed exclusively on surface-isomorphic patterns (i.e., \textit{Thematic target}, \textit{N-extended}, \textit{NUM P NUM} and \textit{Proper name inglobation} patterns), and evaluation on the full balanced test set (henceforth, \texttt{PSEUDO}). 
\end{itemize}

The aim of these configurations is to investigate whether providing (\texttt{OTHER}), as opposed to withholding (\texttt{PSEUDO}), structurally informative negative examples that make constructional boundaries explicit has an impact on the performance of the probing classifier.

\begin{figure}[htbp]
    \centering
    
    \begin{subfigure}{0.48\columnwidth}
        \centering
        \includegraphics[width=\linewidth]{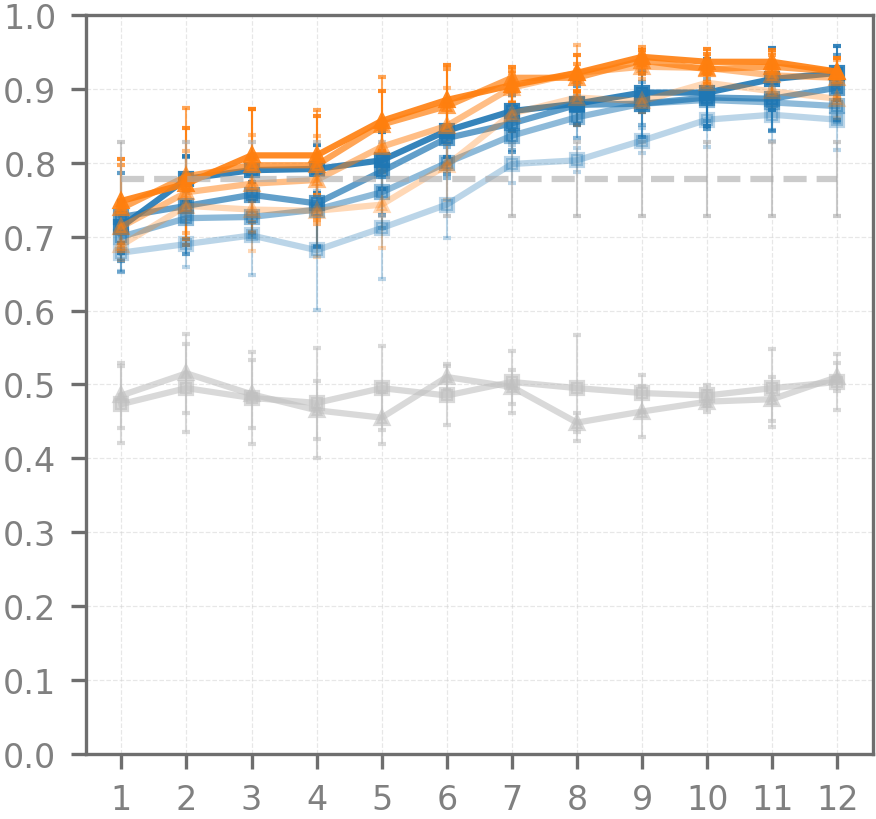}
        \caption{bert-cased-ita}
        \label{fig:a1}
    \end{subfigure}
    \hfill
    \begin{subfigure}{0.48\columnwidth}
        \centering
        \includegraphics[width=\linewidth]{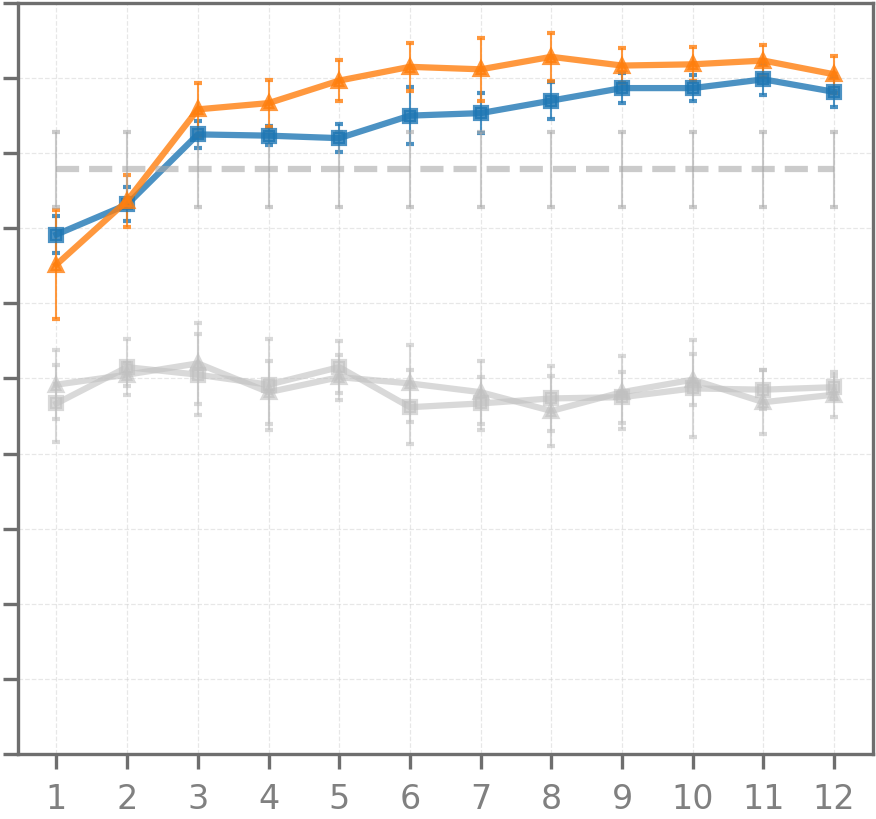}
        \caption{mBERT}
        \label{fig:a2}
    \end{subfigure}
    
    \vspace{0.5em}
    
    \begin{subfigure}{0.48\columnwidth}
        \centering
        \includegraphics[width=\linewidth]{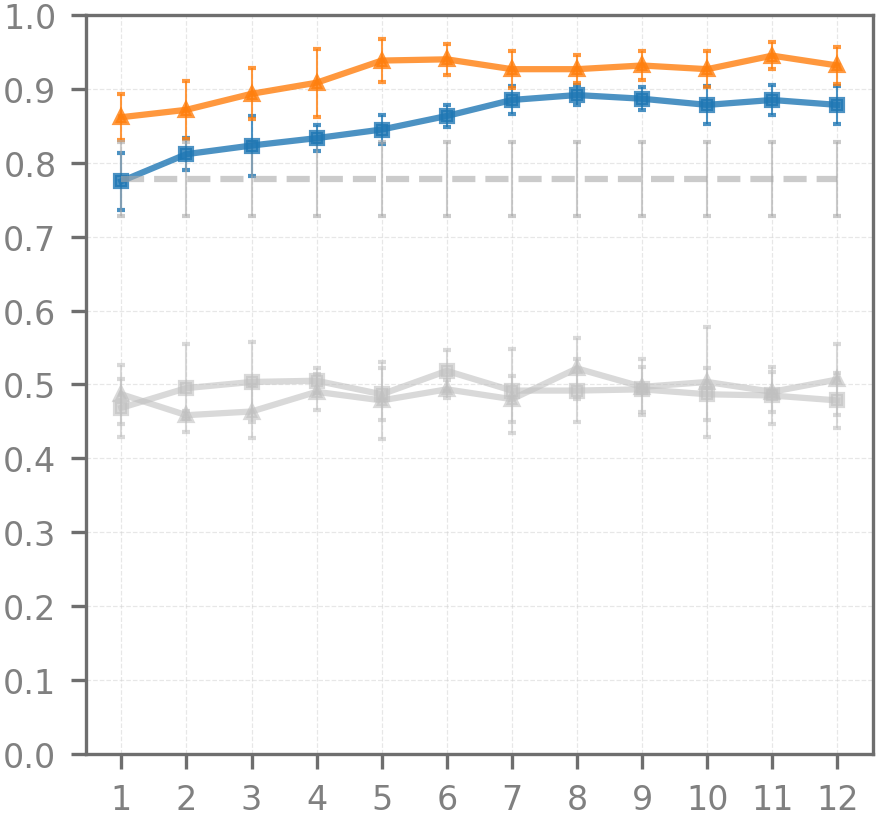}
        \caption{UmBERTo}
        \label{fig:a3}
    \end{subfigure}
    \hfill
    \begin{subfigure}{0.48\columnwidth}
        \centering
        \includegraphics[width=\linewidth]{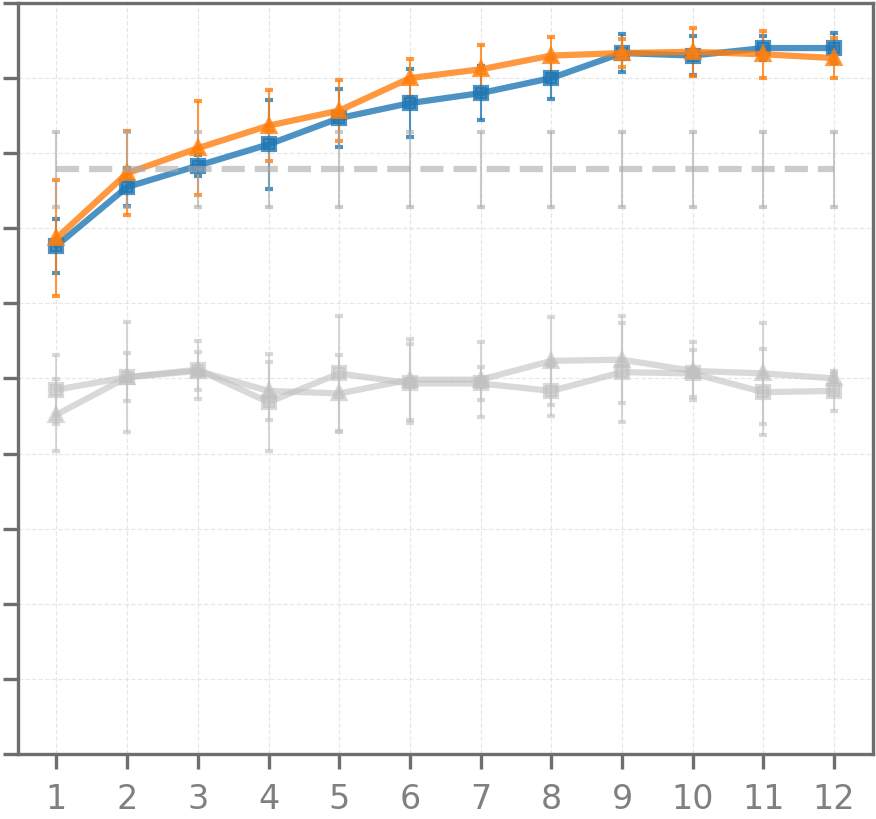}
        \caption{XLM-RoBERTa}
        \label{fig:a4}
    \end{subfigure}
    
    \caption{Accuracy of \unk (red lines, square dots) and \prep (orange lines, triangular dots) on Construction identification for the \texttt{SIMPLE} configuration. As in the following plots, the accuracy of the five probing classifiers resulting from the five random splits is averaged. Dashed grey line represents FastText baseline. Continuous grey lines refer to control classifiers. Figure (\ref{fig:a1}) includes decremental training configurations, line shading becomes progressively lighter as the number of training instances decreases (480 → 240 → 120 → 60). No substantial performance differences emerge across configurations.
    Figure~\ref{fig:pcaex1UNK} in Appendix provides a qualitative visualisation of the embedding space across layers.
    }
    \label{fig:ex1}
\end{figure}

With regard to configuration (i), as shown in Figure~\ref{fig:ex1}, probing classifiers trained on contextual embeddings, regardless of the specific model considered, achieve strong performance in distinguishing \npn Cxns from distractors, both for \unk and \prep.

As in \citet{scivetti_Construction_2025}, the control classifier remains close to chance level, consistent with the selectivity criterion discussed by \citet{hewitt_designing_2019}. 
The static lexical baseline performs well above chance: FastText probe provides a meaningful lower bound, which is outperformed by contextual embeddings only in middle to late layers, consistently with the interpretation that deeper layers encode more abstract information that is not available to static embeddings alone \citep{hewitt-manning-2019-structural, tenney-etal-2019-bert, scivetti_Construction_2025}.

\begin{figure}[htbp]
    \centering
    
    \begin{subfigure}{0.48\columnwidth}
        \centering
        \includegraphics[width=\linewidth]{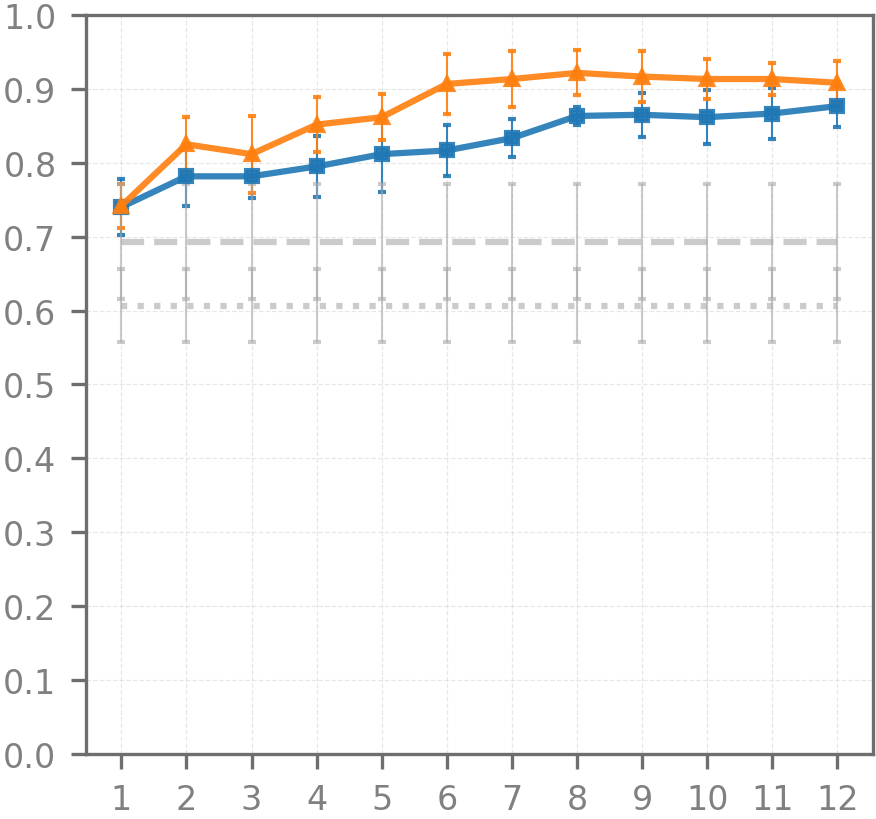}
        \caption{Exp. 1 - bert-cased-eng}
        \label{fig:c1}
    \end{subfigure}
    \hfill
    \begin{subfigure}{0.48\columnwidth}
        \centering
        \includegraphics[width=\linewidth]{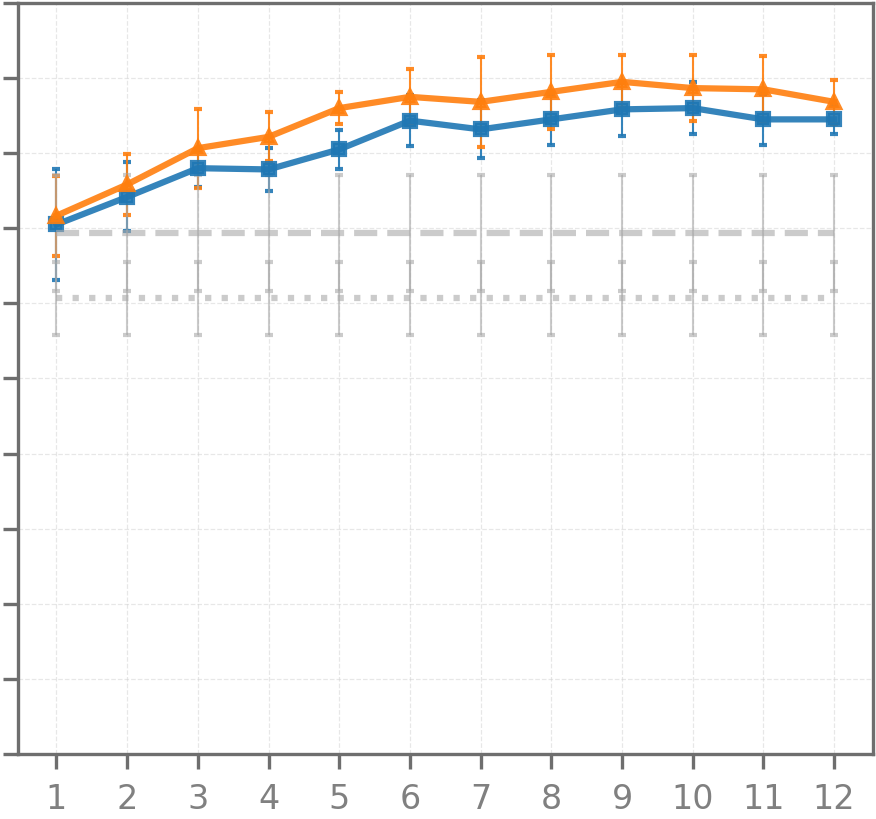}
        \caption{Exp. 1 - mBERT}
        \label{fig:c2}
    \end{subfigure}
    
    \vspace{0.5em}
    
    \begin{subfigure}{0.48\columnwidth}
        \centering
        \includegraphics[width=\linewidth]{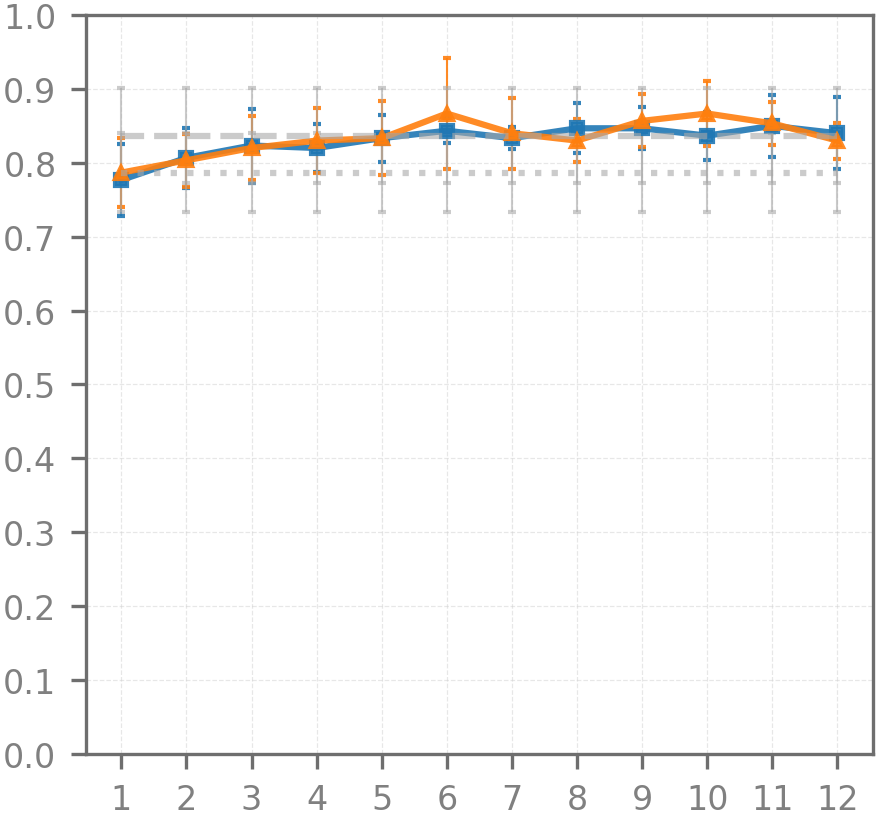}
        \caption{Exp. 2 - bert-cased-eng}
        \label{fig:c3}
    \end{subfigure}
    \hfill
    \begin{subfigure}{0.48\columnwidth}
        \centering
        \includegraphics[width=\linewidth]{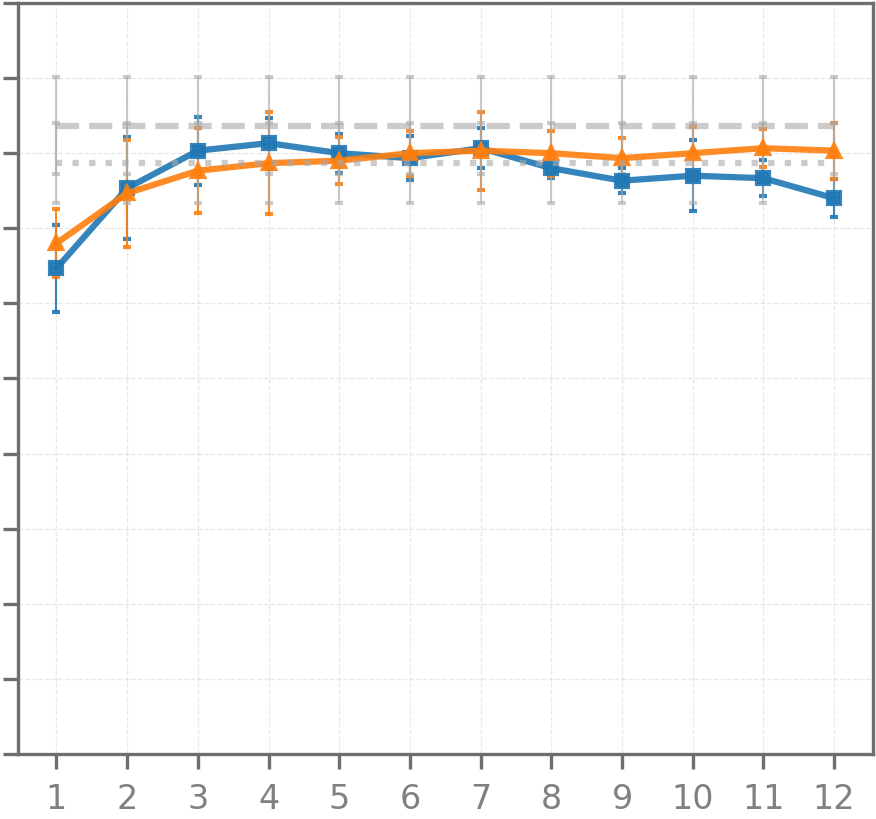}
        \caption{Exp. 2 - mBERT}
        \label{fig:c4}
    \end{subfigure}
    
    \caption{Accuracy of \unk (red lines, square dots) and \prep (orange lines, triangular dots) on both experiments on English data from \citet{scivetti_Construction_2025}. Dashed grey line represents FastText baseline. Dotted grey line represents GloVe baseline.}
    \label{fig:ex1-scivetti}
\end{figure}

Results are confirmed on English \npn pattern as well: global performance is consistent both for \unk and \prep embeddings, but GloVe and FastText baselines perform quite differently on the task (Figure~\ref{fig:ex1-scivetti}).

Finally, the fact that \unk and \prep yield similar performance is informative. Replacing the preposition with \unk removes direct access to lexical information, yet the probe remains highly accurate. This suggests that successful discrimination does not crucially depend on access to the preposition’s lexical semantics, but can be achieved on the basis of information contributed by the surrounding nouns and the broader sentential context. At the same time, the comparable performance of \prep suggests that lexical information about the preposition does not substantially increase accuracy beyond what is already available in the contextual configuration. These results support the conclusion that the probes capture information in BERT representations that reliably distinguishes the \npn Cxn from near-minimal distractor counterparts beyond what is possible through lexical semantic cues alone.

\begin{figure}[htbp]
    \centering
    
    \begin{subfigure}{0.48\columnwidth}
        \centering
        \includegraphics[width=\linewidth]{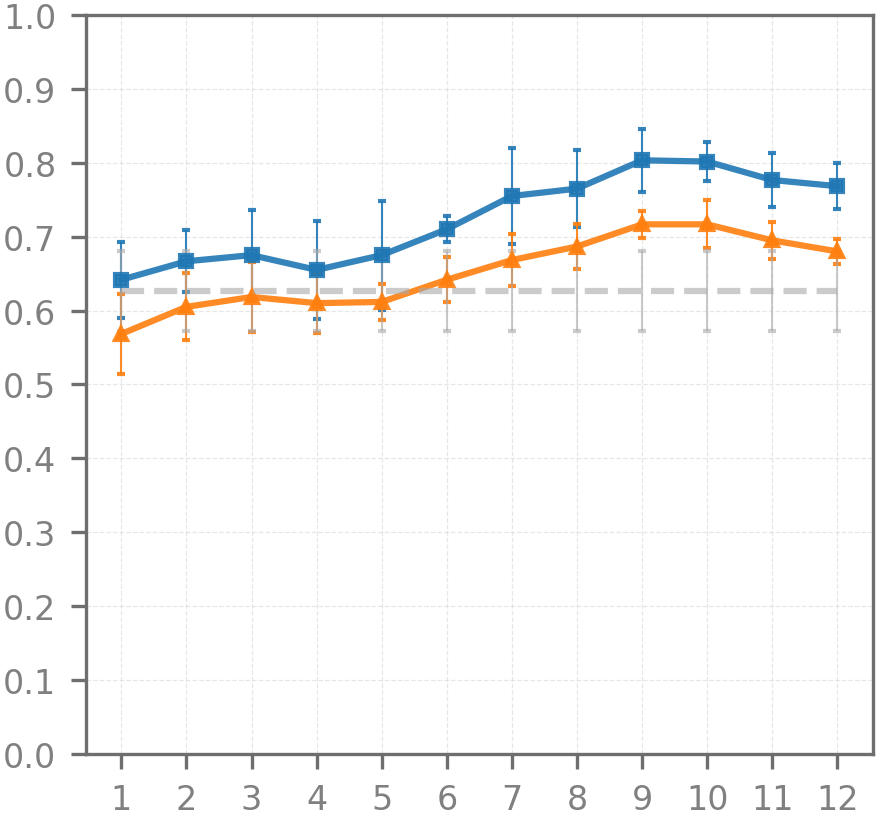}
        \caption{\texttt{OTHER}}
        \label{fig:b1}
    \end{subfigure}
    \hfill
    \begin{subfigure}{0.48\columnwidth}
        \centering
        \includegraphics[width=\linewidth]{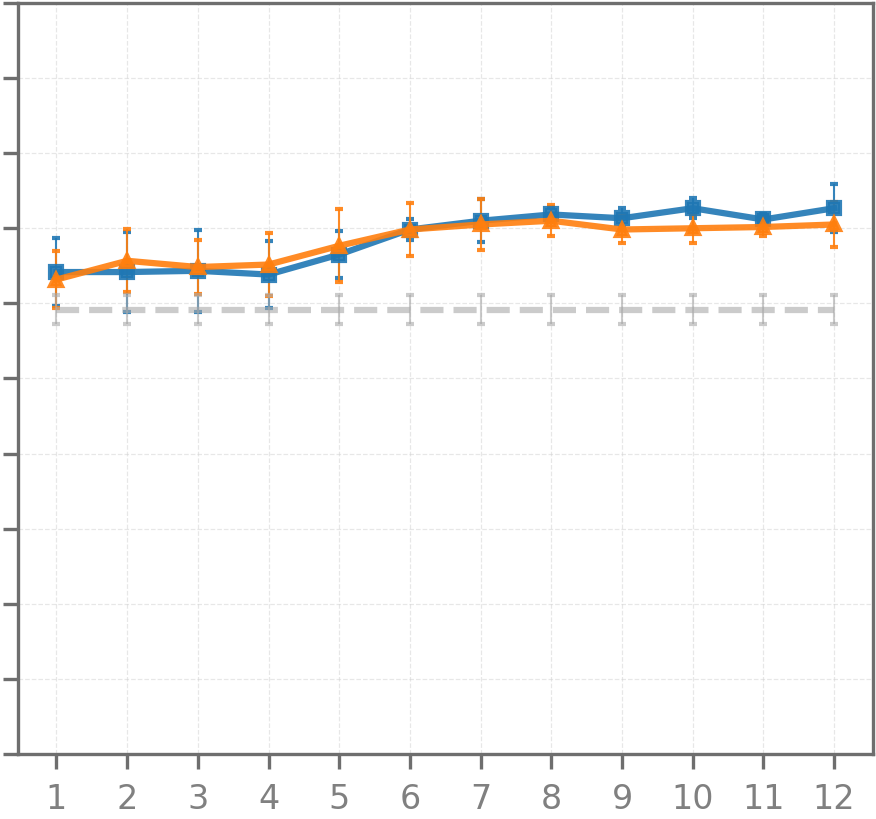}
        \caption{\texttt{PSEUDO}}
        \label{fig:b2}
    \end{subfigure}
\caption{Accuracy of \unk (red lines, square dots) and \prep (orange lines, triangular dots) on Construction identification for the \texttt{OTHER} and \texttt{PSEUDO} configurations. Dashed grey line represents FastText baseline.}
\label{fig:other-pseudo}
\end{figure}

Configurations (ii) and (iii) yield slightly lower overall performance, while still consistently outperforming the baseline (Figure~\ref{fig:other-pseudo}).

The distribution of misclassifications (see Appendix \ref{Confusion matrices}) highlights three main patterns, which are consistently observed across all models in the BERT family.
In the \texttt{SIMPLE} configuration, all models exhibit very few classification errors, in line with the high overall accuracies. The vast majority of errors are False Positives, i.e. distractors incorrectly classified as constructions. This pattern suggests that constructional information is robustly encoded in the representations.
The \texttt{PSEUDO} configuration illustrates how strongly the nature of the distractor instances defines the task itself. In this setting, the model is exposed to actual \npn constructions alongside distractors that are only superficially isomorphic but lack constructional status. In the test set structurally distinct constructions are more frequently misclassified. This suggests that, given the type of distractors encountered during training, the classifier might effectively learn to distinguish between constructional and non-constructional instances, rather than on deeper structural or construction-specific properties.
In the \texttt{OTHER} configuration, \prep performance appears to be influenced by an imbalance in the training data, due to the constraints of this configuration. Since most distractors that instantiate a different Cxn are realised with the preposition \textit{a}, instances with \textit{su} are more frequently misclassified. This suggests that \prep representations are sensitive to prepositional frequency effects. By contrast, \unk representations do not seem to exhibit the same bias, indicating a more construction-oriented encoding less dependent on lexical skew.
Taken together, these results highlight the importance of carefully defining what counts as a minimal pair for the Cxn under investigation, in order to faithfully translate the underlying theoretical question into a computational task.
In probing experiments performance is strongly shaped by the experimental setup, in particular by the definition of distractors and minimal pairs. This suggests that probing results should be interpreted with care, as they reflect not only properties of the model, but also properties of the task design.

\section{Disambiguation task}
\label{sec:disambiguation}

Given the very high performance achieved in the experiment about the identification of \npn Cxn, extending the analysis beyond form, we now turn to examining the semantic dimension of the Cxn. Our setup is a multinomial three-class disambiguation problem: we only focus on the Cxn (1), (2), (3) and Cxn (4) in Table \ref{tab:npn_constructions}, which are associated to the three possible meanings of \textit{juxtaposition/contact}, \textit{succession/iteration/distributivity} or \textit{greater\_plurality/accumulation}. 

\begin{figure}[htbp]
    \centering
    
    \begin{subfigure}{0.48\columnwidth}
        \centering
        \includegraphics[width=\linewidth]{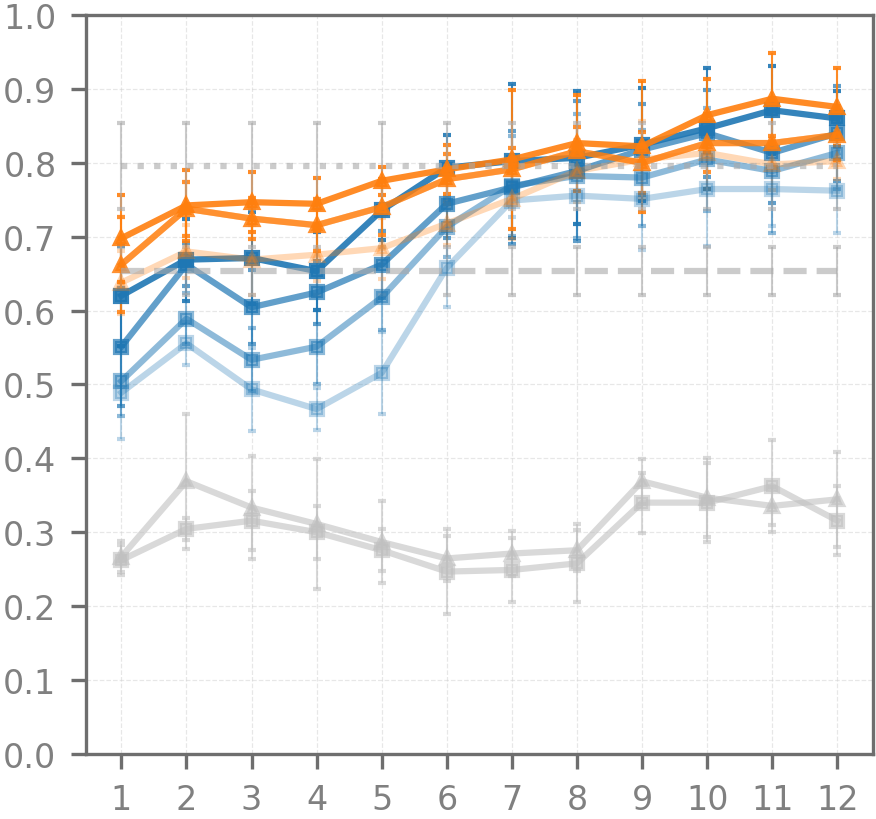}
        \caption{bert-cased-ita}
        \label{fig:g1}
    \end{subfigure}
    \hfill
    \begin{subfigure}{0.48\columnwidth}
        \centering
        \includegraphics[width=\linewidth]{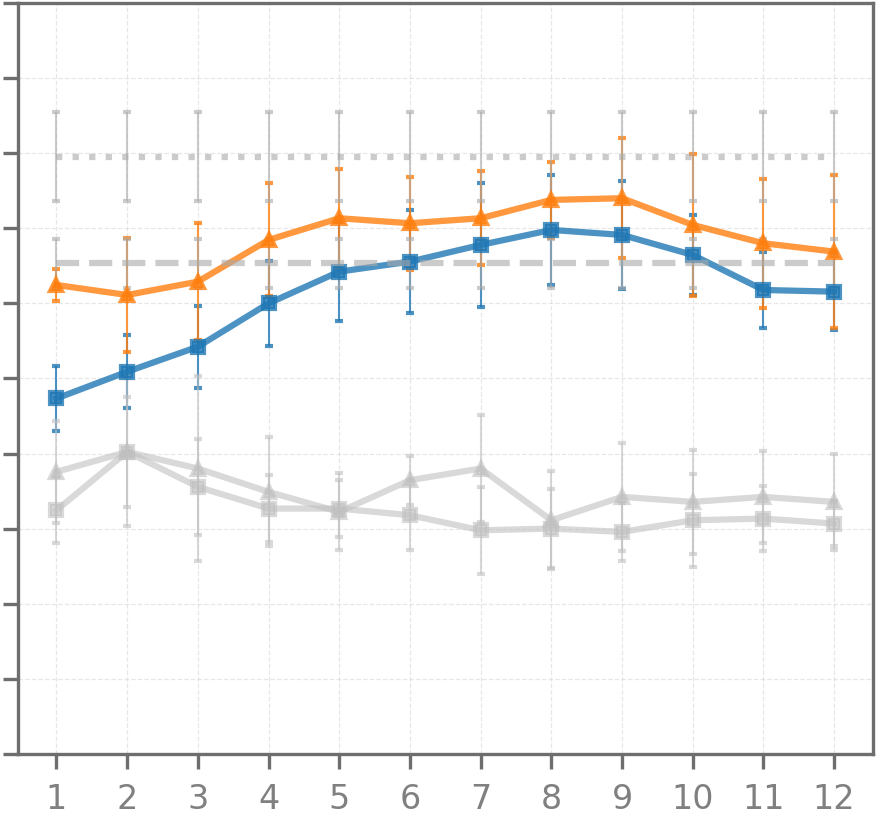}
        \caption{mBERT}
        \label{fig:g2}
    \end{subfigure}
    
    \vspace{0.5em}
    
    \begin{subfigure}{0.48\columnwidth}
        \centering
        \includegraphics[width=\linewidth]{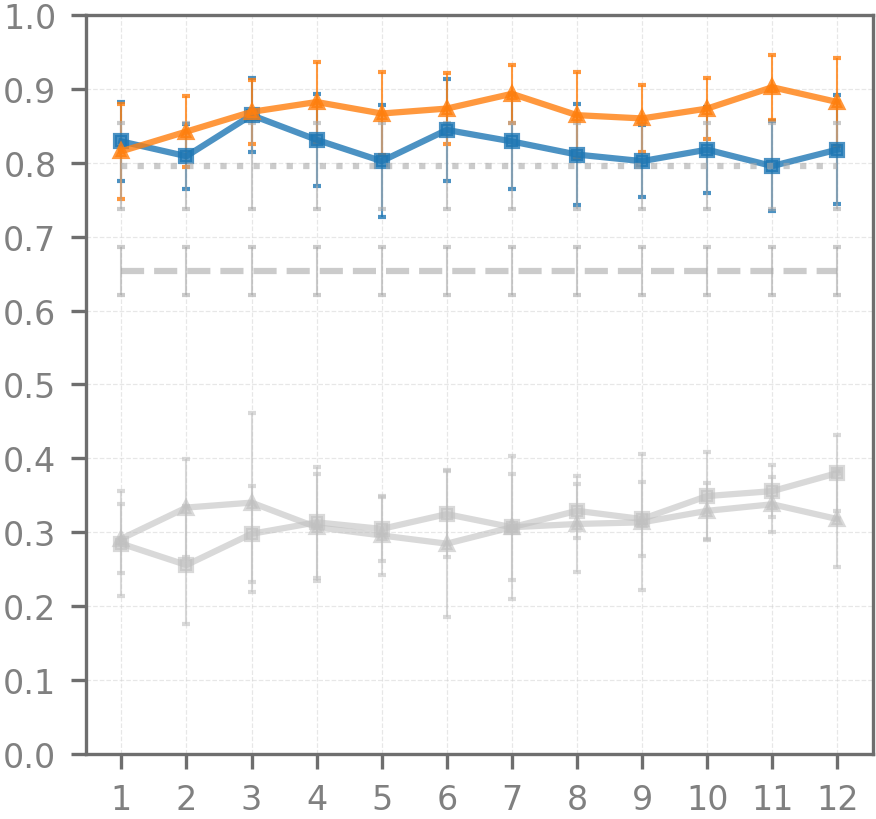}
        \caption{UmBERTo}
        \label{fig:g3}
    \end{subfigure}
    \hfill
    \begin{subfigure}{0.48\columnwidth}
        \centering
        \includegraphics[width=\linewidth]{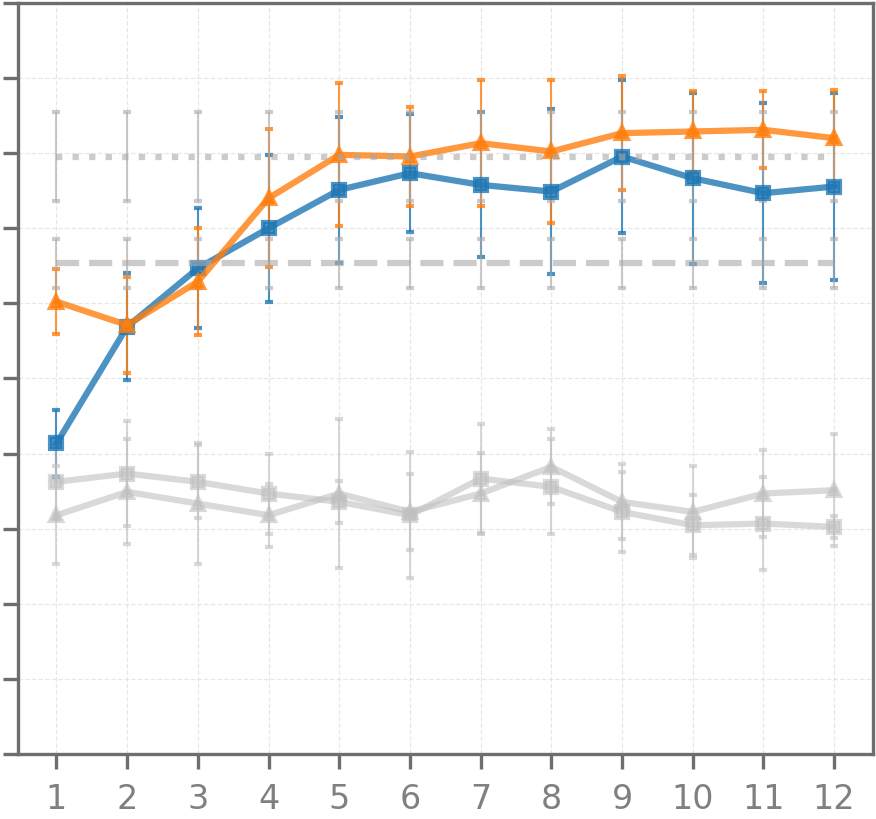}
        \caption{XLM-RoBERTa}
        \label{fig:g4}
    \end{subfigure}
    
    \caption{Accuracy of \unk (red lines, square dots) and \prep (orange lines, triangular dots) on Construction disambiguation task. Dashed grey line represents FastText baseline. Dotted grey line represents morphological FastText baseline. Continuous grey lines refer to control classifiers. Model (\ref{fig:g1}) includes decremental training configurations, line shading becomes progressively lighter as the number of training instances decreases (480 → 240 → 120 → 60). No substantial performance differences emerge across configurations. Figures~\ref{fig:pcaex2UNK} and \ref{fig:PCAex2PREP}, in Appendix, provide a qualitative visualisation of the embedding space across layers.}

    \label{fig:ex2}
\end{figure}

Performance for both \unk and \prep embeddings, as shown in Figure~\ref{fig:ex2}, are above the baselines. In addition to the static baseline computed on the lemma of the reduplicated noun in the Cxn, this experiment includes a stronger baseline condition. Since one of the defining properties distinguishing Cxn (3) from (4) is the grammatical number of the reduplicated noun, we computed an additional FastText baseline using vectors built from the morphologically inflected noun: this proves substantially stronger than the lemma-based one. BERT models surpass it only in the highest layers, suggesting that lower layers do not encode information beyond what is already available from surface morphological cues, while higher layers capture more abstract constructional distinctions.

Error analysis for both \unk and \prep contextual embeddings reveal a highly structured pattern across the BERT family.
The probing classifier achieves near-perfect performance in identifying the \textit{greater\_plurality/accumulation} and \textit{succession/iteration/distributivity} meanings when instantiated with the preposition \textit{su} `on'. 
Although overall performance remains strong, some confusion emerges between \textit{juxtaposition/contact} and \textit{succession/iteration/distributivity} when both are instantiated with the preposition \textit{a} `at/to'. In other words, the model shows greater difficulty in distinguishing between semi-schematic Cxns that share the same surface preposition. Importantly, \textit{greater\_plurality/accumulation} is distinguished not only semantically but also through a salient surface morphological constraint (singular or plural number of the reduplicated noun), which may facilitate classification, as the the additional morphologically informed baseline suggests. Future work could isolate this formal property in a controlled design in order to assess the relative contribution of surface morphology and constructional semantics to probing performance.

A substantial proportion of the errors involves \npn Cxn (1) and (2), for which, respectively, labels \textit{juxtaposition/contact} and \textit{succession/iteration/distributivity} are predicted. These cases frequently involve high-frequency and highly conventionalised types, such as \textit{gomito a gomito} (`elbow to elbow'), \textit{petto a petto} (`chest to chest'), \textit{fronte a fronte} (`forehead to forehead'), and spatial-configuration patterns such as \textit{balcone a balcone} (`balcony to balcony'), \textit{uscio a uscio} (`doorway to doorway'), and \textit{porta a porta} (`door to door').

\ea\label{ex:guancia306}

\textit{Qui lavorano i revisori \textbf{gomito a gomito} con i membri di Cosea}.\\
`Here auditors work \textbf{elbow to elbow} with members of Cosea\footnote{Cosea is an Italian company, active in the Emilia region.}'\\
\textsc{Annotated: juxt./contact;\\
Predicted: succ./iter./distributivity}

\z

\ea\label{ex:corpo91}

\textit{Faccia a faccia e \textbf{petto a petto}, Big Jim serrò le mani sulle braccia di Andy e lo guardò negli occhi.}.\\
`Face to face and \textbf{chest to chest}, Big Jim closed his hands on Andy's arms and looked at him in his eyes.'\\
\textsc{Annotated: juxt./contact;\\
Predicted: succ./iter./distributivity}

\z

Examples (\ref{ex:guancia306}) and (\ref{ex:corpo91}) show that, although the Cxn canonically encodes physical contact or spatial adjacency, the larger event structure evokes dynamic interaction unfolding in time (e.g. working sessions, fighting sequences). The probe may therefore privilege an interpretation compatible with iterativity or sequential event structure over a purely configurational reading.

A similar oscillation emerges in spatial-distribution expressions.

\ea\label{ex:porta}

\textit{Dopo anni passati prima come venditore \textbf{porta a porta} di assicurazioni e di calzature per donne (...) questo è il suo momento.}.\\
`After years spent first as a \textbf{door to door} salesman of insurance policies and women’s shoes (...) this is his moment.'\\
\textsc{Annotated: succ./iter./distributivity;\\
Predicted: juxt./contact}

\z

In these cases, such as Example (\ref{ex:porta}), the model appears to oscillate between an interpretation based on contact or adjacency between entities along a spatial path and  an interpretation based on distribution over successive sites. The alternation suggests that, for entrenched lexicalised types, the boundary between \textit{juxtaposition/contact} and \textit{succession/iteration/distributivity} could be intrinsically gradient rather than categorical.

These results motivate a more focused examination of the \textit{succession/iteration/distributivity} label, as it is compatible with multiple prepositions and does not rely on a strong surface constraint (unlike \textit{greater\_plurality/accumulation}). Its classification therefore provides a more stringent test of whether contextual embeddings capture abstract constructional meaning independently of lexical realisation.

\subsection{Semantic generalisation}

In order to create an experimental setup that allow us to test the probing classifier on unseen semi-schematic Cxns, we annotated 100 instances on Cxns (5) and (7) from Table \ref{tab:npn_constructions} with sentential context extracted from CORIS: these have meaning of \textit{succession/iteration/distributivity} and are realised by the prepositions \textit{per} `by' and \textit{dopo} `after'.

We trained the probing classifier on the full dataset comprising the three semantic labels together with the distractor instances, and then tested it exclusively on the newly annotated \textit{per} `by' and \textit{dopo} `after' instances. 
The composition of this dataset is shown in Appendix ~\ref{tab:perdopo}.

\begin{figure}[htbp]
    \centering
    \includegraphics[width=0.5\linewidth]{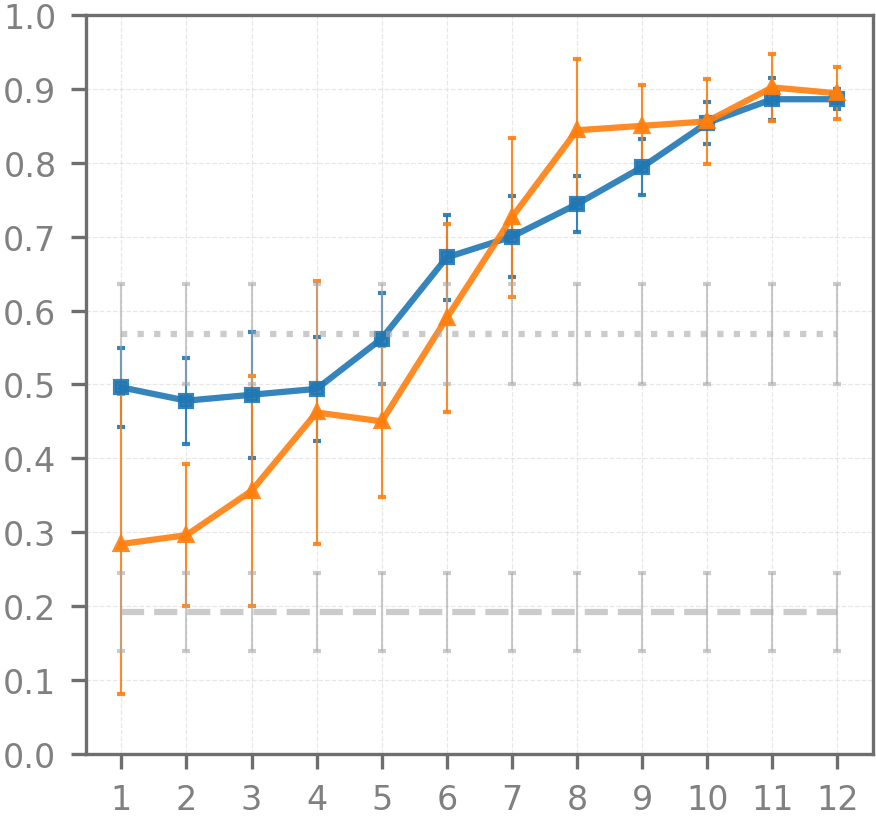}
\caption{Accuracy of \unk (red lines, square dots) and \prep (orange lines, triangular dots) on Construction disambiguation. Dashed grey line represents FastText baseline. Dotted grey line represents morphological FastText baseline.}
\label{fig:ex2_per_dopo}
\end{figure}

Figure~\ref{fig:ex2_per_dopo} reports accuracy across layers for \texttt{[UNK]} and \texttt{PREP} embeddings, together with static baselines. We can observe that \texttt{[UNK]} and \texttt{PREP} representations support robust generalisation to unseen prepositions, with performance reaching high accuracy in late layers. 
The task is intrinsically harder, as it is demonstrated by the drop in performance for both baselines. Nonetheless, results are consistent across different model in BERT family. \texttt{[UNK]} starts from moderate performance in the earliest layers and increases steadily, approaching 0.9 in the final layers. \texttt{PREP} shows a different trajectory: it is markedly weaker and more variable in the lowest layers, where it performs close to baseline, but in 4--8 layers exhibits a sharp improvement, converging with \texttt{[UNK]} in late layers. This pattern suggests that early-layer \texttt{PREP} representations are strongly related to lexical information about the preposition, while higher layers increasingly integrate contextual information that can support a semantics-based decision even when the preposition itself is unseen in training.

These findings provide converging evidence that the semantic information exploited by the probing classifier is not merely a learned association between a small set of prepositions and labels. Instead, the late-layer gains for \texttt{[UNK]} and especially for \texttt{PREP} are consistent with an increasingly abstract encoding of constructional meaning, where the model can map unseen prepositional instantiations (\textit{per} `by', \textit{dopo} `after') onto a semantic category encountered in training (\textit{succession/distributivity/iteration}). Under this interpretation, the experiment supports the conclusion that while lower BERT layers, particularly for \texttt{PREP}, remain more sensitive to surface lexical form, higher layers encode a representation rich enough to sustain generalisation across competing semi-schematic Cxns \citep{masini2024npn} expressing \textit{succession/distribution/iterativity} which can be realised by any of the prepositions \textit{a} `at/to', \textit{su} `on', \textit{per} `by', and \textit{dopo} `after'. 

\section{Conclusion}

We presented two probing experiments addressing the identification and semantic disambiguation of Italian \npn constructions. To this end, we introduced an extended dataset including both constructional instances and carefully designed distractors, allowing for a controlled evaluation of construction-sensitive encoding.

We extended and enriched the setup of \cite{scivetti_Construction_2025} by comparing \unk and \prep contextual embeddings, implementing alternative training configurations grounded in different minimal-pair and distractor definitions, and introducing a generalisation test on unseen prepositions instantiating unseen semi-schematic Cxns.

The aim of this study is to investigate whether the internal representations of models in the BERT family are compatible with a constructionist perspective, with respect to the Italian \npn constructional family. This question is addressed through a set of experiments targeting both form-related and meaning-related properties of the constructions.

The results suggest that distinguishing constructional patterns relies on the joint contribution of multiple factors, including lexical distribution and morphological cues. As indicated by the baseline performance, none of these factors in isolation is sufficient to fully resolve the task; rather, they appear to jointly support the information that is recoverable for classification.

The identification results show that constructional information is detectable in contextual embeddings. However, they also demonstrate that performance is highly sensitive to the composition of the training set: the nature of the distractors and the operationalisation of minimal pairs substantially shape the classification task. In this respect, careful interpretative caution is required, in line with recent observations on task sensitivity in probing studies (see also \citet{dunn2025llms}).

The disambiguation experiment further reveals that static embeddings constitute strong baselines. Lexical information, especially when coupled with morphological cues such as number of the reduplicated noun, proves highly informative and can account for a significant portion of the variance. 
Crucially, however, when the task involves generalisation to unseen semi-schematic constructions instantiating the same semantic value, contextual embeddings enable the classifier to achieve high accuracy precisely where static vectors show a marked performance drop. This suggests that higher layers encode more abstract constructional regularities that go beyond type-level lexical information.

Taken togheter, the results are consistent with a constructionist perspective, in that successful classification appears to rely on the joint contribution of multiple cues rather than on isolated lexical features.

Multilingual models (mBERT and XLM-R respectively) deserve specific attention: while their performance remain largely comparable to monolingual models in the identification task, the situation appears different for disambiguation. In disambiguation task, in fact, both mBERT and XLM-R underperform their monolingual counterpart. While the difference is negligible for English, the drop in performance is particularly evident for Italian data. This confirms the importance of extending interpretability investigations to languages other than English: taking a language-specific perspective like the one that CxG offers can help evaluating language-specific abilities of multilingual models.

Overall, the findings seem to provide empirically grounded support for an abstraction path in PLMs towards an higher-level construction linking the semi-schematic Cxns associated with the semantics of \textit{succession/distribution/iterativity}, realized by the prepositions \textit{a} `at/to', \textit{su} `on', \textit{per} `by', and \textit{dopo} `after' within the \npn family.

At the same time, the results remain confined to the representational space within BERT's models and any theoretical implications must therefore be interpreted with caution. 

In order to provide empirical grounding for the theoretical debate concerning the schematization of the \npn construction, the probing results are currently being cross-validated through behavioural experiments that directly compare human judgments with model predictions. This step will be essential to determine whether the computational evidence actually reflects cognitively plausible constructional abstractions and to more rigorously assess the theoretical validity of the proposed \npn schematization.

\section{Limitations}

The present study is subject to several limitations.

First, the analysis is restricted to a single constructional family, namely the Italian \npn Cxs. Although multiple prepositions (\textit{a} `at/to', \textit{su} `on', \textit{per} `by', \textit{dopo} `after') are included, they instantiate closely related constructions within the same constructional network, differing primarily in degree of productivity and conventionalisation. As such, the findings should be interpreted as evidence about this specific constructional domain rather than as a general account of constructional information in internal representations of the model. Extending the analysis to Cxns with different syntactic profiles and levels of schematicity would be necessary to assess the broader validity of the observed patterns.
This is especially relevant given that \npn Cxns predominantly fulfil an adverbial function, which limits the diversity of syntactic configurations examined.

Second, the study focuses on models within the BERT family. While this allows for controlled comparison across variants, these models share core architectural and training properties, including encoder-only structure and a masked language modelling objective. Consequently, the results cannot be straightforwardly generalised to other classes of models. In particular, architectures with different training objectives or representational dynamics may encode constructional information differently. Future work should therefore extend the analysis to decoder-only models, in order to assess whether the results observed here are specific to masked language modelling or reflect more general properties of transformer-based representations, including those shaped by next-token prediction.

Third, the methodology is limited to diagnostic probing. This design choice ensures comparability with prior work on the English \npn construction \citep{scivetti_Construction_2025}, but it also constrains the scope of the conclusions. In particular, it remains unclear whether the information encoded in the model’s internal representations and exploited by the probe is causally involved in the model’s behaviour on the task. In other words, while probing suggests that the representational space can be analyzed in terms of constructional information, 
it does not establish that the model itself relies on this information when making predictions.



\section{Ethics Statement}

Annotators were recruited within an advanced Master's-level course as part of structured educational activities. Participation was entirely voluntary and had no impact on students' evaluation or academic standing. All participants were informed about the objectives of the study and the intended use of the collected data.

\section{Bibliographical References}\label{sec:reference}

\bibliographystyle{lrec2026-natbib}
\bibliography{lrec2026-example}

\newpage
\section*{Appendix}
\addcontentsline{toc}{section}{Appendix}
\label{sec:appendix}

\setcounter{subsection}{0}
\renewcommand{\thesubsection}{\Alph{subsection}}
\renewcommand{\thesection}{}

\makeatletter
\renewcommand{\thefigure}{\Alph{subsection}.\arabic{figure}}
\@addtoreset{figure}{subsection}
\renewcommand{\thetable}{\Alph{subsection}.\arabic{table}}
\@addtoreset{table}{subsection}
\makeatother

\subsection{Distractors}
\label{sec:distractors}

\begin{tcolorbox}[	
  title={\emph{\textsc{PNPN}} - Construction - (\textit{a})},
  colback=red!2!white,   
  colframe=red!25!black,  
  boxrule=0.5pt,
  arc=1mm,
  left=3pt, right=3pt, top=3pt, bottom=3pt, breakable
]
\textbf{Description:} 

The \textsc{pnpn} Construction \([P\ N_{i}\ P\ N_{i}]\) encodes a meaning of \textit{iterativity}, \textit{succession}, and \textit{transition} between elements belonging to the same semantic category. In this Construction, the nominal slots are occupied by the same noun, repeated in a discontinuous form, while the preceding preposition ---typically \textit{a} or \textit{da} --- specifies the direction or type of relation between the two occurrences.

\vspace{1.8mm}
\hrule
\vspace{1.8mm}

\textbf{Example (ID 720):}
\emph{Il tutto tra l'altro avviene nel giro di pochi mesi con una successione \textbf{da agenzia ad agenzia} quasi automatica} \\
`All of this, moreover, happens within a few months with an almost automatic succession \textbf{from agency to agency}'.
\end{tcolorbox}

\begin{tcolorbox}[
  title={\emph{Verbal} - Construction - (\textit{a})},
  colback=red!2!white,   
  colframe=red!25!black,  
  boxrule=0.5pt,
  arc=1mm,
  left=3pt, right=3pt, top=3pt, bottom=3pt, breakable
]
\textbf{Description:} 

The verbal Construction \([V\ N_{i}\ a\ N_{i}]\) is realised by a restricted set of transitive verbs that allow both a direct object and a prepositional complement introduced by \textit{a}. Among the most frequent are verbs of accumulation or transfer, such as \textit{aggiungere} (`to add'), \textit{unire} (`to join'), \textit{sommare} (`to sum'), and \textit{accostare} (`to bring together'). This Construction exhibits a high degree of semantic conventionalisation: the identity of the nouns generates an intensifying or cumulative meaning that goes beyond the literal meaning of the verb.

\vspace{1.8mm} 
\hrule
\vspace{1.8mm}

\textbf{Example (ID 1027):}
\emph{Quel fatto \textbf{aggiunse irritazione a irritazione}.} \\
`That fact \textbf{added irritation to irritation}'.

\end{tcolorbox}

\begin{tcolorbox}[
  title={\emph{NUM P NUM} - isomorphic pattern — (\textit{a})},
  colback=red!2!white,   
  colframe=red!25!black,  
  boxrule=0.5pt,
  arc=1mm,
  left=3pt, right=3pt, top=3pt, bottom=3pt, breakable
]
\textbf{Description:} 

This pattern expresses a relationship of \textbf{equivalence} or \textbf{balance} between two numerical values that are identical or opposed (identical, in the case of superficial identity with \textit{\npn})

\vspace{1.8mm}
\hrule
\vspace{1.8mm}

\textbf{Example (ID 1498):}
\emph{Il risultato non si sbloccava e si era sempre sullo \textbf{zero a zero}.} \\
`The result did not get unlocked and it was always at \textbf{zero to zero}'.

\end{tcolorbox}

\begin{tcolorbox}[
  title={\emph{N extended} - isomorphic pattern - (\textit{a}, \textit{su})},
  colback=red!2!white,   
  colframe=red!25!black,  
  boxrule=0.5pt,
  arc=1mm,
  left=3pt, right=3pt, top=3pt, bottom=3pt, breakable
]
\textbf{Description:} 

These instances superficially replicate the \npn Construction, but one or both occurrences of the noun are part of a larger complex phrase or multiword expression and do not constitute an autonomous instance of the same syntactic category. Consequently, they do not convey the iterative, distributive, or juxtapositional relations that define actual \npn constructions.

\vspace{1.8mm} 
\hrule
\vspace{1.8mm}

\textbf{Example (ID 229):}
\emph{Le due telecamere alternavano \textbf{primi piani a piani a figura intera} da angolazioni diverse.} \\
`The two cameras alternated \textbf{close-ups to full shots} from different angles'.\\
\textbf{Example (ID 2193):}
\emph{La Commissione può ora rimborsare \textbf{spese generali di base su base forfetaria} o versare importi forfetari per piccoli progetti.} \\
`The Commission can now reimburse basic overhead costs on a flat-rate basis or pay lump-sum amounts for small projects'.

\end{tcolorbox}






\begin{tcolorbox}[
  title={\emph{Proper name inglobation} - isomorphic pattern — (\textit{su})},
  colback=red!2!white,   
  colframe=red!25!black,  
  boxrule=0.5pt,
  arc=1mm,
  left=3pt, right=3pt, top=3pt, bottom=3pt, breakable
]
\textbf{Description:} 

Accidental lexical repetition in which one or both noun occurrences belong to a proper name. In such cases, the repetition has no constructive or syntactic function, resulting merely from the contiguous presence of the same lemma in two elements of the sentence.

\vspace{1.8mm} 
\hrule
\vspace{1.8mm}

\textbf{Example (ID 2189):}
\emph{La posizione del \textbf{Comune di Arezzo} \textbf{su Arezzo Fiere} è chiarissima.} \\
`The position of the Municipality of Arezzo on Arezzo Fiere is very clear'.

\end{tcolorbox}

\begin{tcolorbox}[
  title={\emph{N su N giù} - Construction - (\textit{su})},
  colback=red!2!white,   
  colframe=red!25!black,  
  boxrule=0.5pt,
  arc=1mm,
  left=3pt, right=3pt, top=3pt, bottom=3pt, breakable
]
\textbf{Description:} 

Combines nominal repetition with the directional pair \textit{su–giù} (`up, down'), producing a rhythmic and iterative effect. It typically occurs in colloquial or performative contexts, marking repetition or cyclic succession of events or actions, but can also convey an iterative or ironic meaning, evoking repetition and excess in a rhythmically salient way. The su–giù pair contributes to a sense of saturation and cyclical movement.

\vspace{1.8mm}
\hrule
\vspace{1.8mm}

\textbf{Example (ID 2174):}
\emph{Come se ci fosse bisogno di una voce dall’ alto che dica \textbf{"bandiere su
bandiere giù"}.} \\
`As if there were a need for a voice from above saying 'flags up, flags down' '.

\end{tcolorbox}

\begin{tcolorbox}[
  title={\emph{Num P Num} - isomorphic pattern — (\textit{su})},
  colback=red!2!white,   
  colframe=red!25!black,  
  boxrule=0.5pt,
  arc=1mm,
  left=3pt, right=3pt, top=3pt, bottom=3pt, breakable
]
\textbf{Description:} 

Expresses a proportional or evaluative relation between two numerical values, typically used in quantificational or performative contexts such as voting, scores, or results. The preposition \emph{su} introduces the reference term, establishing a part-whole or ratio relation (“x obtained out of y possible”). When the two numerals are identical, the expression acquires idiomatic and highly conventionaliased value and is especially productive in sports and journalistic language.

\vspace{1.8mm} 
\hrule
\vspace{1.8mm}

\textbf{Example (ID 2127):}
\emph{Per nostra fortuna sono stati fecondati \textbf{3 su 3}, il 100\%, pronti per essere
impiantati in utero.} \\
`Fortunately for us, 3 out of 3, 100'\%, were fertilized, ready to be implanted in the uterus'.

\end{tcolorbox}

\begin{tcolorbox}[
  title={\emph{Thematic target} - isomorphic pattern — (\textit{su})},
  colback=red!2!white,   
  colframe=red!25!black,  
  boxrule=0.5pt,
  arc=1mm,
  left=3pt, right=3pt, top=3pt, bottom=3pt, breakable
]
\textbf{Description:} 

The surface structure N-su-N is preserved, but the function does not correspond to the canonical meanings of the Construction (plurality, iteration, accumulation, distributivity, or connection). These distractors feature a second noun specifying the thematic domain.

\vspace{1.8mm}
\hrule
\vspace{1.8mm}

\textbf{Example (ID 2088):}
\emph{Comunque non spetta a questa introduzione dire delle nostre tre autrici: qui si tratta piuttosto di chiedersi perché, a voler parlare di romanzo popolare in Italia tra i due secoli, si sono scelte tre donne che scrivono per \textbf{donne su donne}.} \\
`Anyway, it is not the role of this introduction to say anything about our three authors: here, it is rather a question of asking why, when one wants to speak of popular novels in Italy between the two centuries, three women who write for women on women were chosen'.

\end{tcolorbox}

\subsection{Training and Test sets}

\begin{table}[h]
\centering

\begin{tabular}{l l r r}
\hline
\textbf{Label} & \textbf{Prep} & \textbf{Train} & \textbf{Test} \\
\hline
CXN   & \textit{a}  & 120 & 30  \\
CXN   & \textit{su} & 120 & 30 \\
\hline
CXN (total) & -- & 240 & 60\\
\hline
DISTR & \textit{a}  & 120 & 30  \\
DISTR & \textit{su} & 120 & 30 \\
\hline
DISTR (total) & -- & 240 & 60  \\
\hline
\textbf{total} & -- & \textbf{480} & \textbf{120} \\
\hline
\end{tabular}
\caption{\texttt{SIMPLE} training test split configuration}
\label{tab:simple}
\end{table}

\begin{table}[h]
\centering
\begin{tabular}{l l r r }
\hline
\textbf{Label} & \textbf{Prep} & \textbf{Train} & \textbf{Test} \\
\hline
CXN   & \textit{a}  & 70  & 30  \\
CXN   & \textit{su} & 70  & 30  \\
\hline
CXN (total) & -- &140  & 60 \\
\hline
DISTR & \textit{a}  & 25  & 30  \\
DISTR & \textit{su} & 115  & 30 \\
\hline
DISTR (total) & -- & 140  & 60 \\
\hline
\textbf{total} & -- & \textbf{280}  & \textbf{120} \\
\hline
\end{tabular}

\caption{\texttt{PSEUDO} training test split configuration}
\label{tab:pseudo}
\end{table}

\begin{table}[h]
\centering

\begin{tabular}{l l r r}
\hline
\textbf{Label} & \textbf{Prep} & \textbf{Train} & \textbf{Test} \\
\hline
CXN   & \textit{a}  & 55 & 30  \\
CXN   & \textit{su} & 55 & 30 \\
\hline
CXN (total) & -- & 110 & 60 \\
\hline
DISTR & \textit{a}  & 105 & 30  \\
DISTR & \textit{su} & 5 & 30 \\
\hline
DISTR (total) & -- & 110 & 60\\
\hline
\textbf{total} & -- & \textbf{220} & \textbf{120} \\
\hline
\end{tabular}

\caption{\texttt{OTHER} training test split configuration}
\label{tab:other}
\end{table}

\begin{table}[h]
\centering

\begin{tabular}{l l r r}
\hline
\textbf{Label} & \textbf{Prep} & \textbf{Train} & \textbf{Test} \\
\hline
Succession   & \textit{a}  & 60 & 15  \\
Succession   & \textit{su} & 60 & 15  \\
\hline
Succession (total) & -- & 120 & 30 \\
\hline
Accumulation & su & 120 & 30 \\
\hline
Juxtaposition & a & 120 & 30 \\
\hline
\textbf{total} & -- & \textbf{360} & \textbf{90} \\
\hline
\end{tabular}

\caption{Semantic disambiguation train--test split configuration.}
\label{tab:ex2split}
\end{table}

\begin{table}[t]
\centering
\begin{tabular}{l l r r}
\hline
\textbf{Label} & \textbf{Prep} & \textbf{Train} & \textbf{Test} \\
\hline
Succession   & \textit{a}  & 30 & --  \\
Succession   & \textit{su} & 30 & --  \\
Succession   & \textit{per}  & -- & 50  \\
Succession   & \textit{dopo} & -- & 50  \\
\hline
Succession (total) & -- & 60 & 100 \\
\hline
Accumulation & \textit{su} & 60 & -- \\
\hline
Juxtaposition & \textit{a} & 60 & -- \\
\hline
Distractor & \textit{a} & 30 & -- \\
Distractor & \textit{su} & 30 & -- \\
\hline
\textbf{Training set (total)} & -- & \textbf{360} & \textbf{100} \\
\hline
\end{tabular}

\caption{\textit{per} and \textit{dopo} train--test split configurations.}
\label{tab:perdopo}
\end{table}

\onecolumn

\subsection{PCA}
\label{sec:pca}

\begin{figure}[h]
\centering

\begin{subfigure}{0.28\textwidth}
  \centering
  \includegraphics[width=\linewidth]{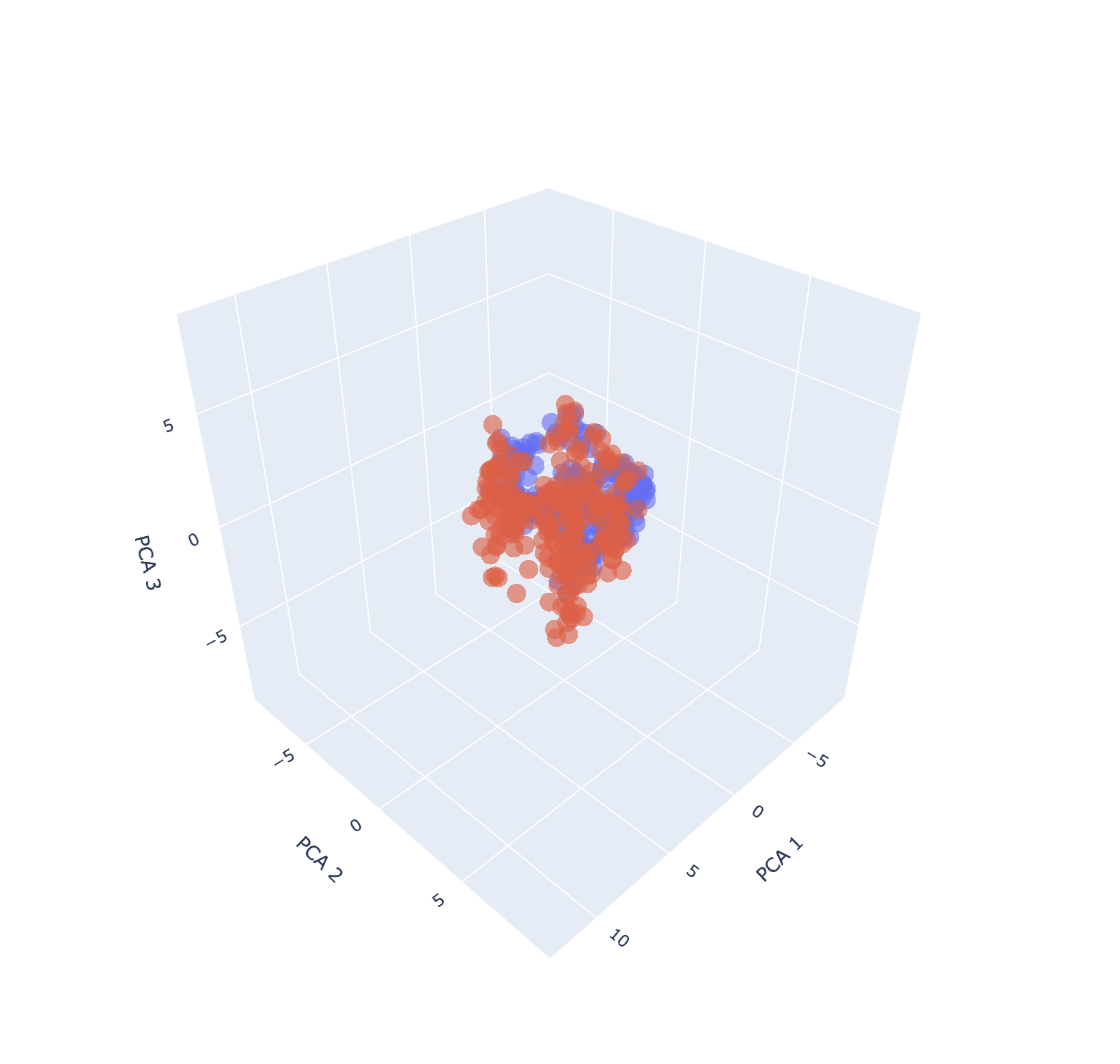}
  \caption{Layer 1}
\end{subfigure}\hfill
\begin{subfigure}{0.28\textwidth}
  \centering
  \includegraphics[width=\linewidth]{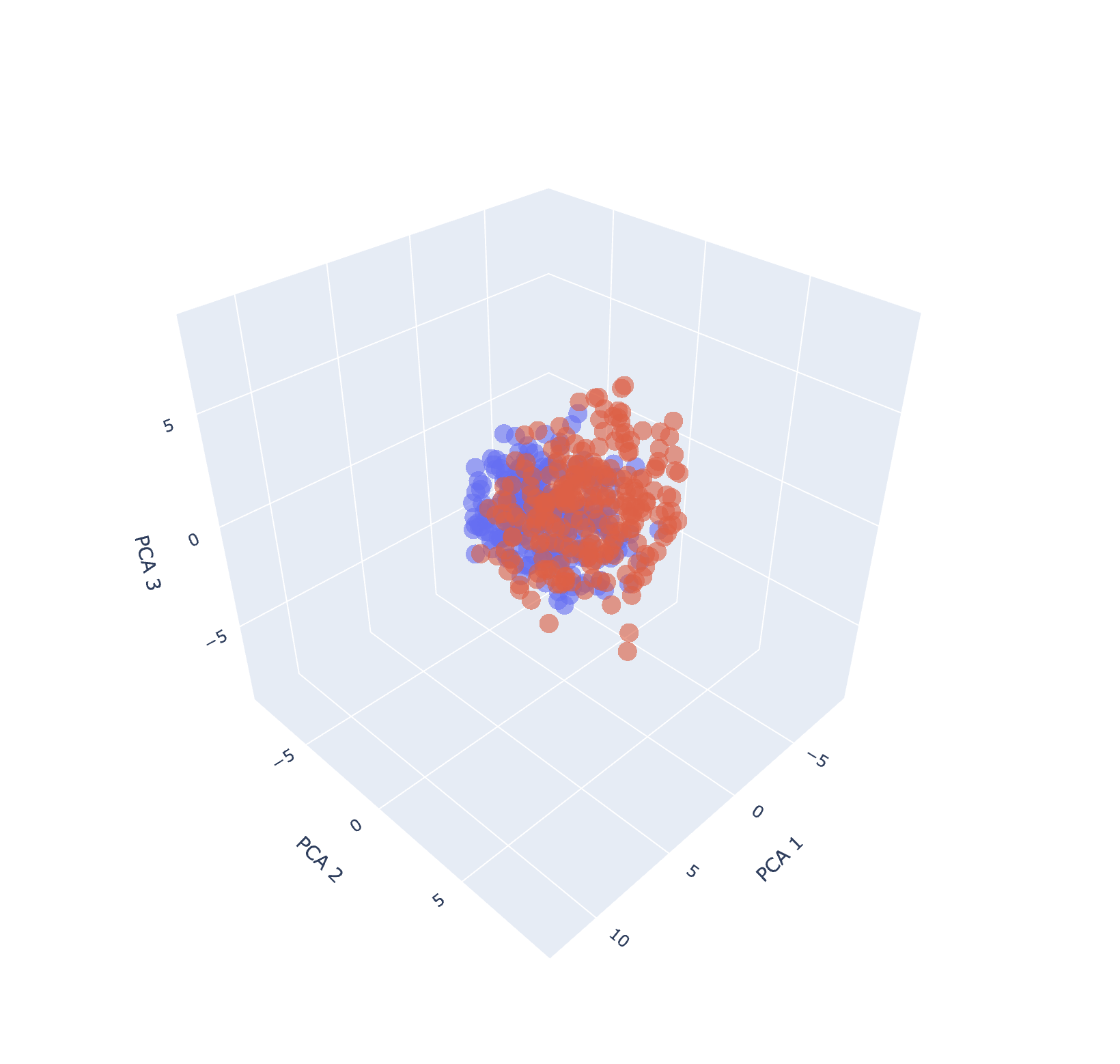}
  \caption{Layer 2}
\end{subfigure}\hfill
\begin{subfigure}{0.28\textwidth}
  \centering
  \includegraphics[width=\linewidth]{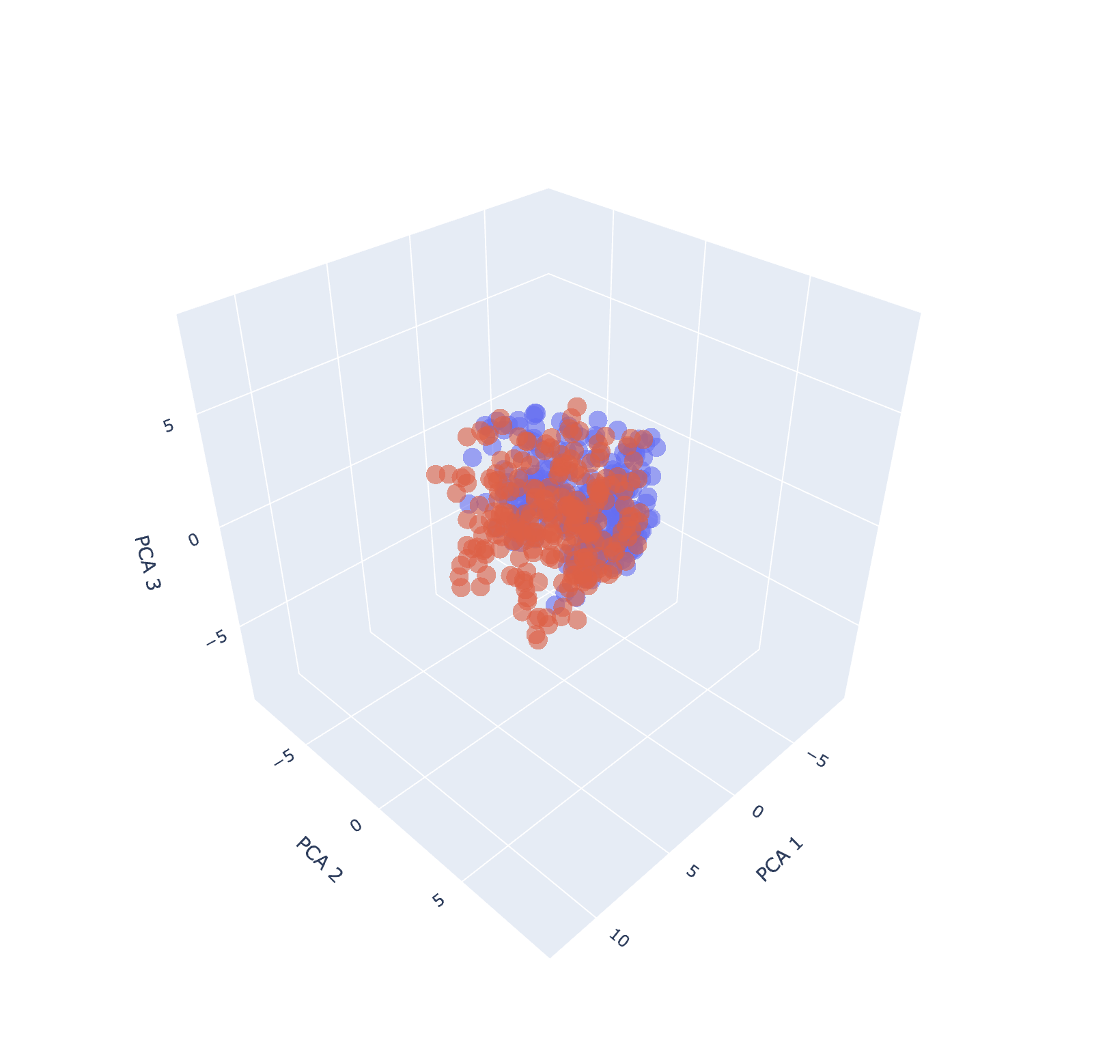}
  \caption{Layer 3}
\end{subfigure}


\begin{subfigure}{0.28\textwidth}
  \centering
  \includegraphics[width=\linewidth]{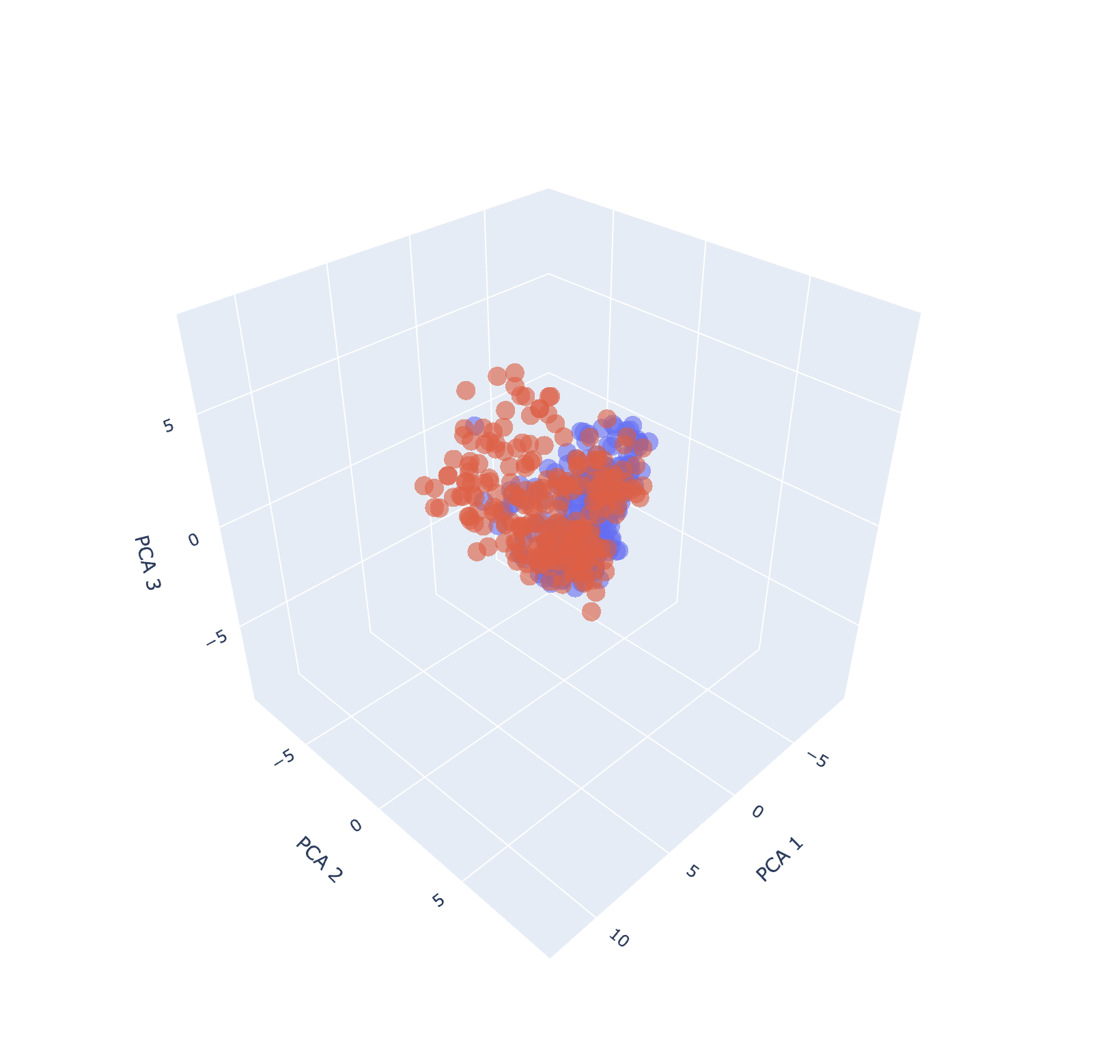}
  \caption{Layer 4}
\end{subfigure}\hfill
\begin{subfigure}{0.28\textwidth}
  \centering
  \includegraphics[width=\linewidth]{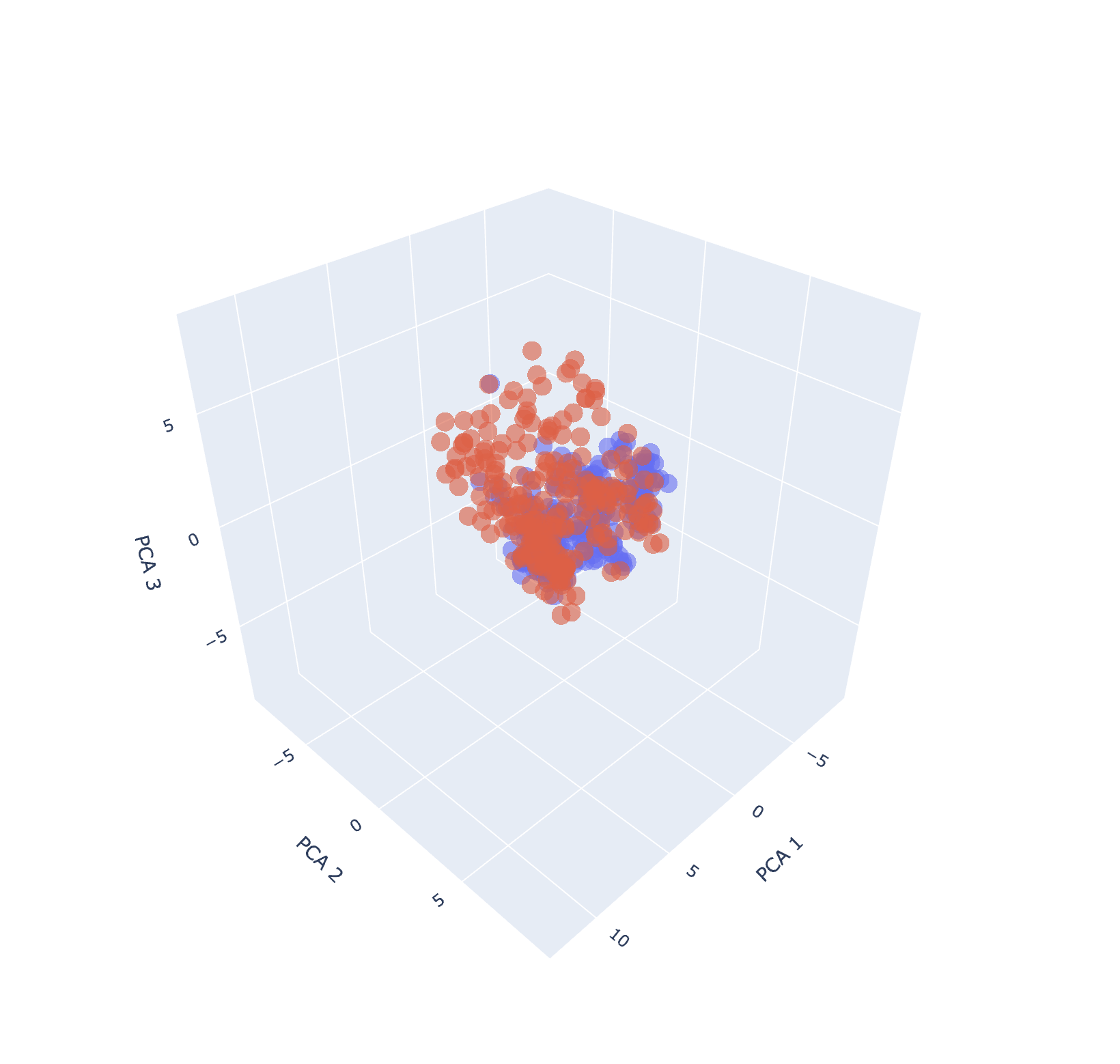}
  \caption{Layer 5}
\end{subfigure}\hfill
\begin{subfigure}{0.28\textwidth}
  \centering
  \includegraphics[width=\linewidth]{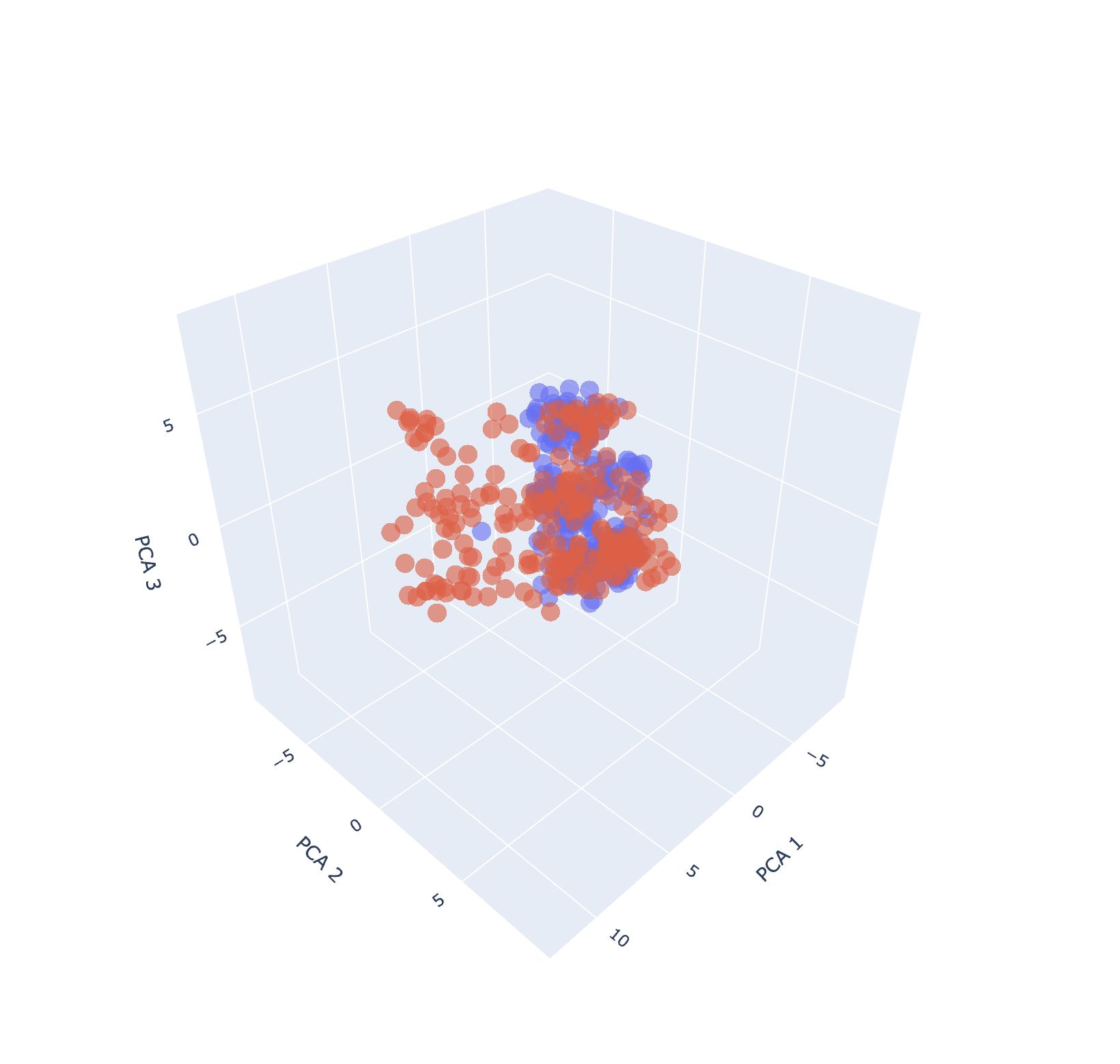}
  \caption{Layer 6}
\end{subfigure}


\begin{subfigure}{0.28\textwidth}
  \centering
  \includegraphics[width=\linewidth]{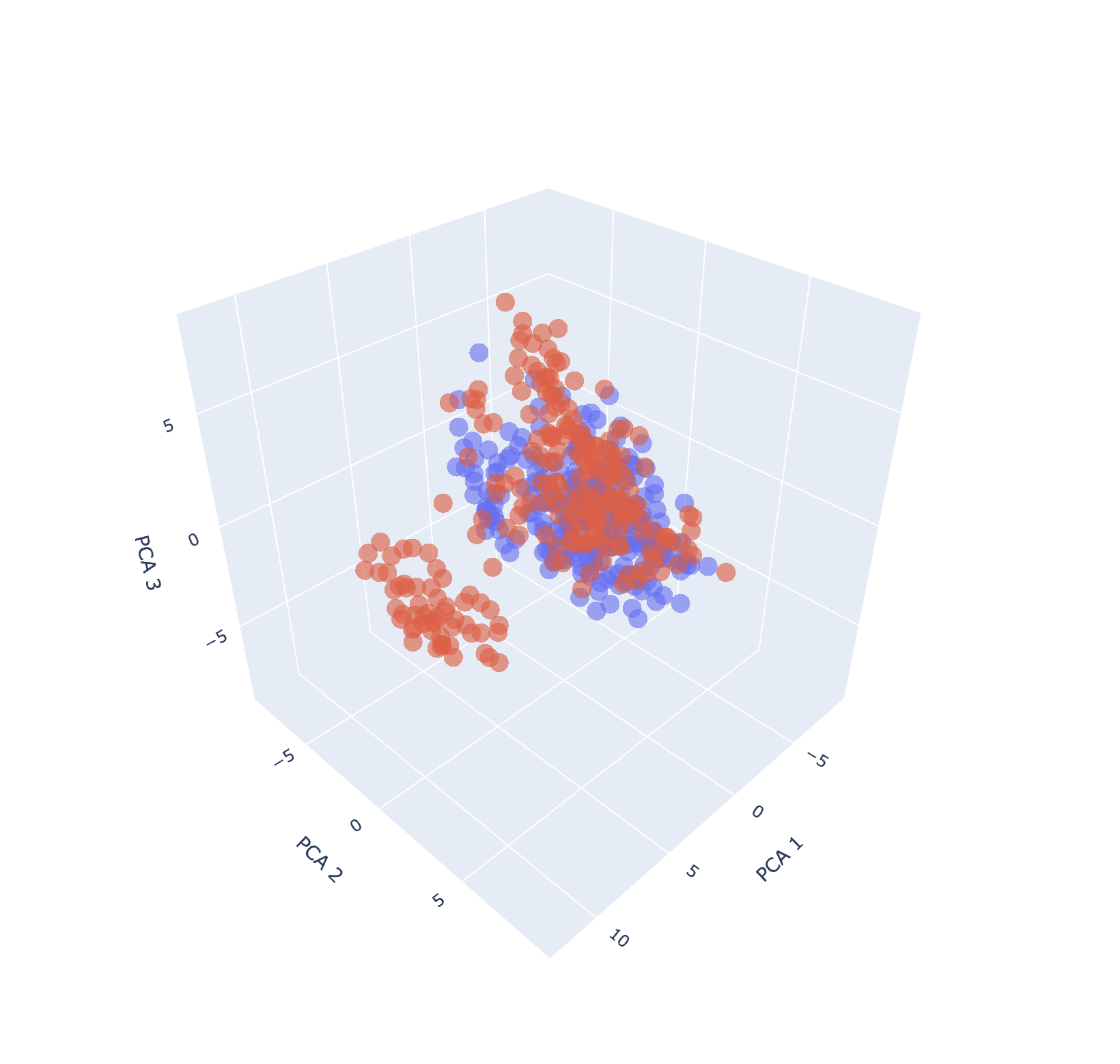}
  \caption{Layer 7}
\end{subfigure}\hfill
\begin{subfigure}{0.28\textwidth}
  \centering
  \includegraphics[width=\linewidth]{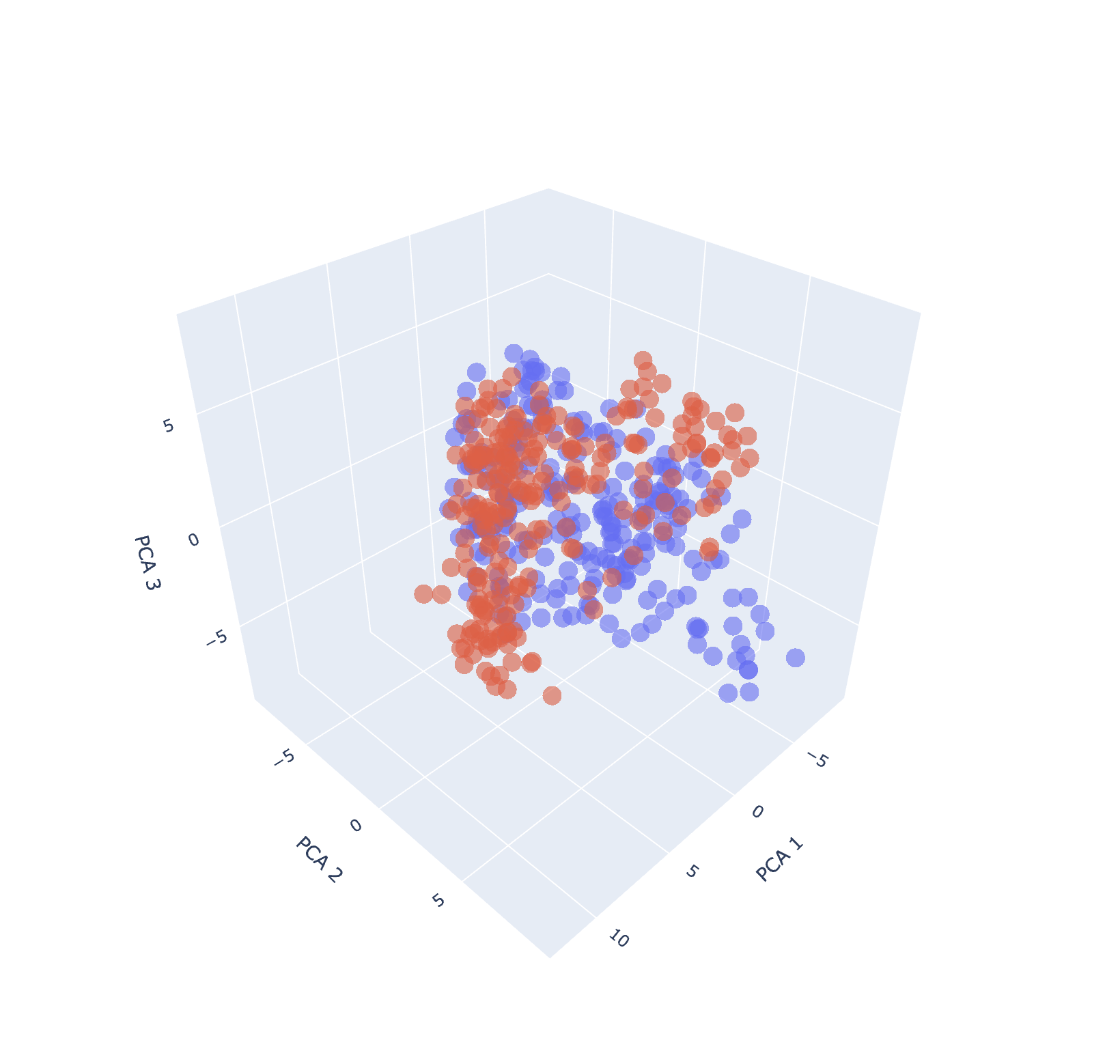}
  \caption{Layer 8}
\end{subfigure}\hfill
\begin{subfigure}{0.28\textwidth}
  \centering
  \includegraphics[width=\linewidth]{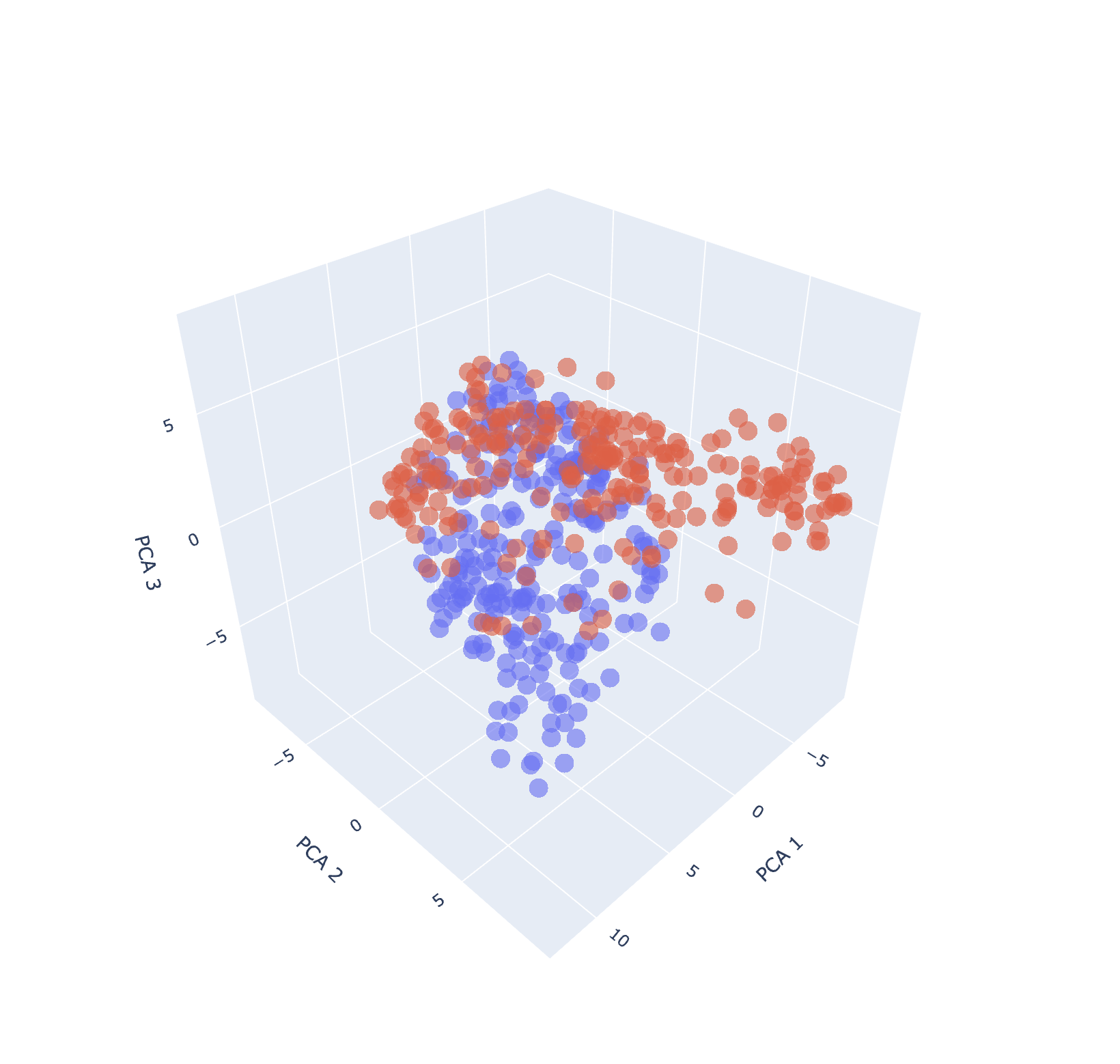}
  \caption{Layer 9}
\end{subfigure}

\vspace{0.3cm}

\begin{subfigure}{0.28\textwidth}
  \centering
  \includegraphics[width=\linewidth]{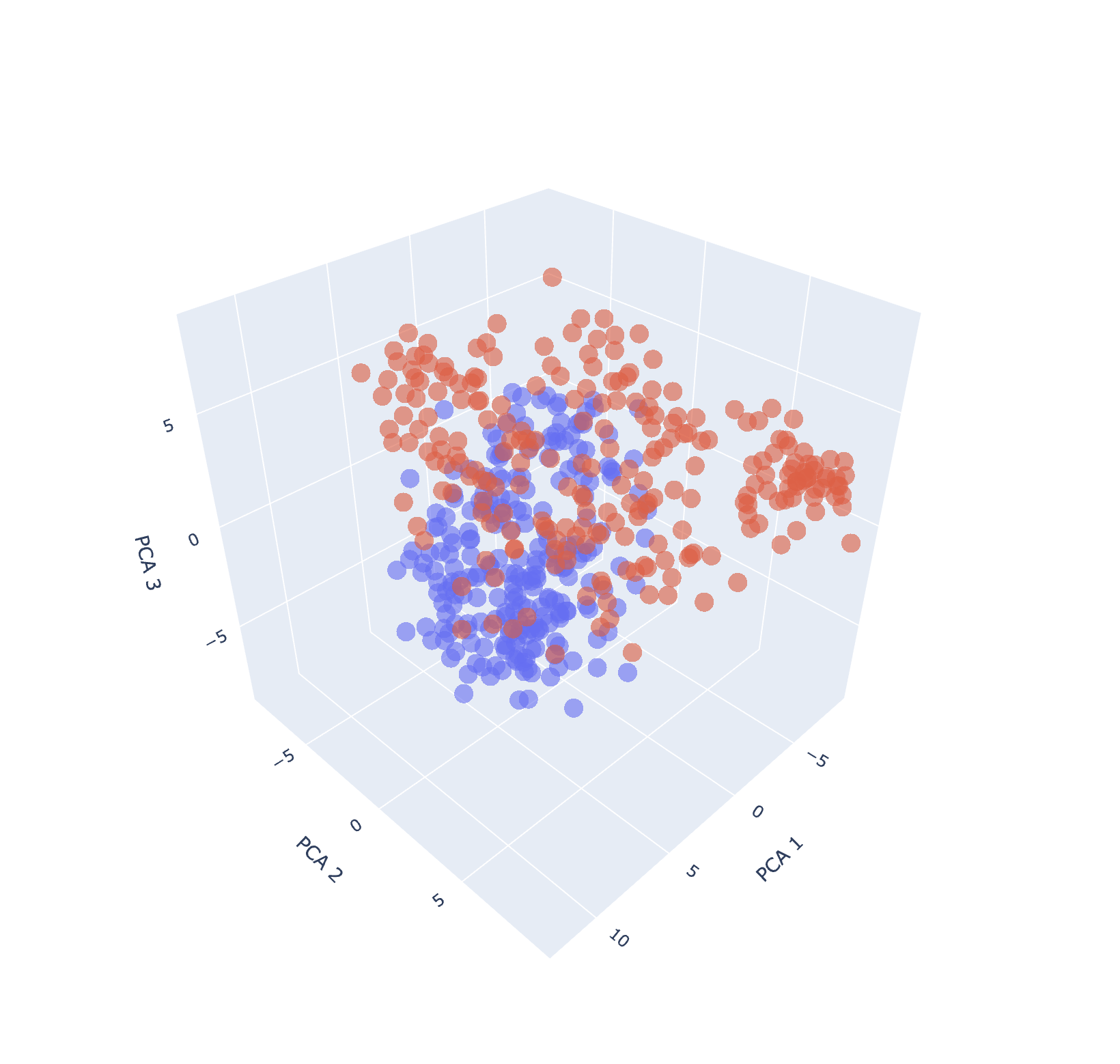}
  \caption{Layer 10}
\end{subfigure}\hfill
\begin{subfigure}{0.28\textwidth}
  \centering
  \includegraphics[width=\linewidth]{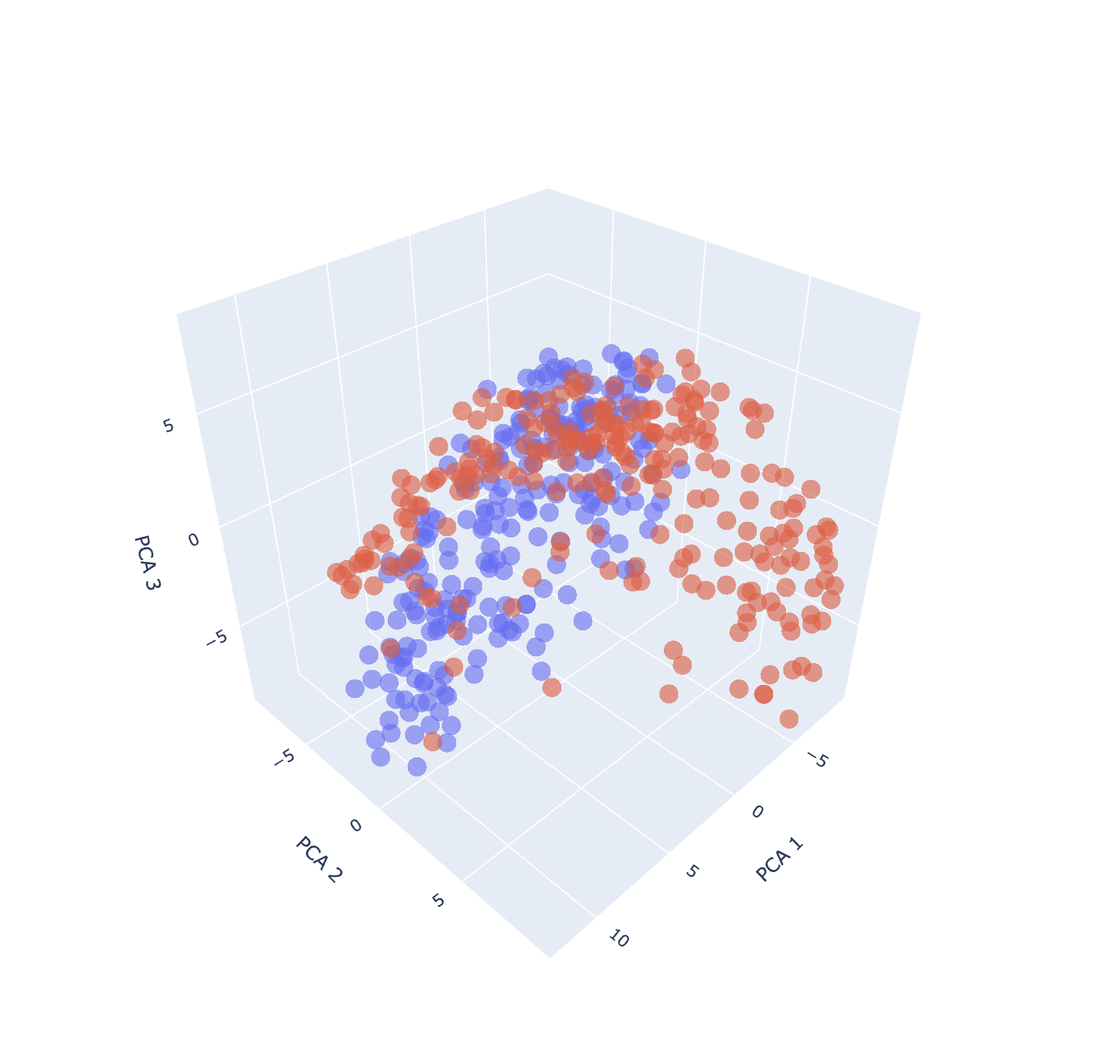}
  \caption{Layer 11}
\end{subfigure}\hfill
\begin{subfigure}{0.28\textwidth}
  \centering
  \includegraphics[width=\linewidth]{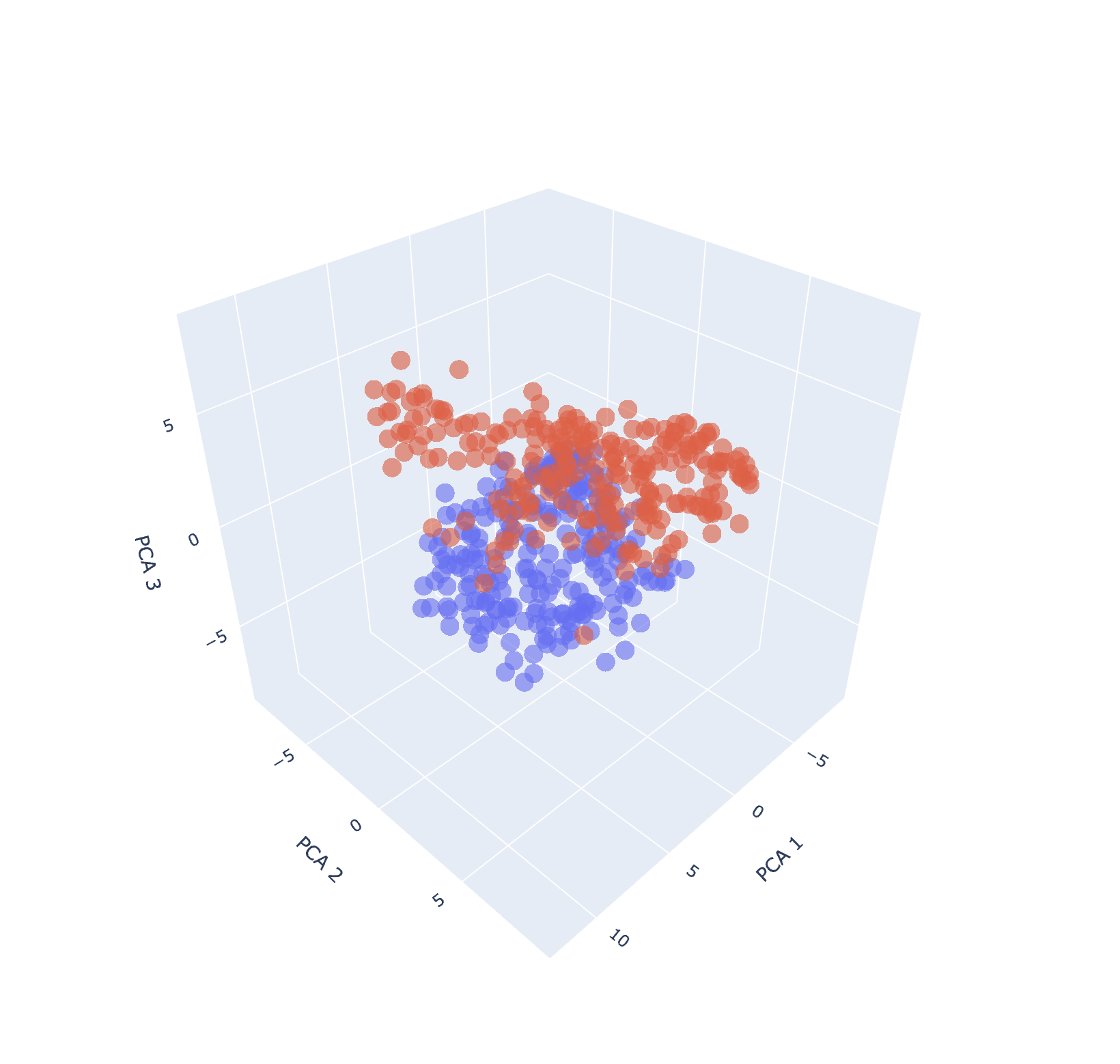}
  \caption{Layer 12}
\end{subfigure}

\caption{PCA-based three-dimensional projection of UNK BERT embeddings across layers for \npn constructions (red) and distractors (red). An animated version of the visualisation is available at \url{https://gretagorzoni00.github.io/NPN_contextual_embeddings/}.}
\label{fig:pcaex1UNK}
\end{figure}

\begin{figure}[h]

\centering

\begin{subfigure}{0.3\textwidth}
  \centering
  \includegraphics[width=\linewidth]{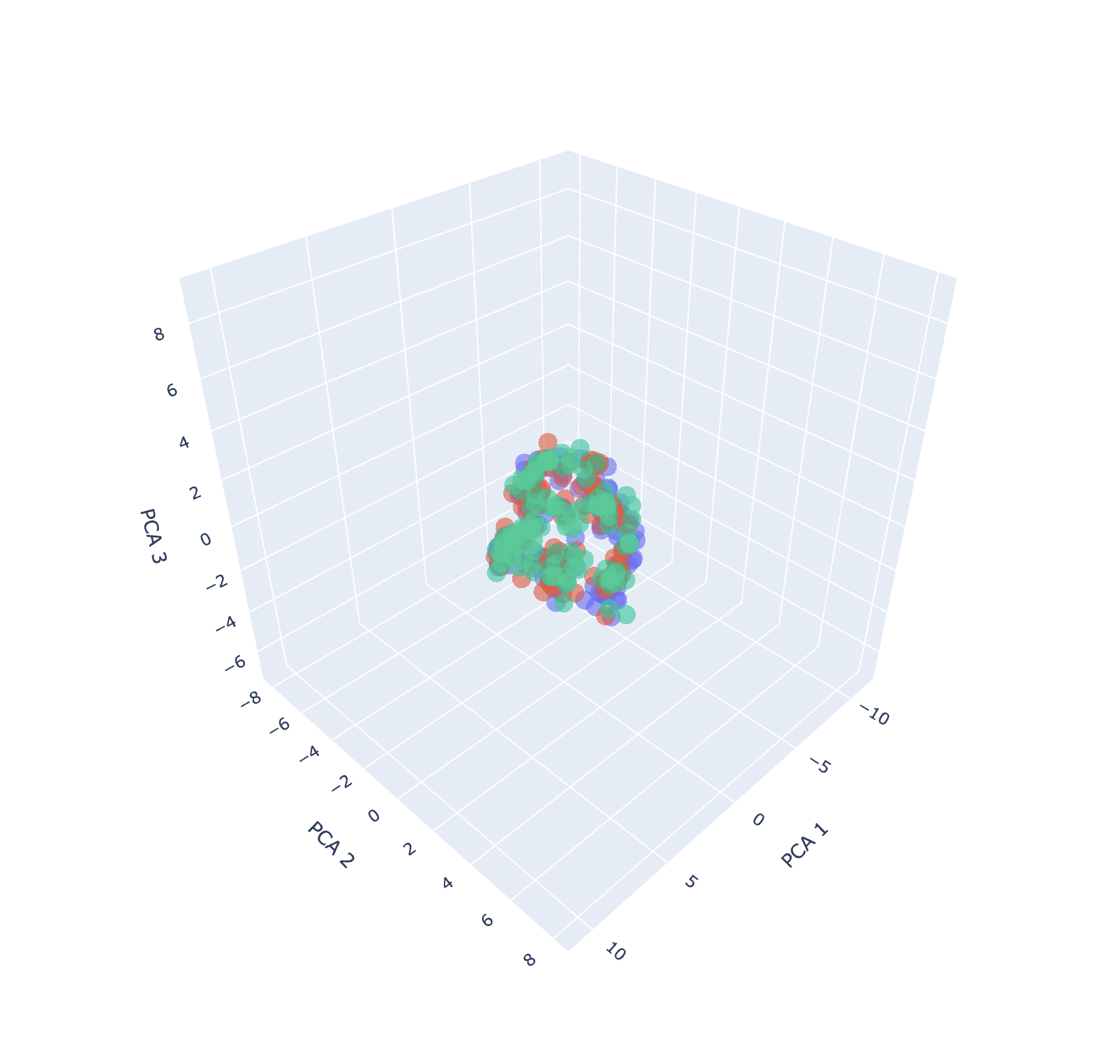}
  \caption{Layer 1}
\end{subfigure}\hfill
\begin{subfigure}{0.3\textwidth}
  \centering
  \includegraphics[width=\linewidth]{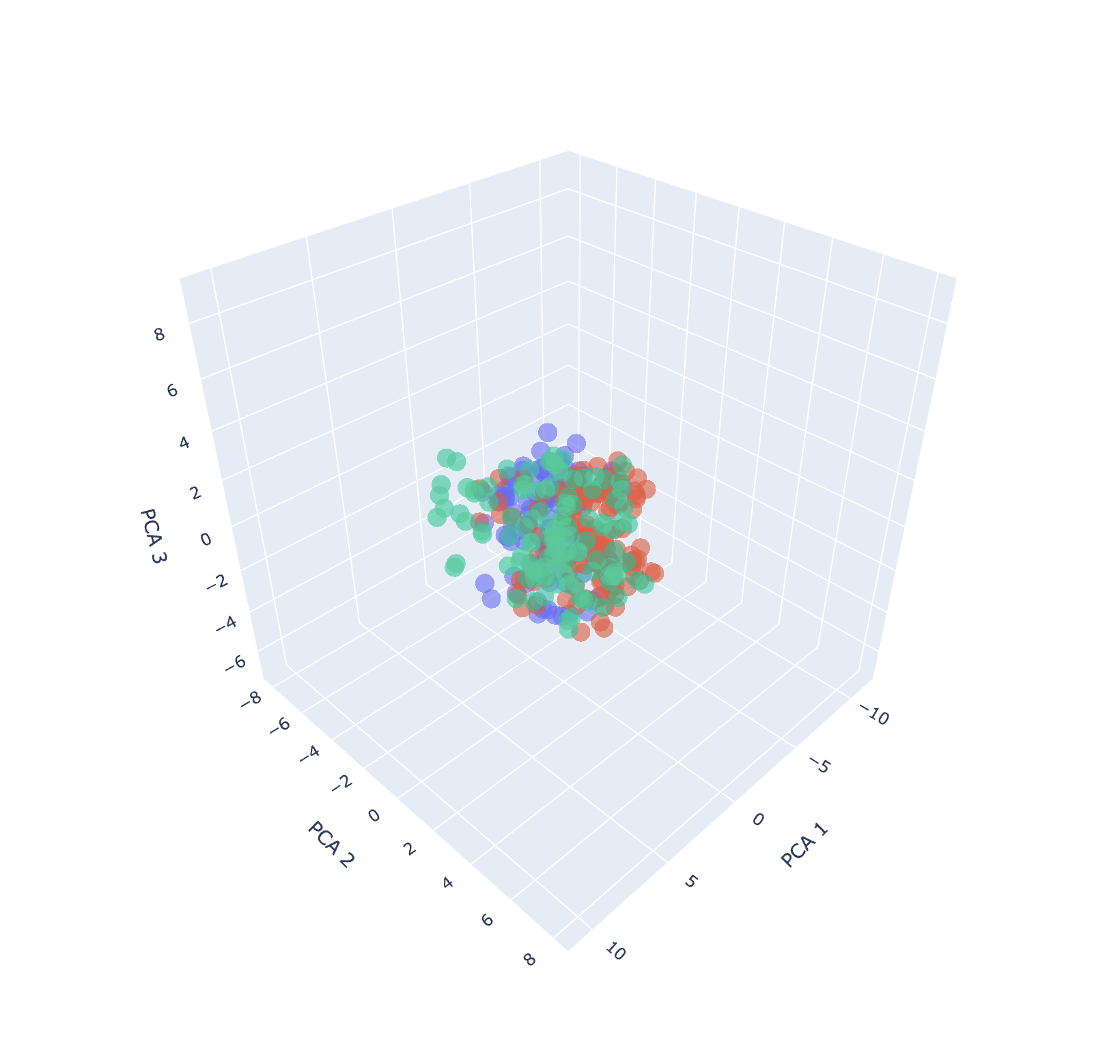}
  \caption{Layer 2}
\end{subfigure}\hfill
\begin{subfigure}{0.3\textwidth}
  \centering
  \includegraphics[width=\linewidth]{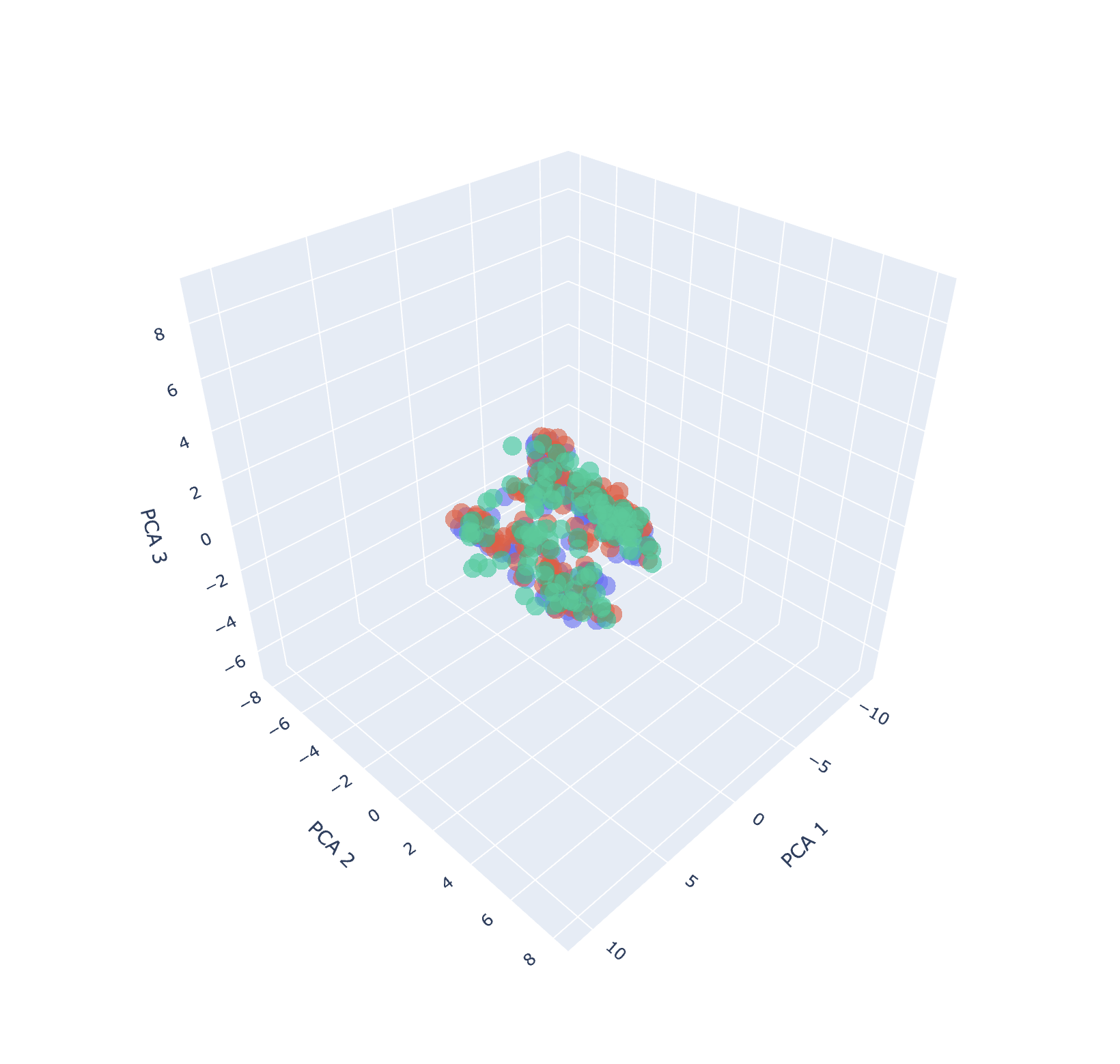}
  \caption{Layer 3}
\end{subfigure}

\begin{subfigure}{0.3\textwidth}
  \centering
  \includegraphics[width=\linewidth]{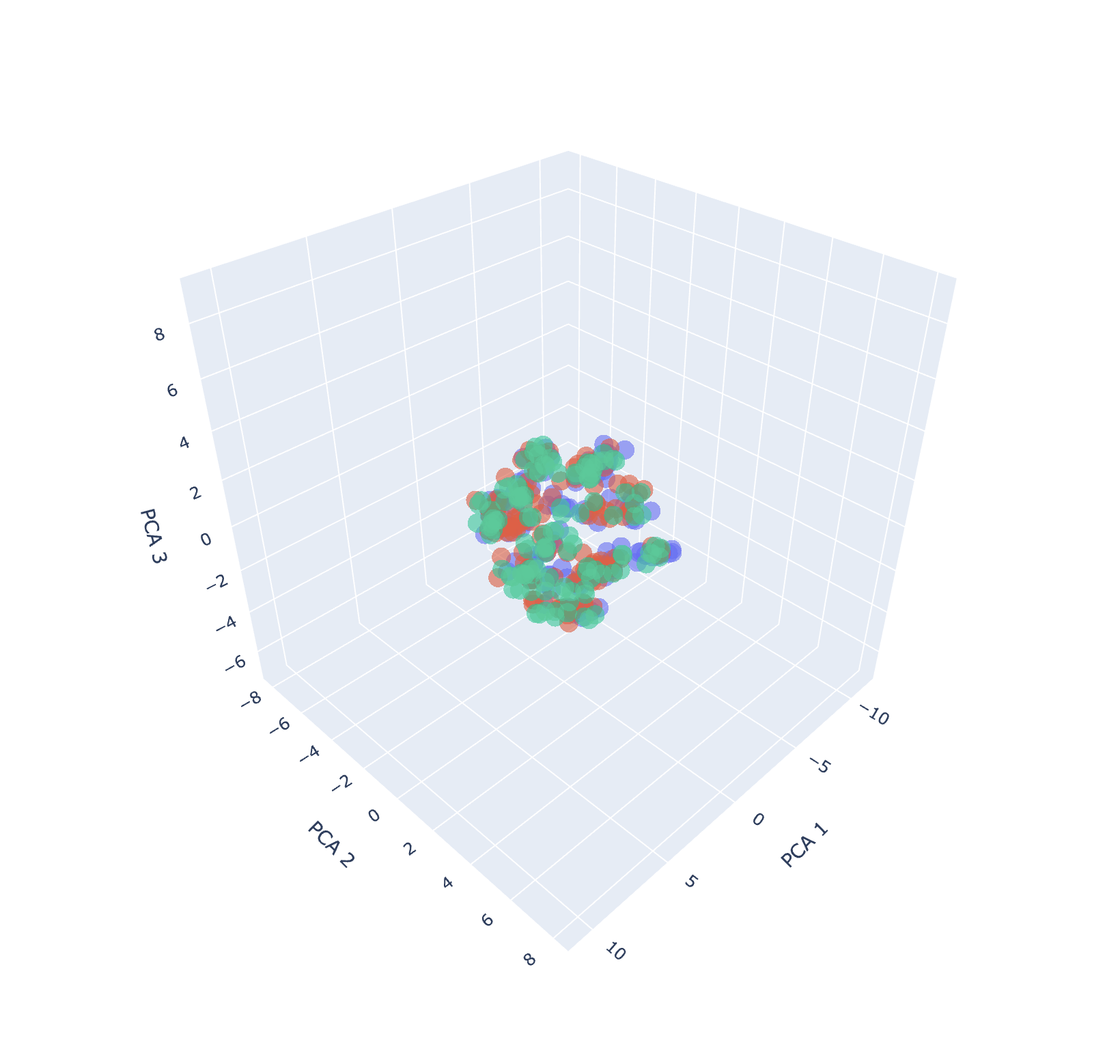}
  \caption{Layer 4}
\end{subfigure}\hfill
\begin{subfigure}{0.3\textwidth}
  \centering
  \includegraphics[width=\linewidth]{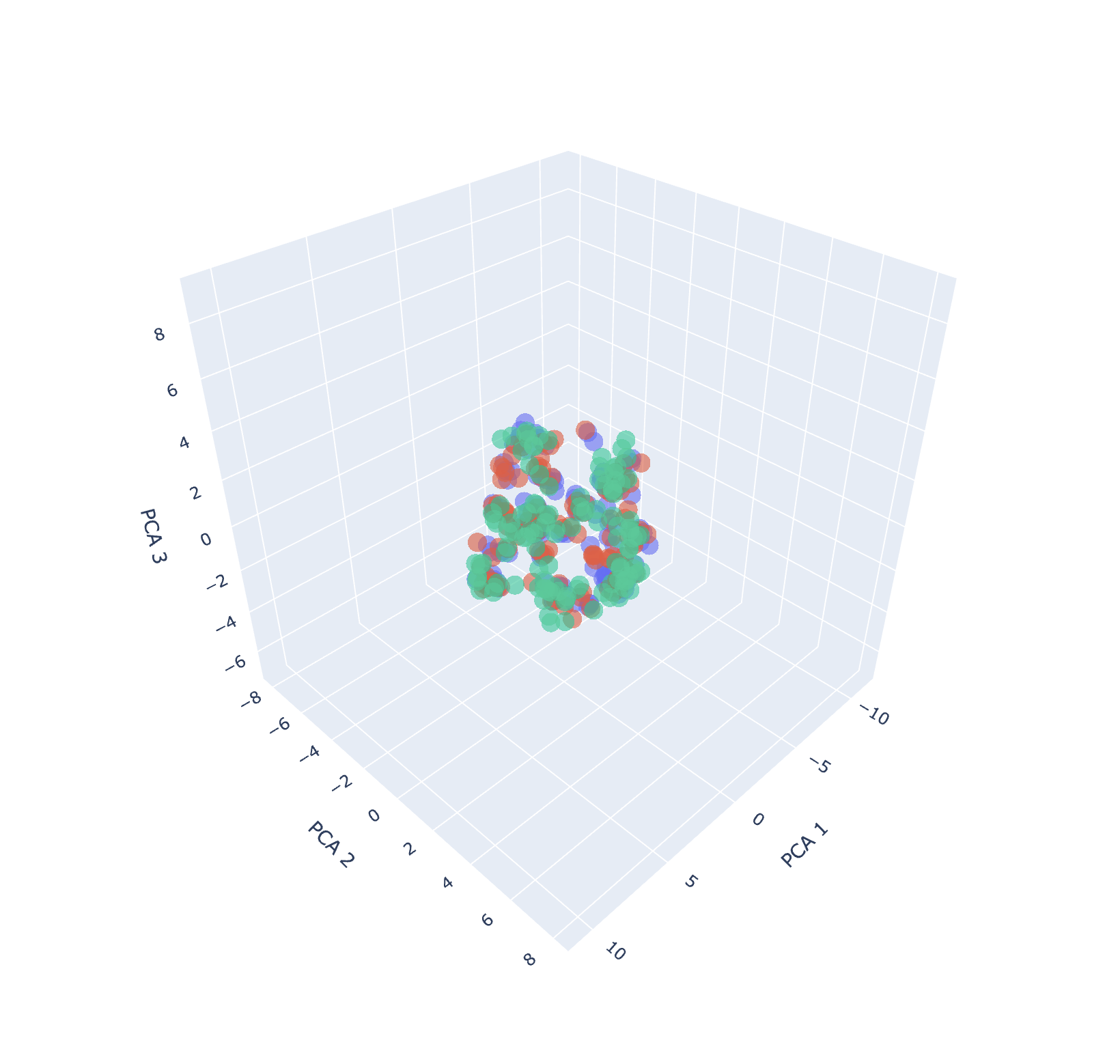}
  \caption{Layer 5}
\end{subfigure}\hfill
\begin{subfigure}{0.3\textwidth}
  \centering
  \includegraphics[width=\linewidth]{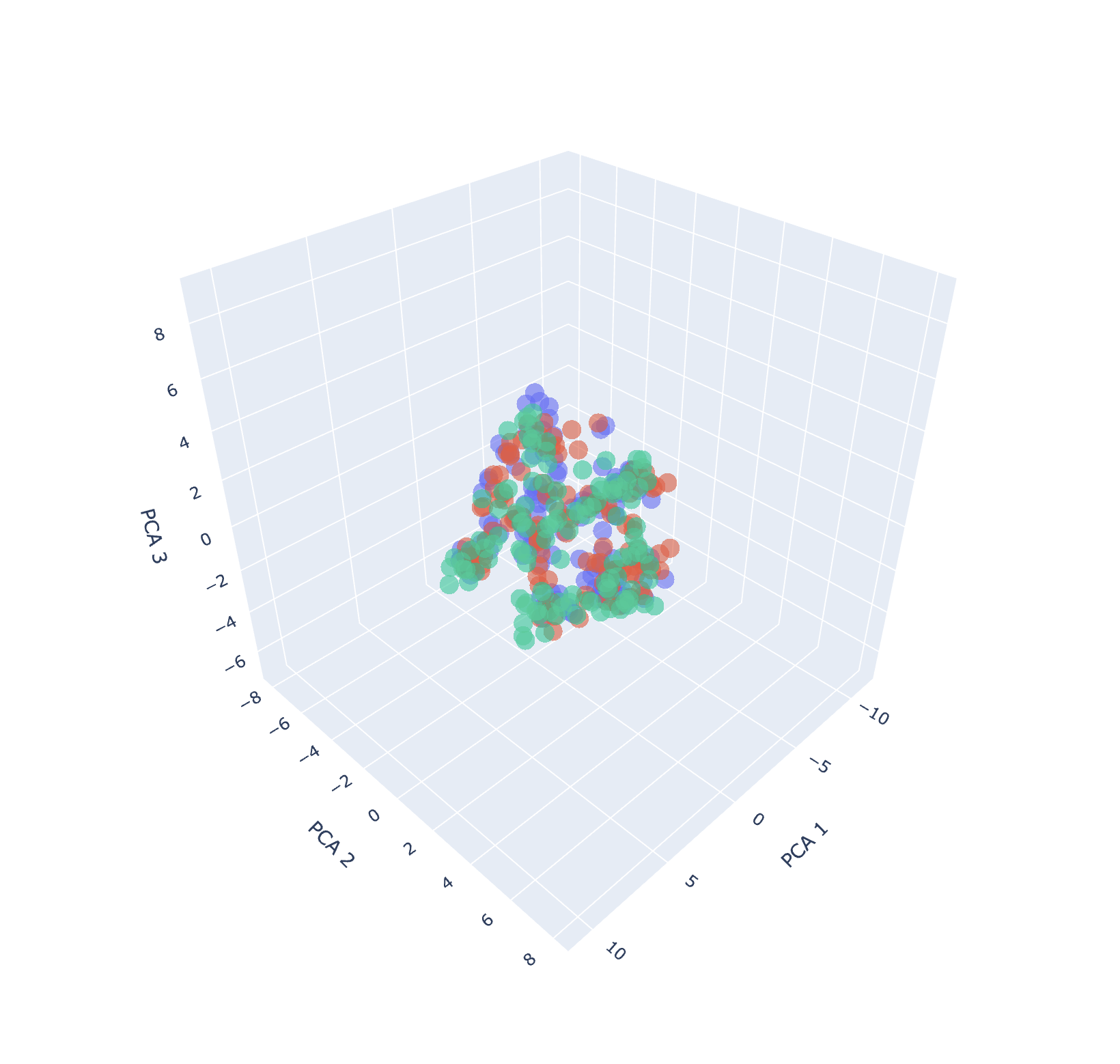}
  \caption{Layer 6}
\end{subfigure}

\begin{subfigure}{0.3\textwidth}
  \centering
  \includegraphics[width=\linewidth]{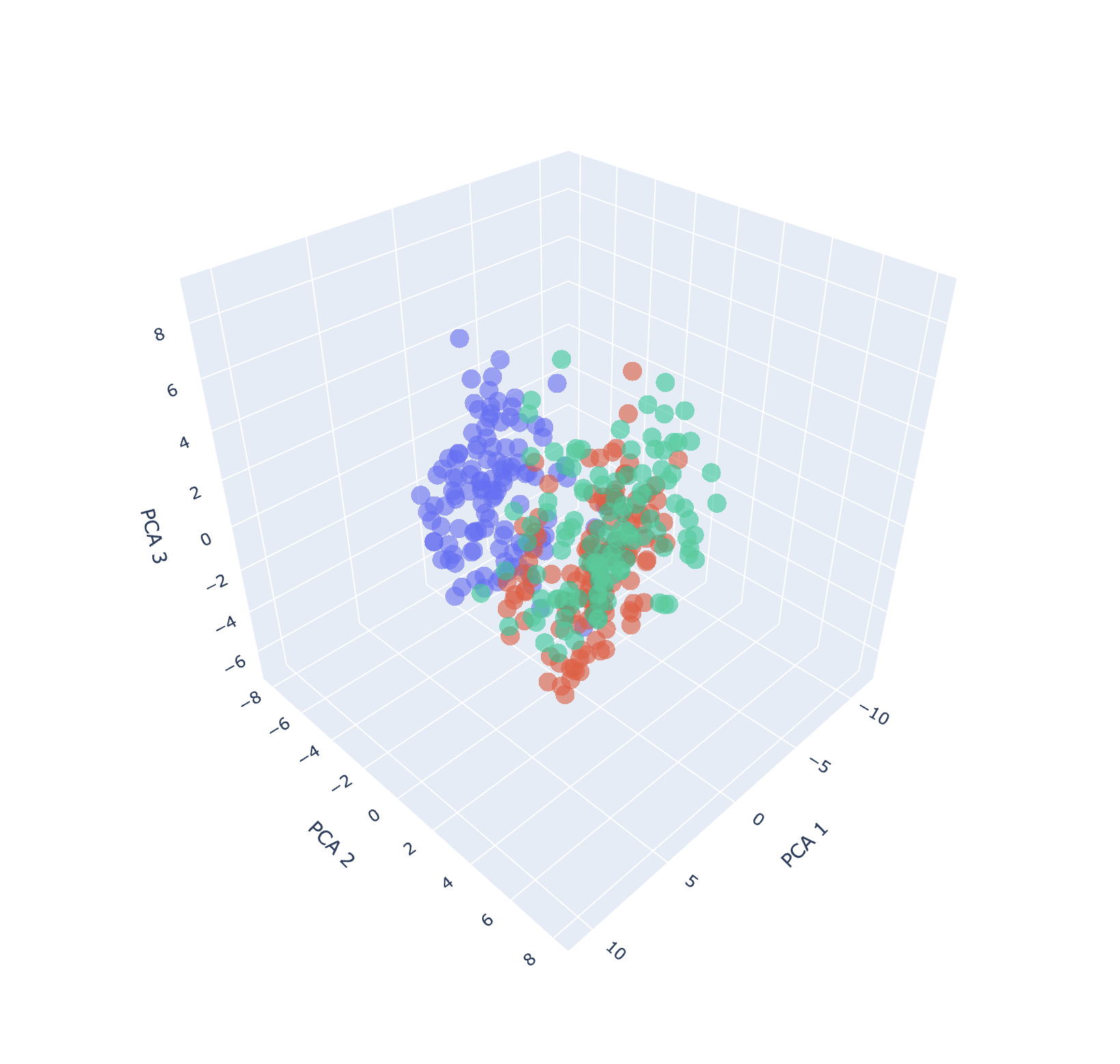}
  \caption{Layer 7}
\end{subfigure}\hfill
\begin{subfigure}{0.3\textwidth}
  \centering
  \includegraphics[width=\linewidth]{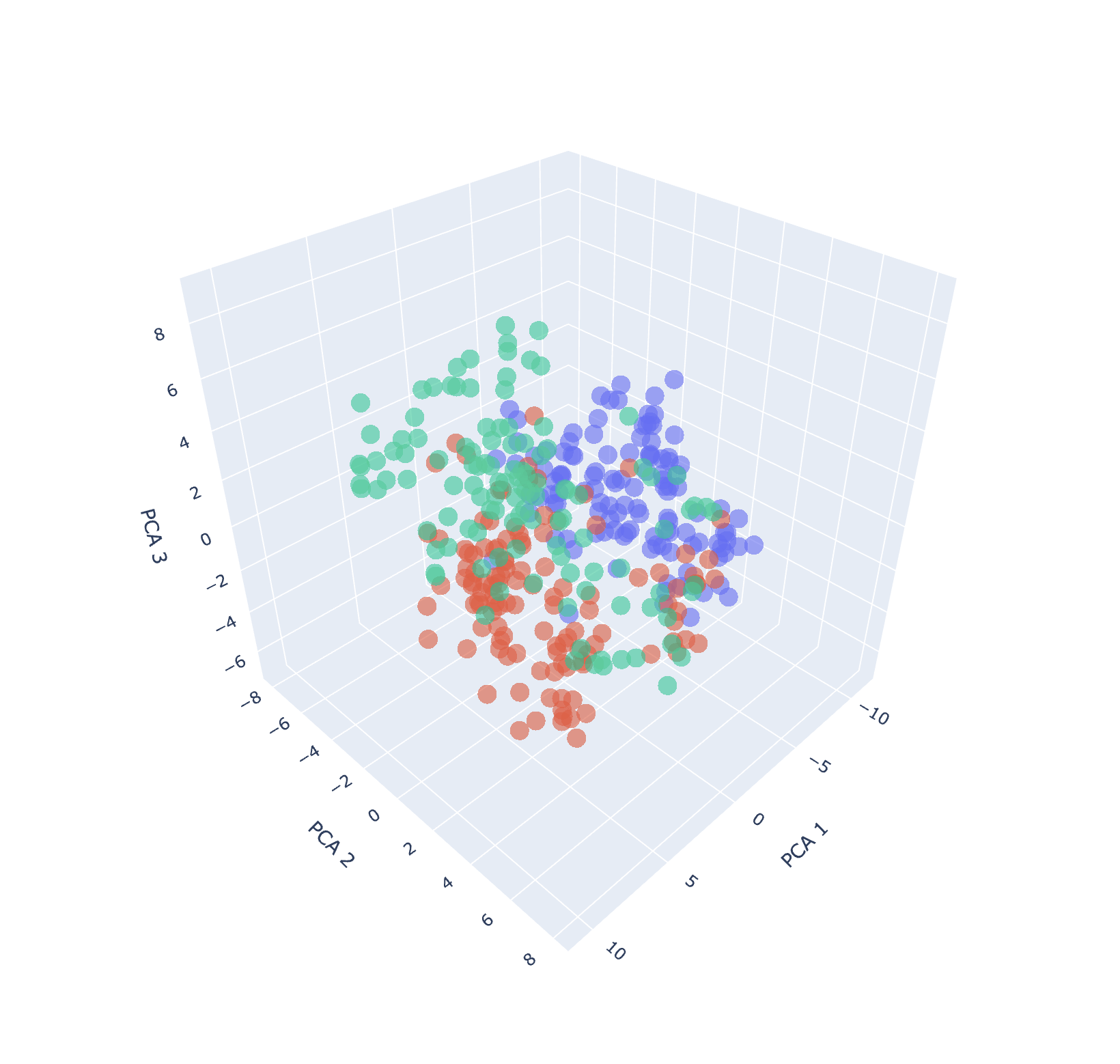}
  \caption{Layer 8}
\end{subfigure}\hfill
\begin{subfigure}{0.3\textwidth}
  \centering
  \includegraphics[width=\linewidth]{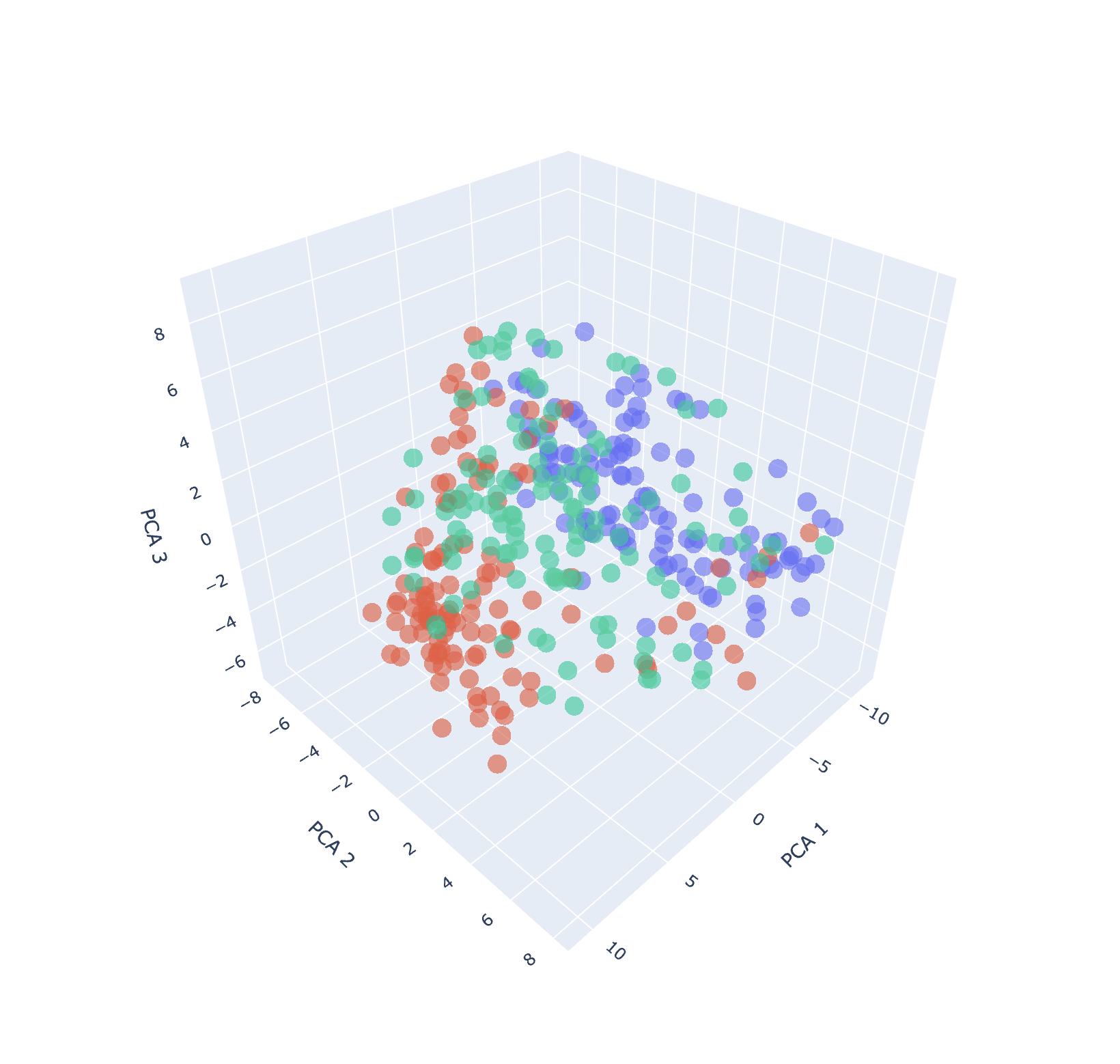}
  \caption{Layer 9}
\end{subfigure}

\begin{subfigure}{0.3\textwidth}
  \centering
  \includegraphics[width=\linewidth]{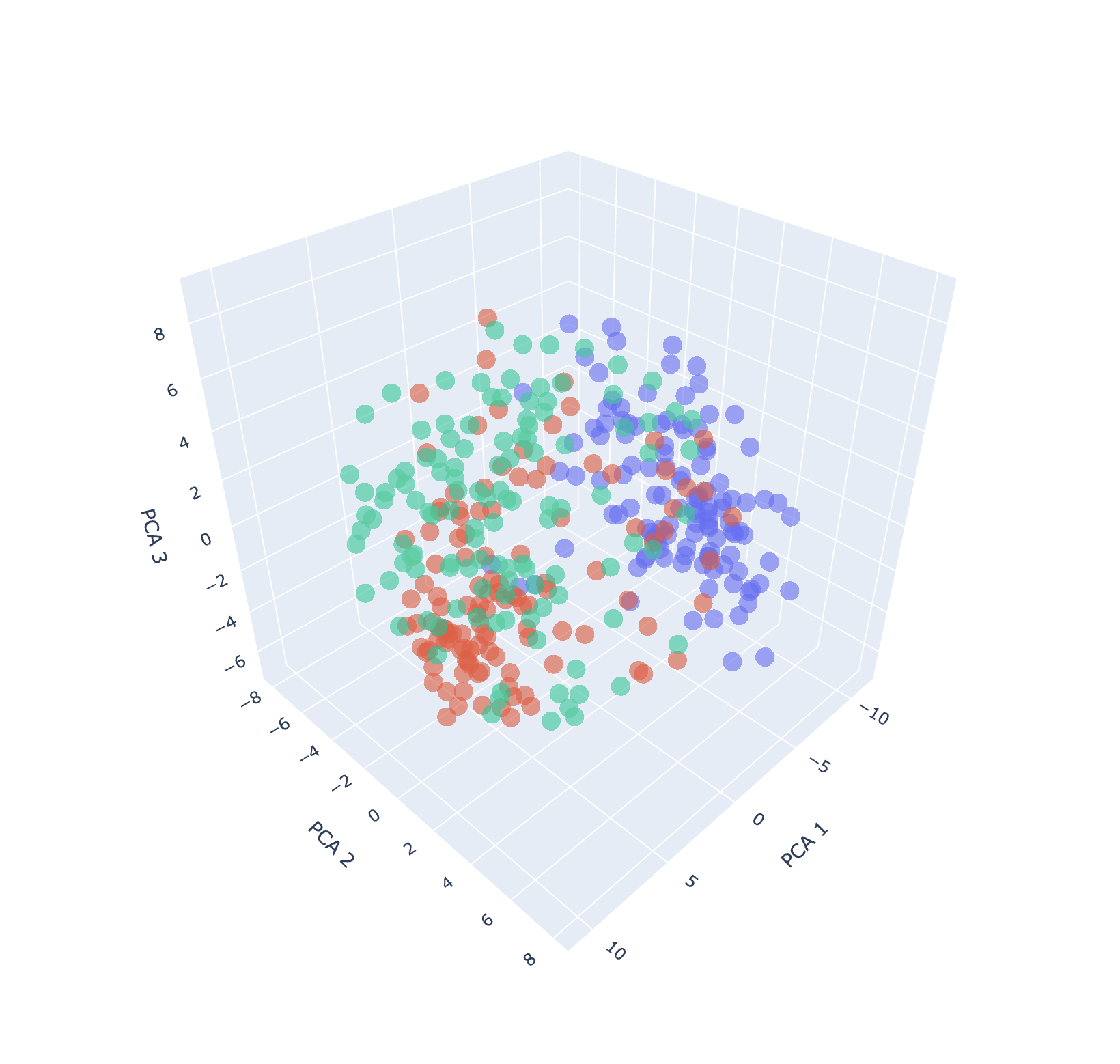}
  \caption{Layer 10}
\end{subfigure}\hfill
\begin{subfigure}{0.3\textwidth}
  \centering
  \includegraphics[width=\linewidth]{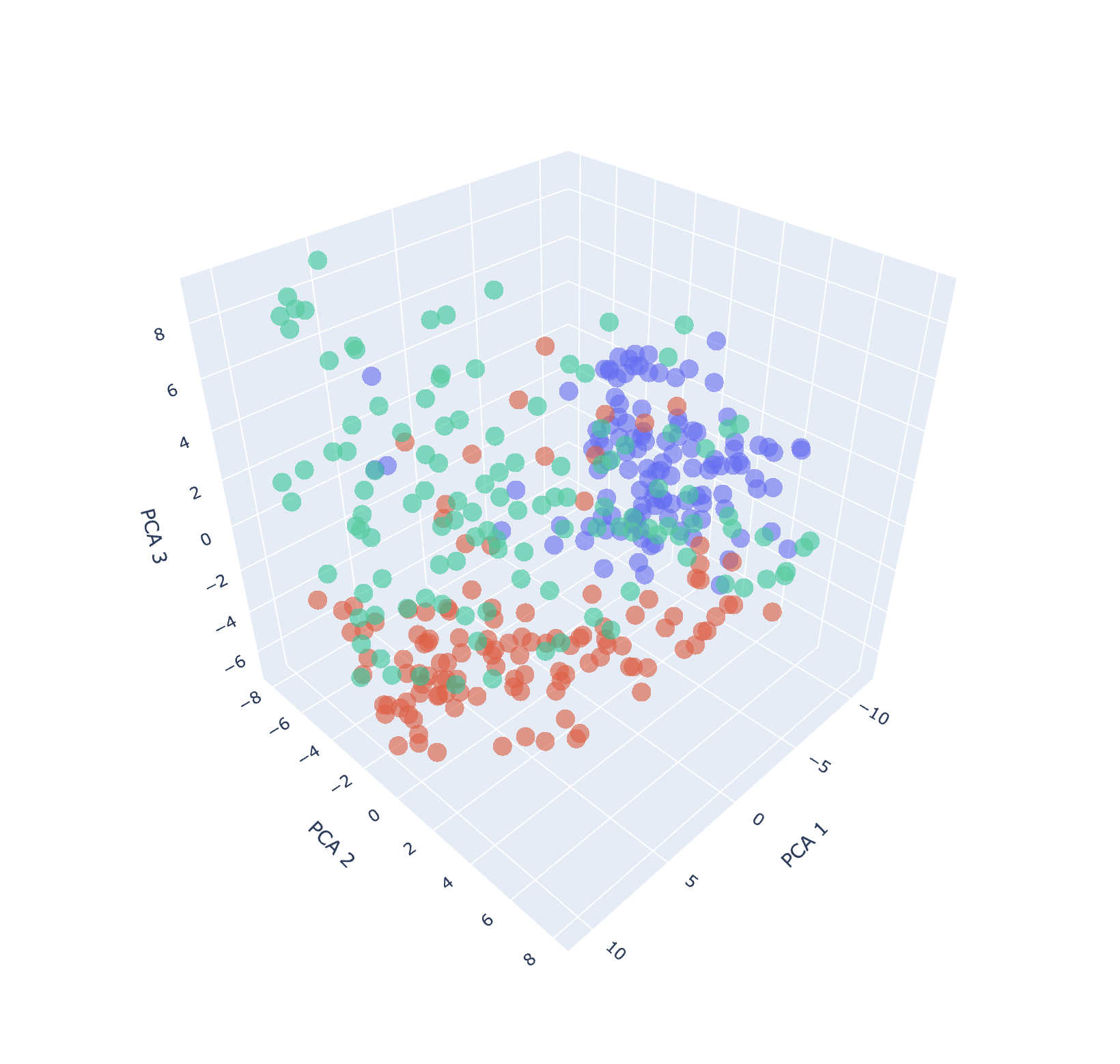}
  \caption{Layer 11}
\end{subfigure}\hfill
\begin{subfigure}{0.3\textwidth}
  \centering
  \includegraphics[width=\linewidth]{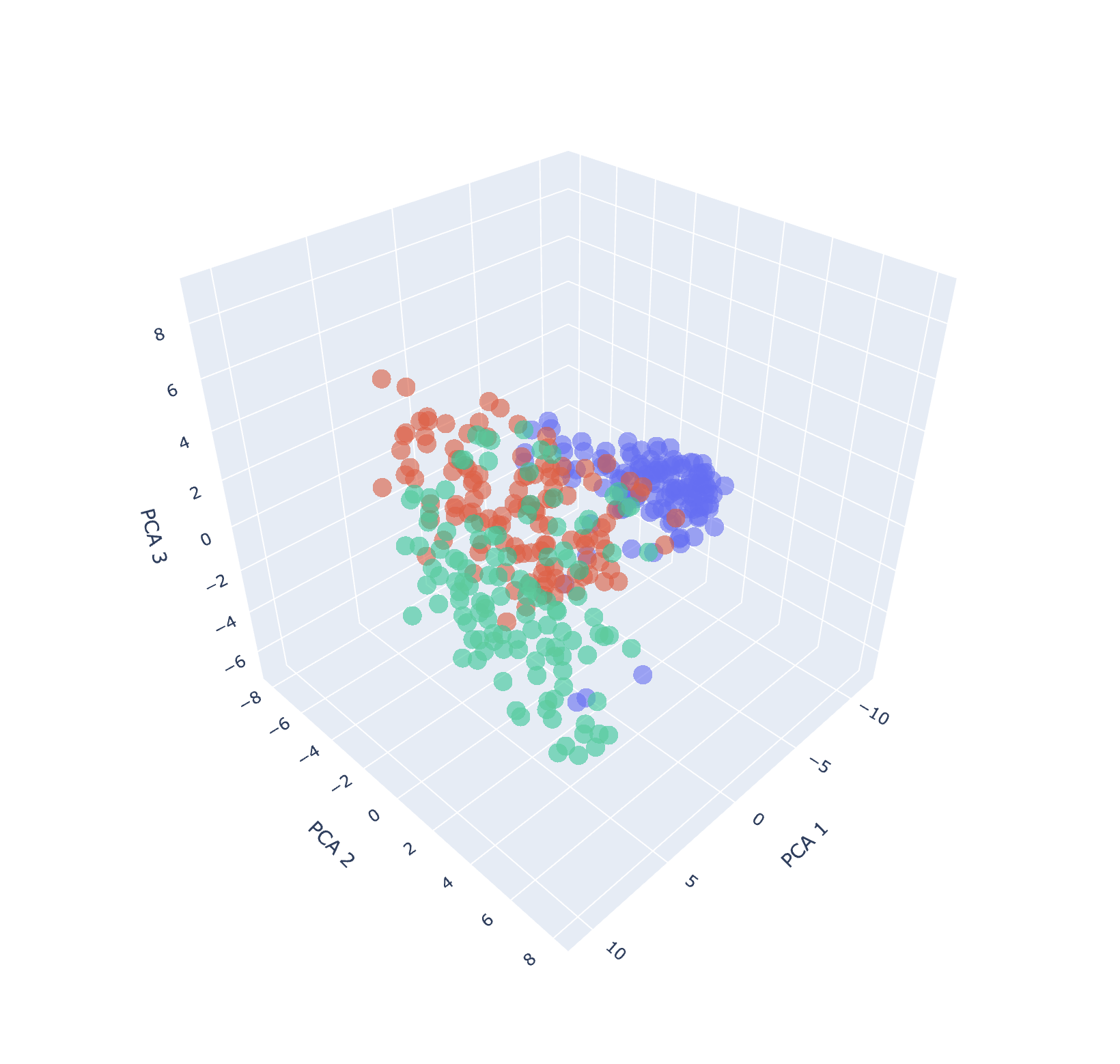}
  \caption{Layer 12}
\end{subfigure}

\caption{Three-dimensional PCA projection of UNK-based BERT contextual embeddings across layers for \npn semi-schematic constructions. Data points are colour-coded by semantic value: \textit{juxtaposition/contact} (red), \textit{greater\_plurality/accumulation} (red), and \textit{succession/iteration/distributivity} (green). The visualisation is based on a balanced subset of 360 instances. An animated version of the visualisation is available at \url{https://gretagorzoni00.github.io/NPN_contextual_embeddings/}.}
\label{fig:pcaex2UNK}
\end{figure}

\begin{figure}[h]

\centering

\begin{subfigure}{0.3\textwidth}
  \centering
  \includegraphics[width=\linewidth]{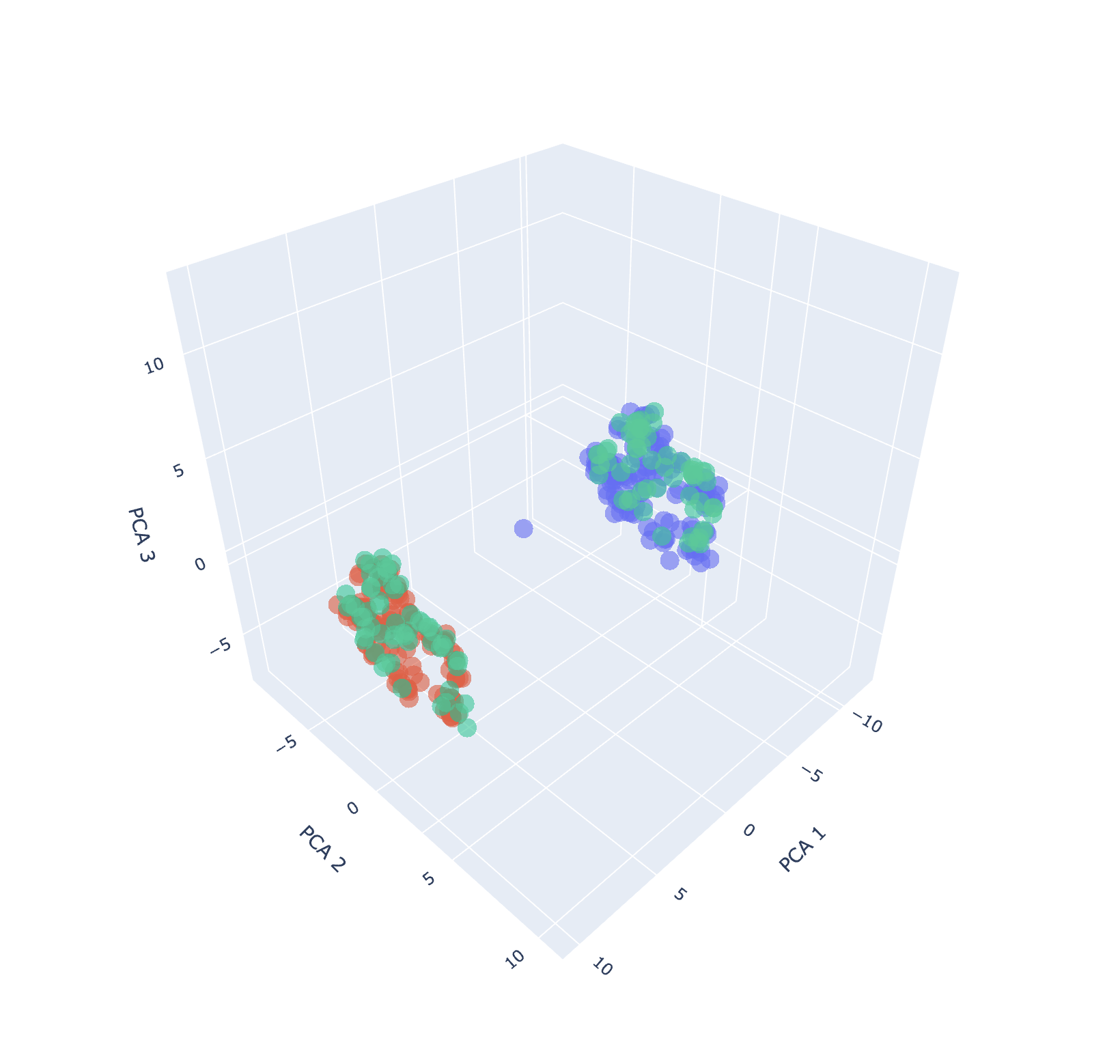}
  \caption{Layer 1}
\end{subfigure}\hfill
\begin{subfigure}{0.3\textwidth}
  \centering
  \includegraphics[width=\linewidth]{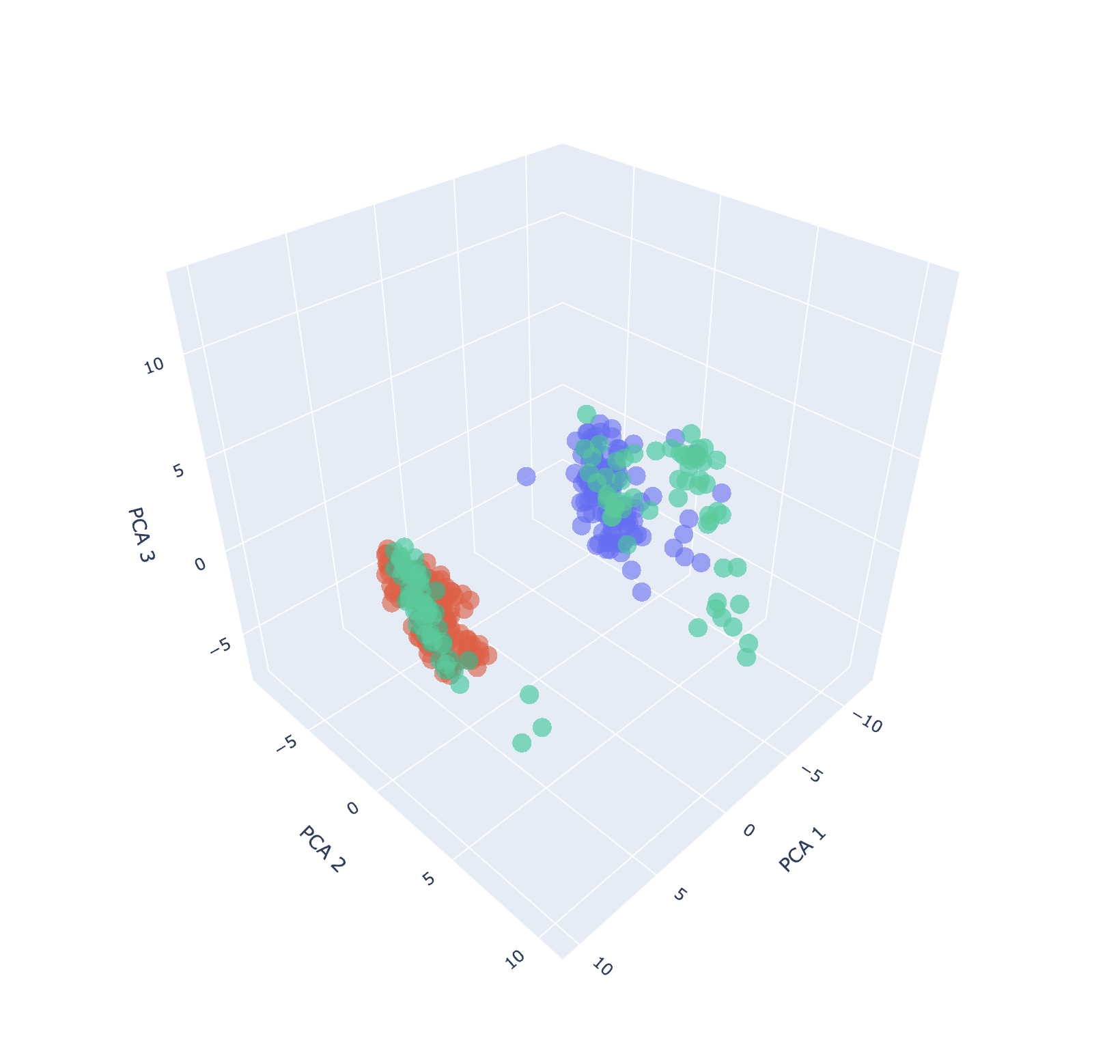}
  \caption{Layer 2}
\end{subfigure}\hfill
\begin{subfigure}{0.3\textwidth}
  \centering
  \includegraphics[width=\linewidth]{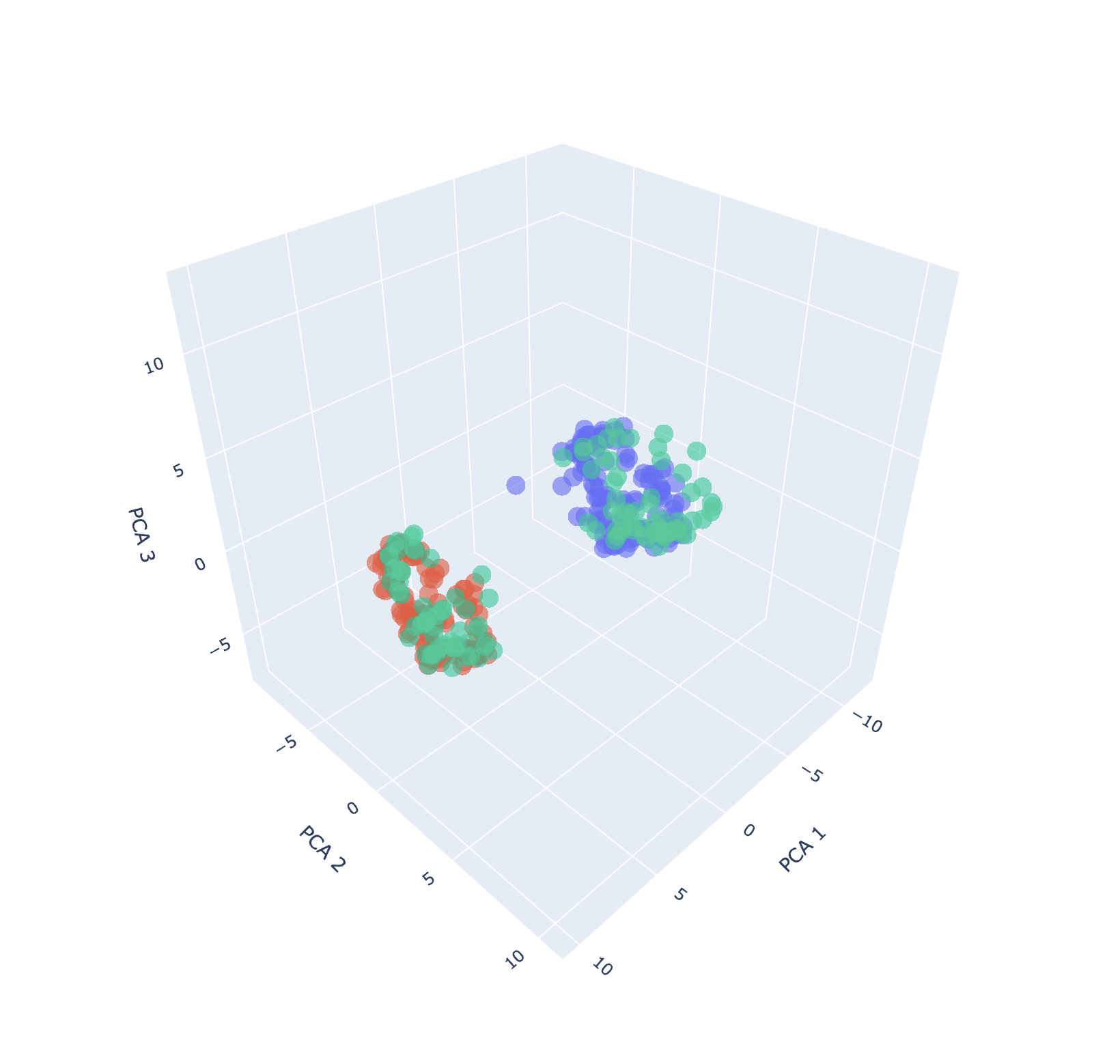}
  \caption{Layer 3}
\end{subfigure}

\begin{subfigure}{0.3\textwidth}
  \centering
  \includegraphics[width=\linewidth]{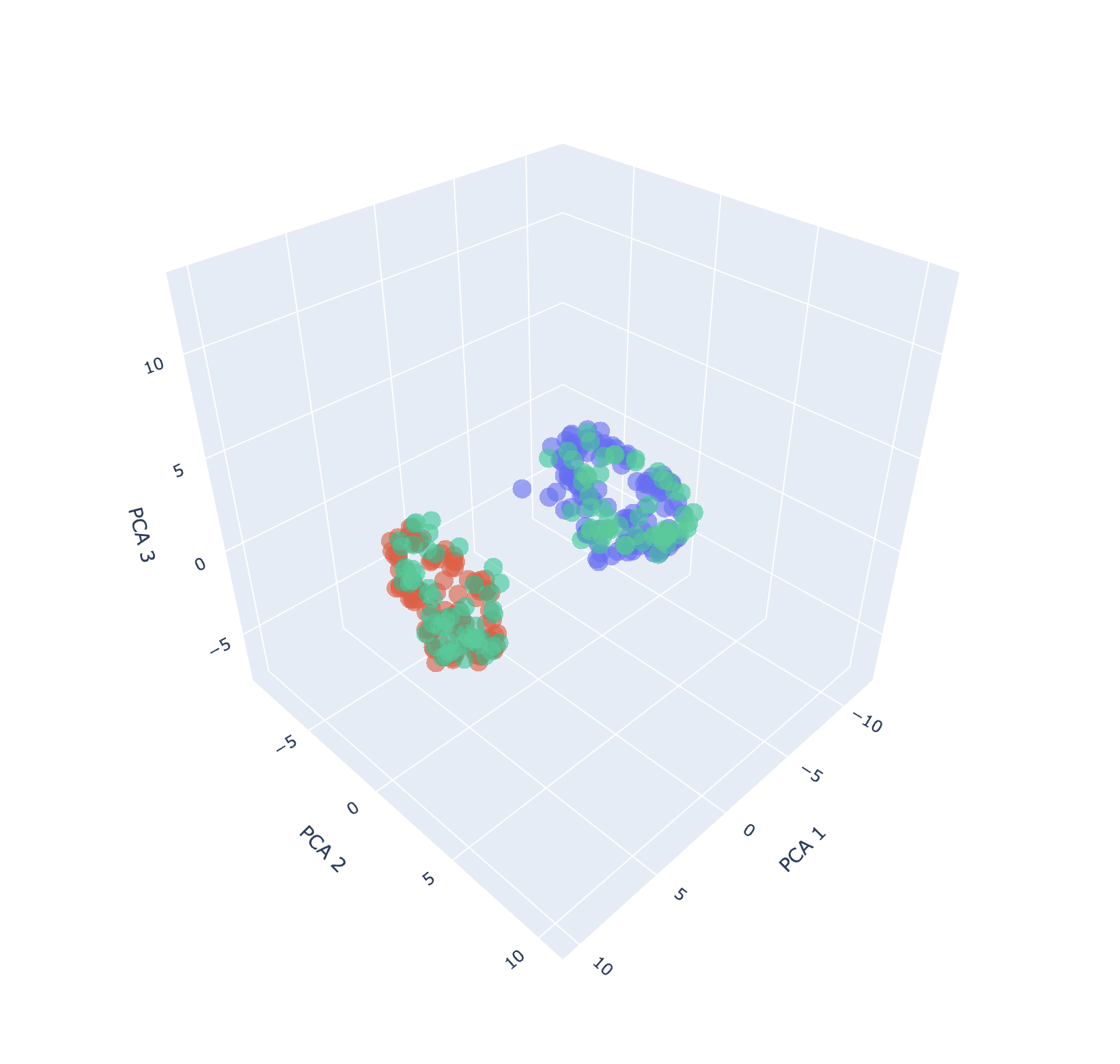}
  \caption{Layer 4}
\end{subfigure}\hfill
\begin{subfigure}{0.3\textwidth}
  \centering
  \includegraphics[width=\linewidth]{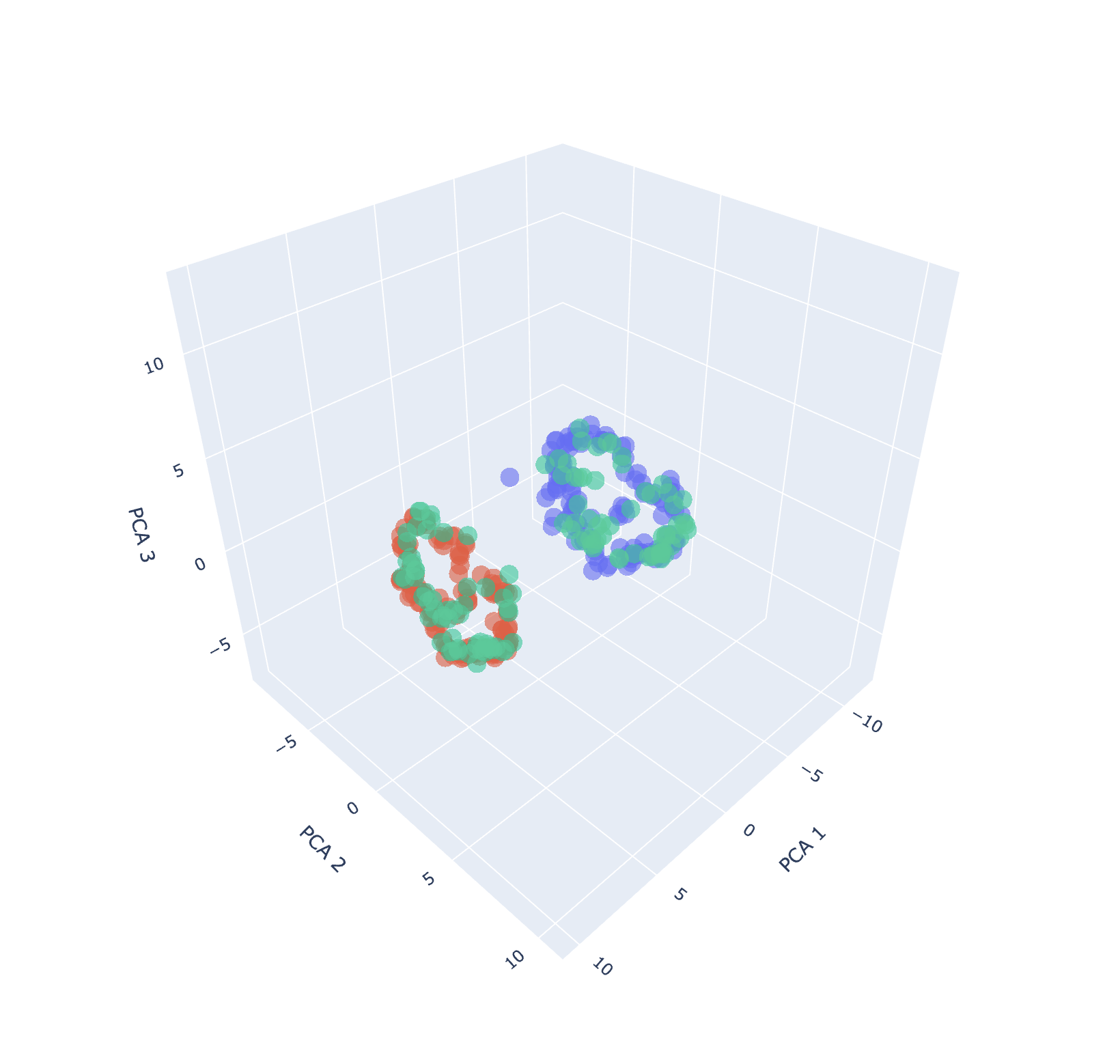}
  \caption{Layer 5}
\end{subfigure}\hfill
\begin{subfigure}{0.3\textwidth}
  \centering
  \includegraphics[width=\linewidth]{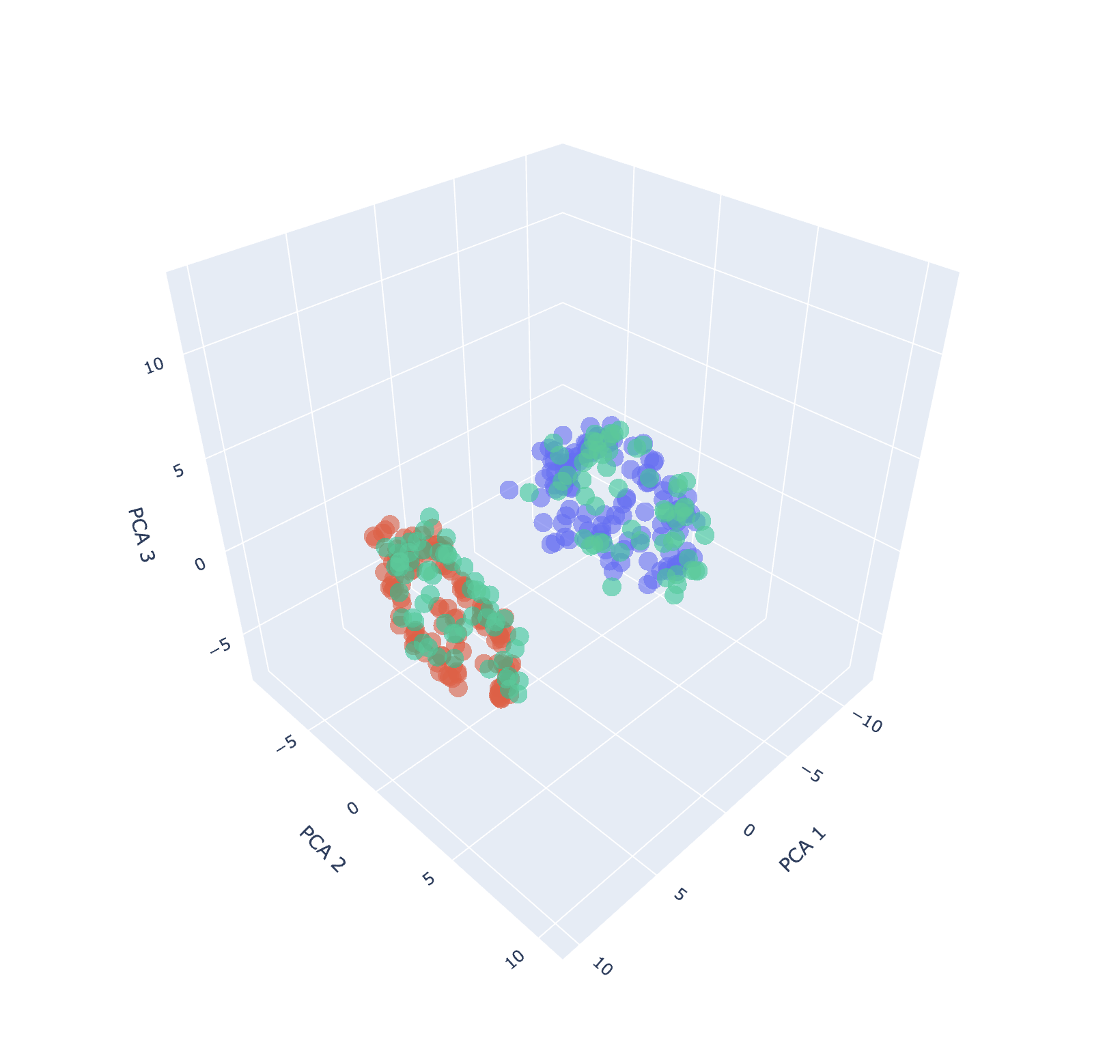}
  \caption{Layer 6}
\end{subfigure}

\begin{subfigure}{0.3\textwidth}
  \centering
  \includegraphics[width=\linewidth]{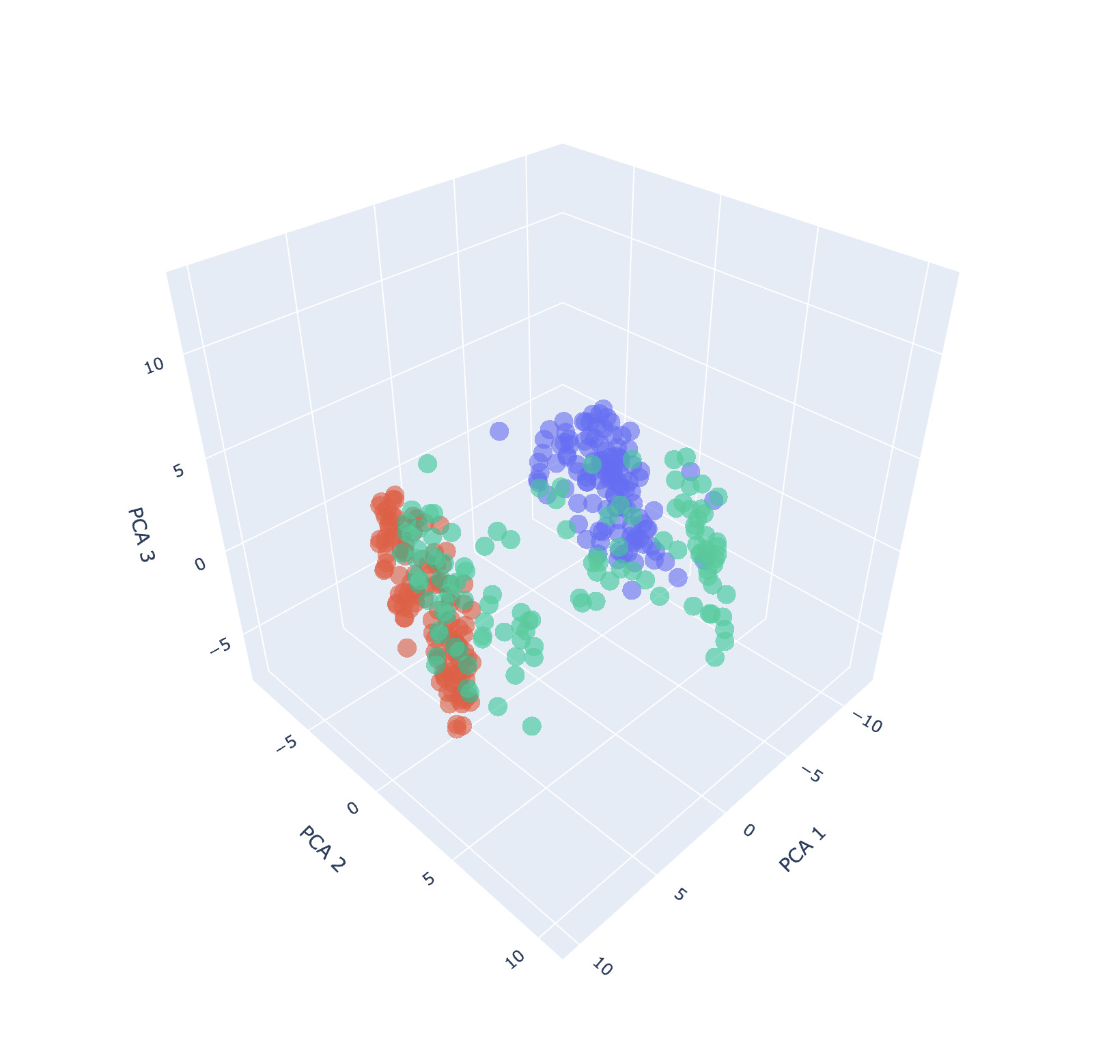}
  \caption{Layer 7}
\end{subfigure}\hfill
\begin{subfigure}{0.3\textwidth}
  \centering
  \includegraphics[width=\linewidth]{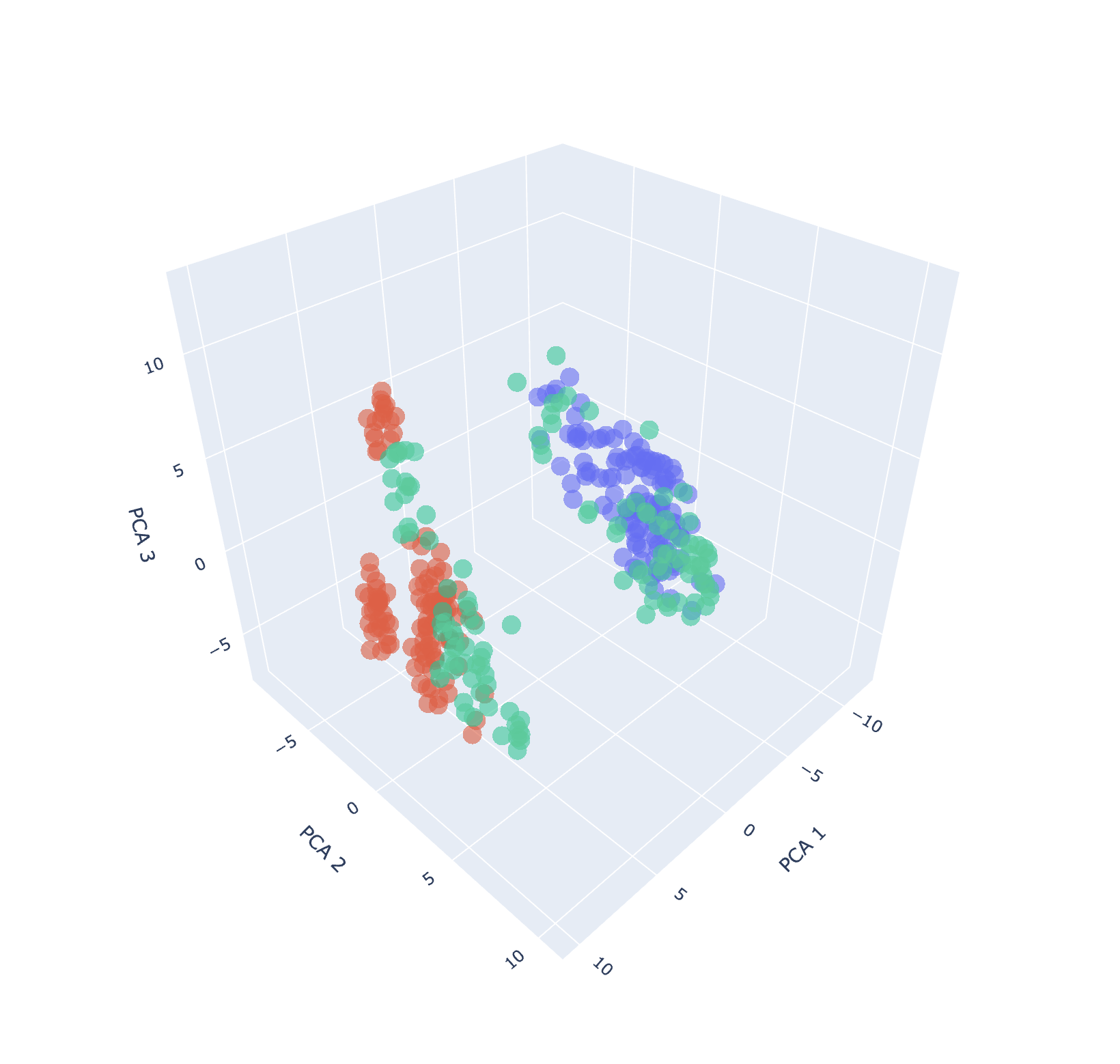}
  \caption{Layer 8}
\end{subfigure}\hfill
\begin{subfigure}{0.3\textwidth}
  \centering
  \includegraphics[width=\linewidth]{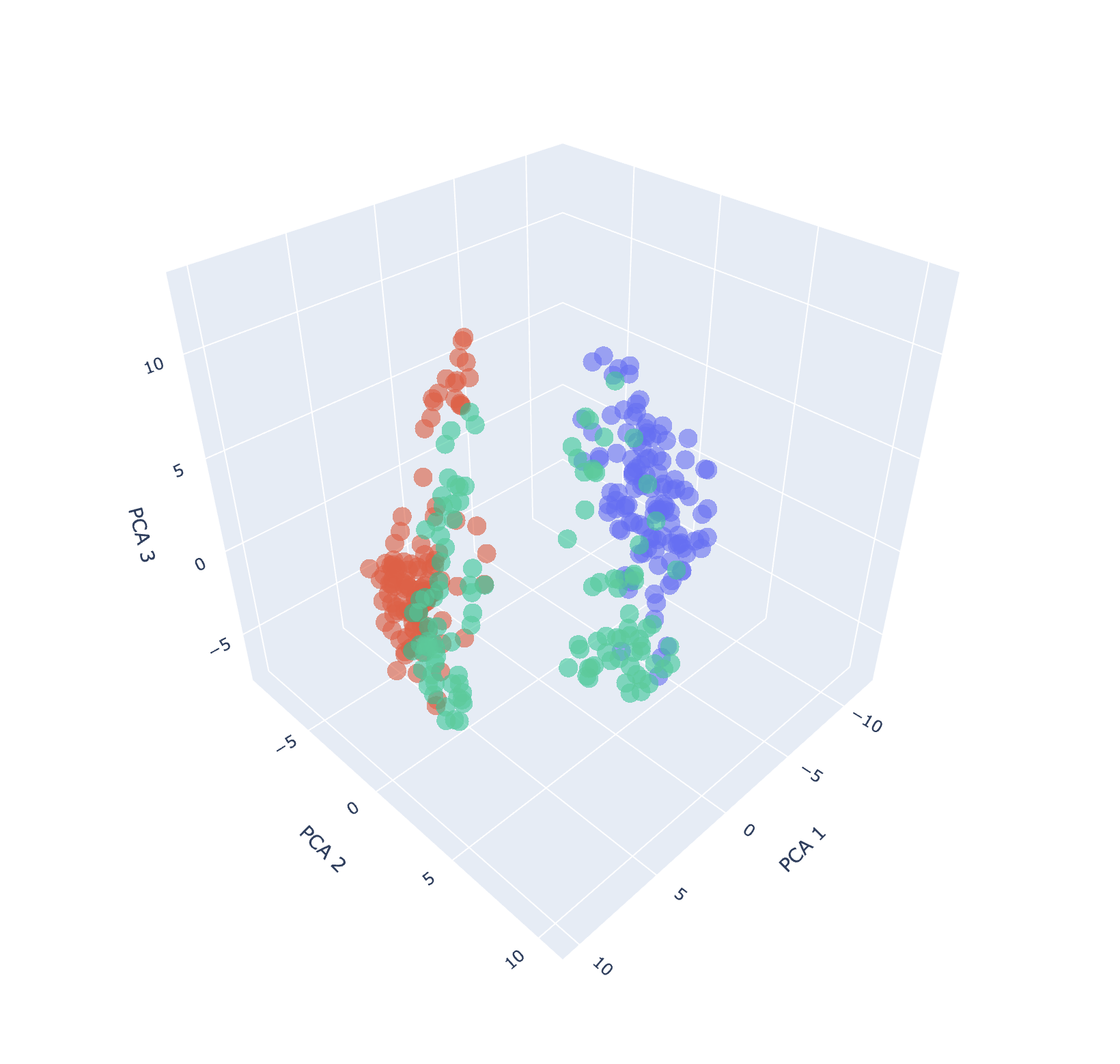}
  \caption{Layer 9}
\end{subfigure}

\begin{subfigure}{0.3\textwidth}
  \centering
  \includegraphics[width=\linewidth]{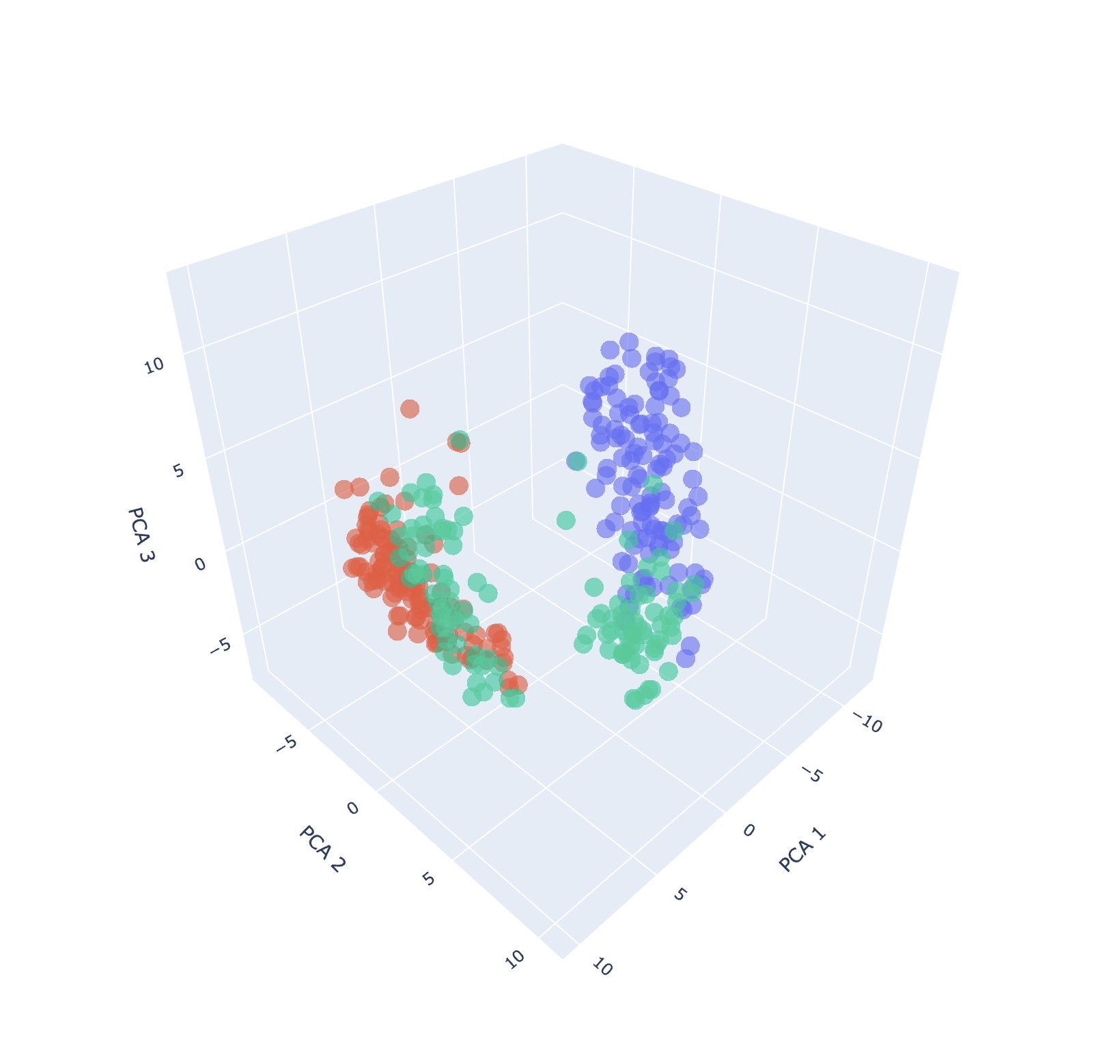}
  \caption{Layer 10}
\end{subfigure}\hfill
\begin{subfigure}{0.3\textwidth}
  \centering
  \includegraphics[width=\linewidth]{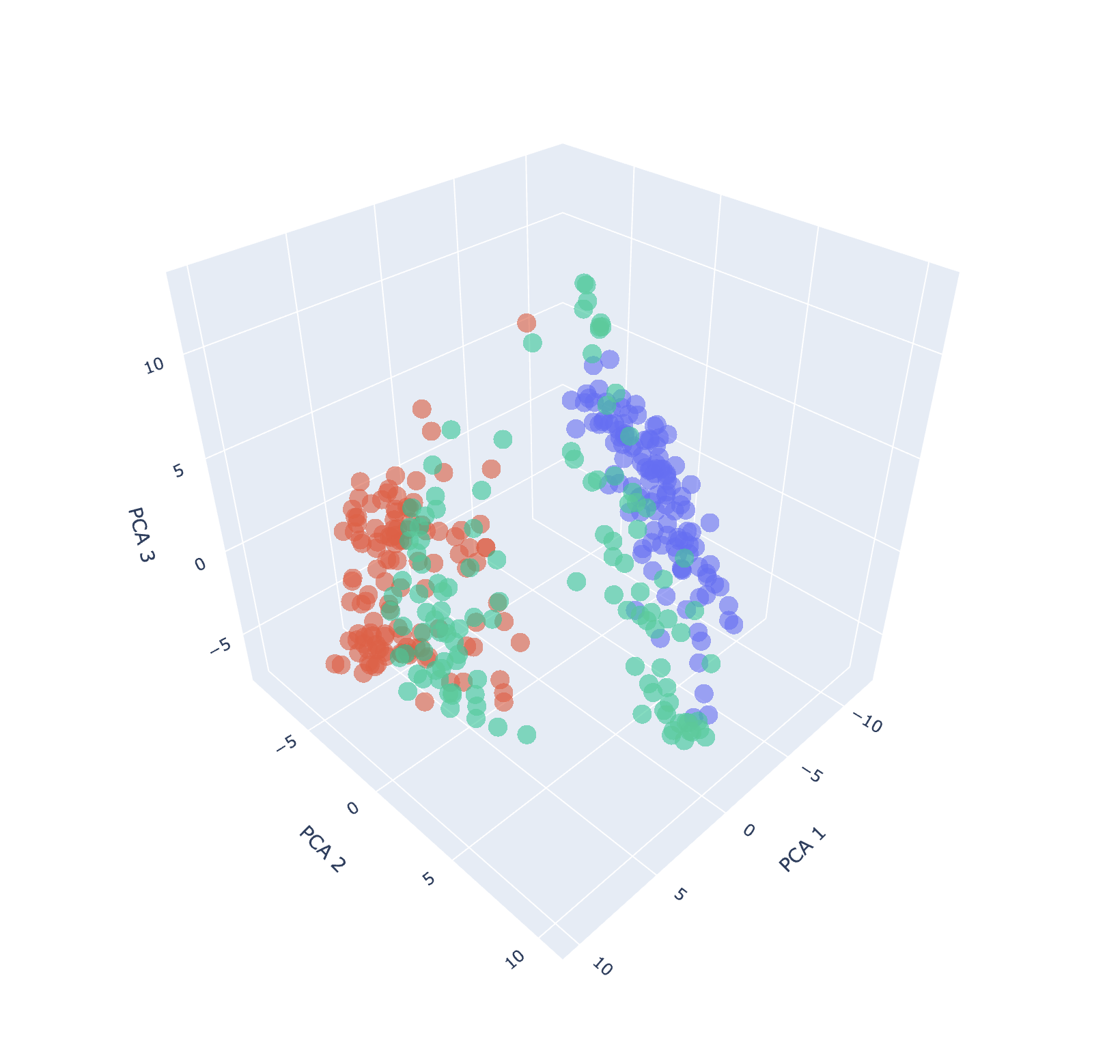}
  \caption{Layer 11}
\end{subfigure}\hfill
\begin{subfigure}{0.3\textwidth}
  \centering
  \includegraphics[width=\linewidth]{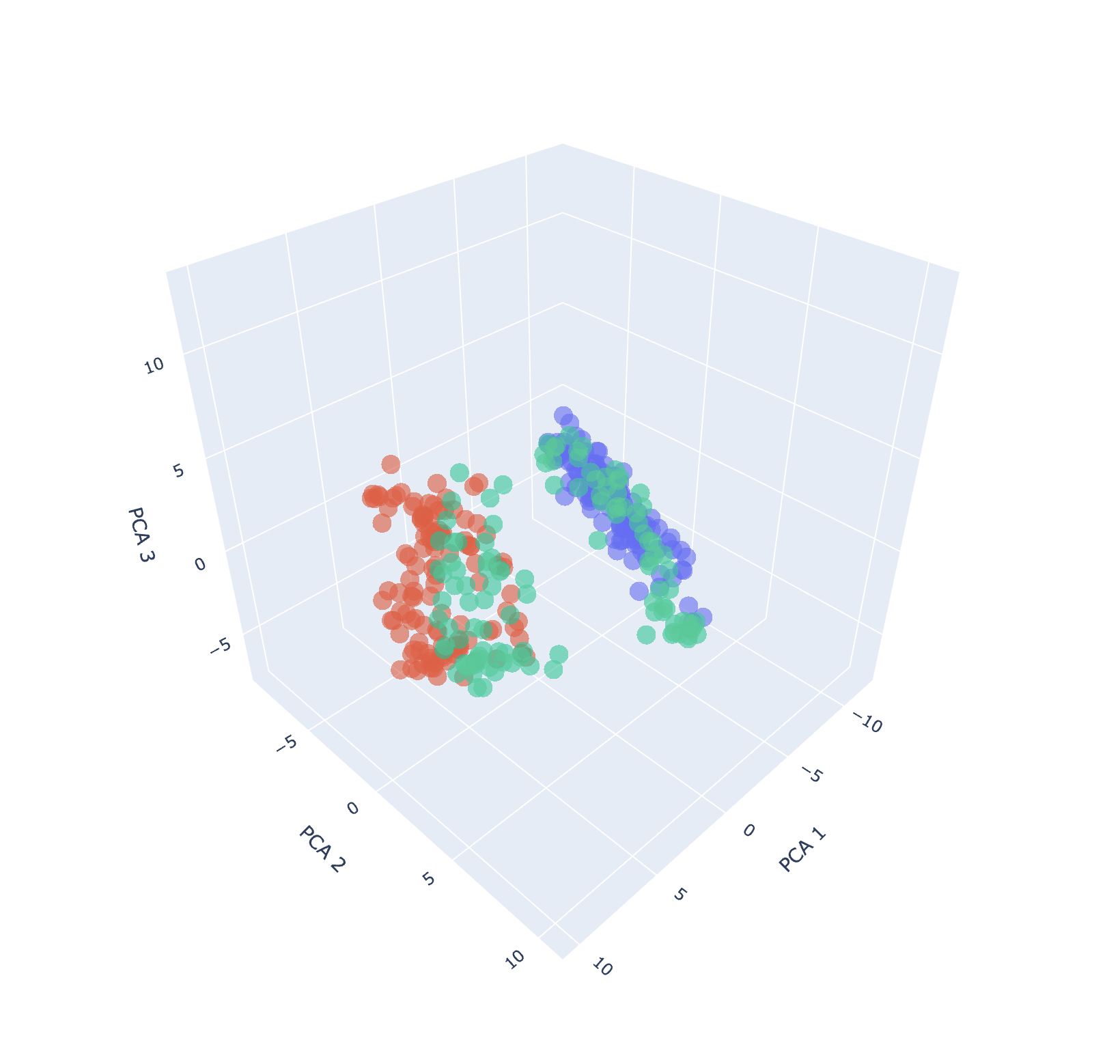}
  \caption{Layer 12}
\end{subfigure}

\caption{Three-dimensional PCA projection of PREP-based BERT contextual embeddings across layers for \npn semi-schematic Constructions. Data points are color-coded by semantic value: \textit{juxtaposition/contact} (red), \textit{greater\_plurality/accumulation} (red), and \textit{succession/iteration/distributivity} (green). The visualisation is based on a balanced subset of 360 instances. An animated version of the visualisation is available at \url{https://gretagorzoni00.github.io/NPN_contextual_embeddings/}.}

\label{fig:PCAex2PREP}
\end{figure}
\clearpage

\subsection{Confusion matrices}
\label{Confusion matrices}

\begin{figure}[h]
    \centering
    \includegraphics[width=\linewidth]{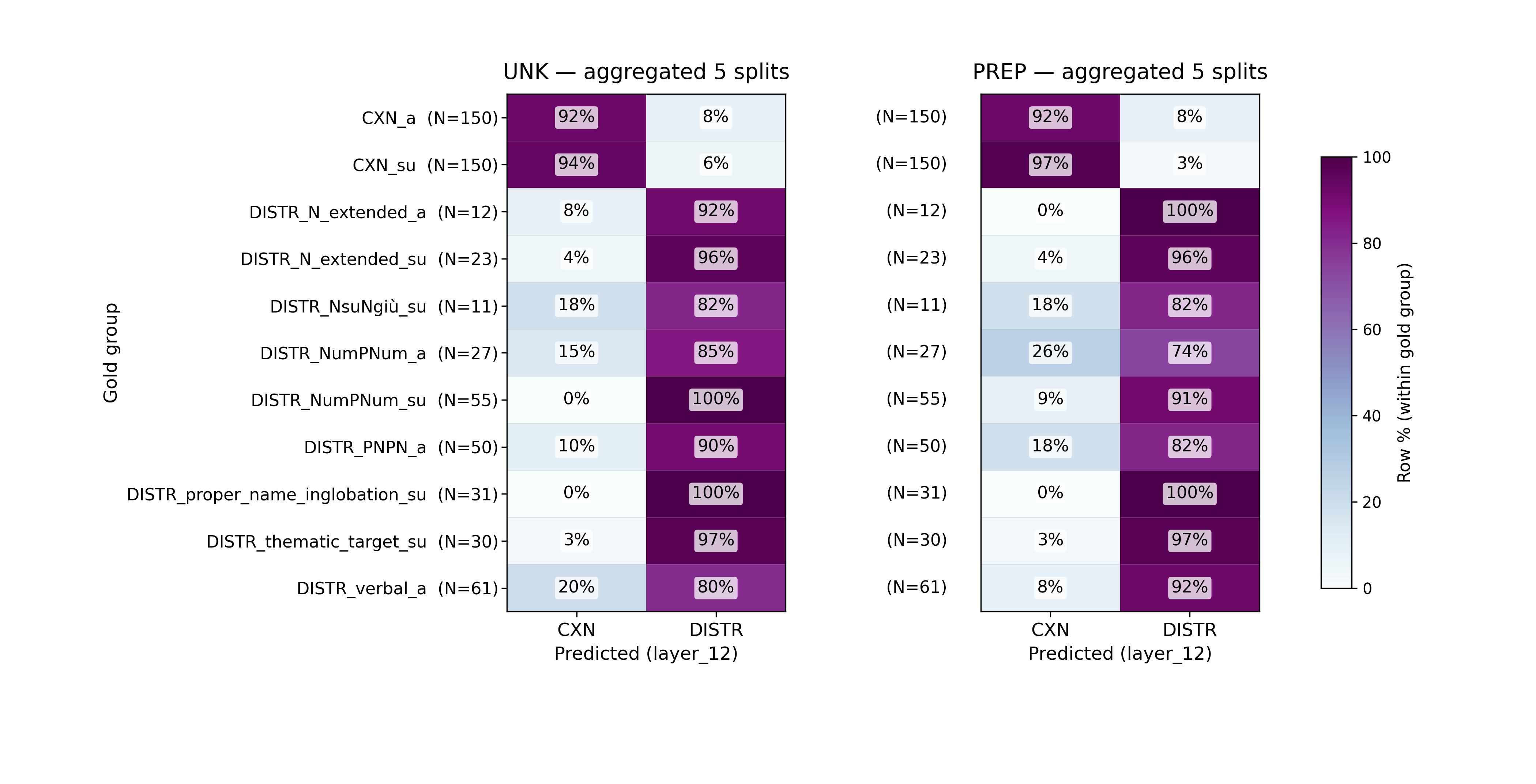}
    \caption{Confusion matrices for the \texttt{SIMPLE} configuration in the Cxn identification experiment. Results for the Italian BERT model (\texttt{bert-base-italian}) are shown for \texttt{[UNK]} embeddings (left) and \texttt{PREP} embeddings (right). The results from the five random splits are aggregated by summing the prediction errors across the five corresponding probing classifiers.}
\label{fig:matrix-ex1simple}
\end{figure}

\begin{figure}[h]
    \centering
    \includegraphics[width=\linewidth]{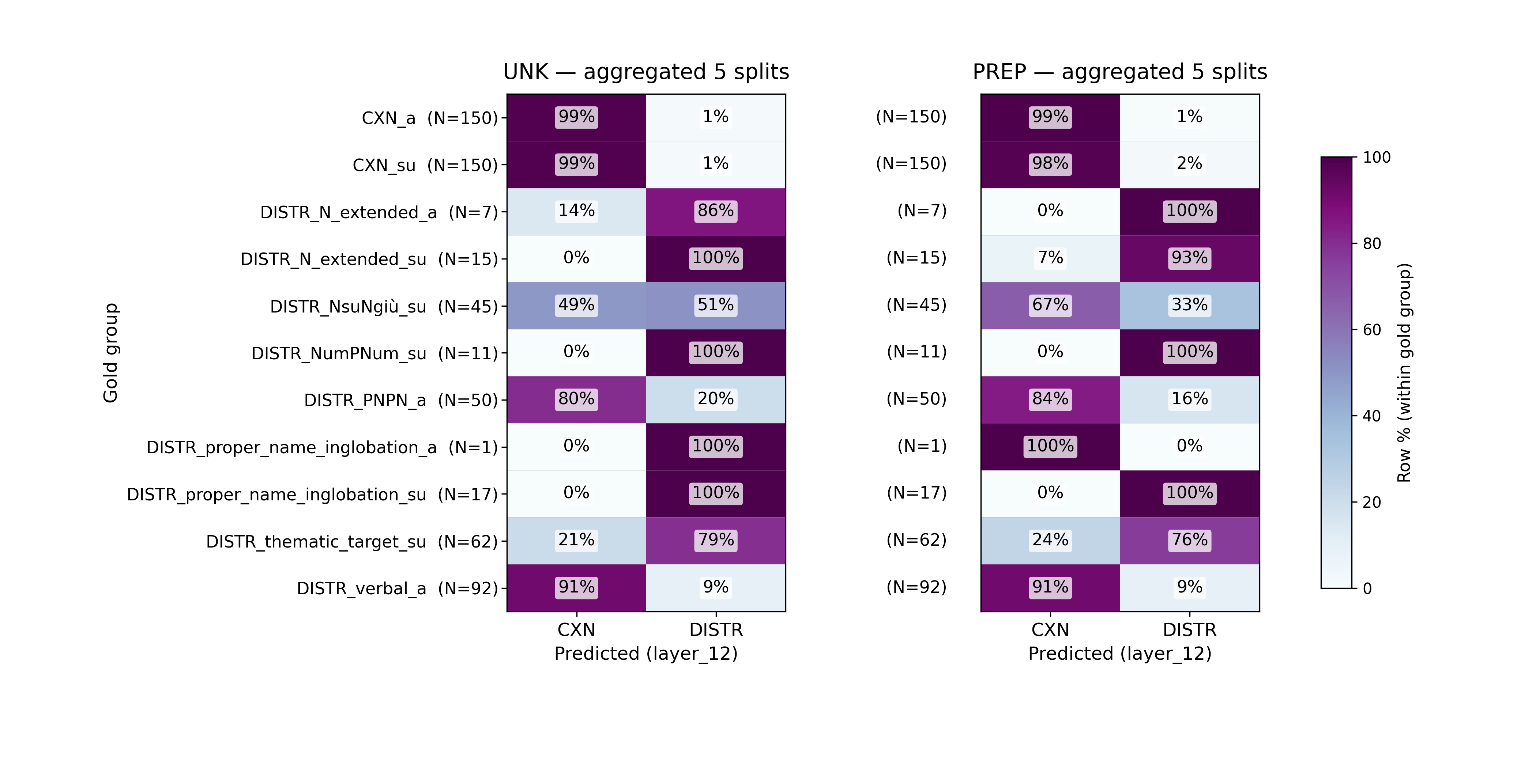}
    \caption{Confusion matrices for the \texttt{PSEUDO} configuration in the Cxn identification experiment. Results for the Italian BERT model (\texttt{bert-base-italian}) are shown for \texttt{[UNK]} embeddings (left) and \texttt{PREP} embeddings (right). The results from the five random splits are aggregated by summing the prediction errors across the five corresponding probing classifiers.}
\end{figure}

\begin{figure}[h]
    \centering
    \includegraphics[width=\linewidth]{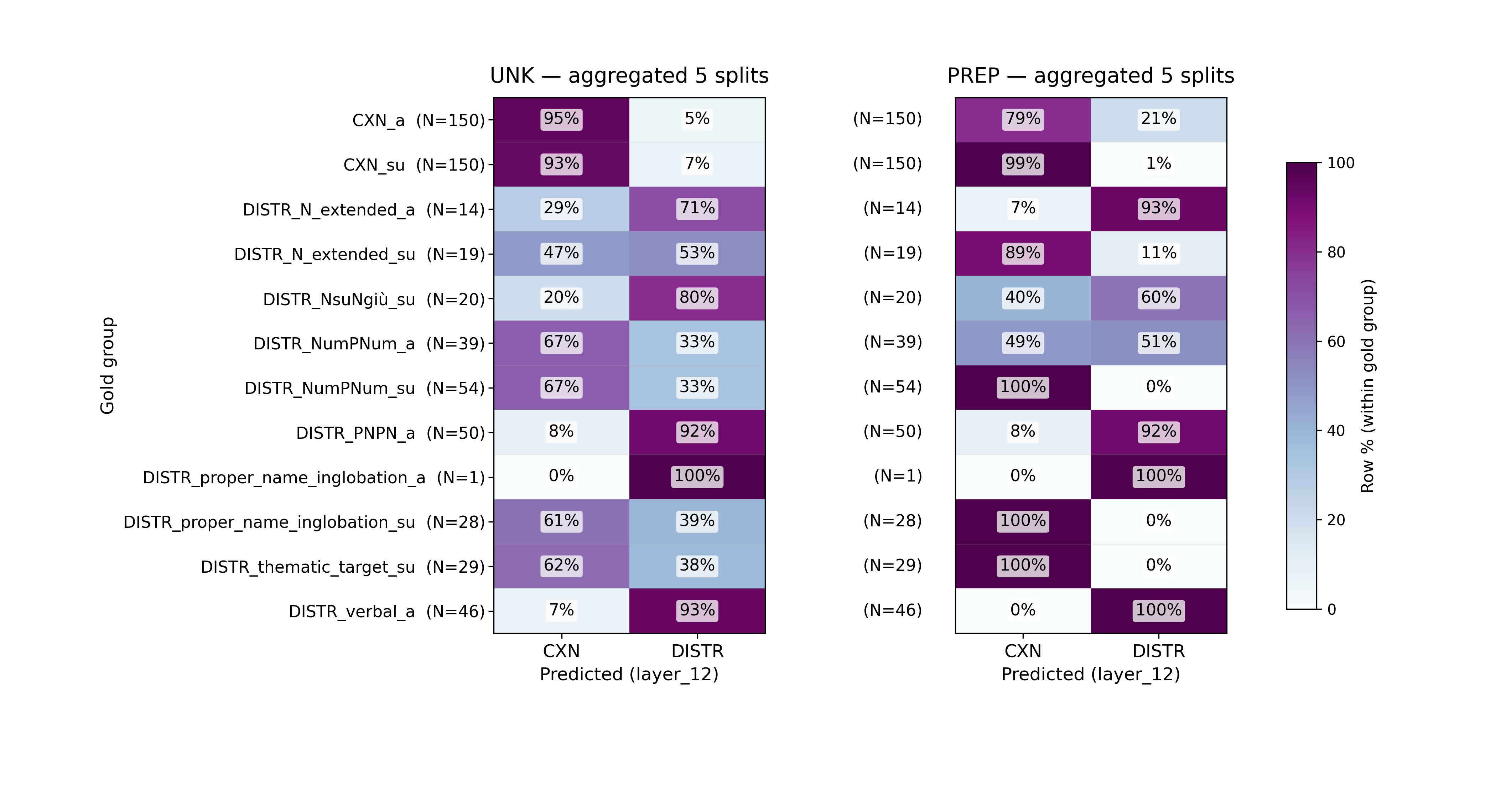}
    \caption{Confusion matrices for the \texttt{OTHER} configuration in the Cxn identification experiment. Results for the Italian BERT model (\texttt{bert-base-italian}) are shown for \texttt{[UNK]} embeddings (left) and \texttt{PREP} embeddings (right). The results from the five random splits are aggregated by summing the prediction errors across the five corresponding probing classifiers.}
\end{figure}

\begin{figure}[h]
    \centering
    \includegraphics[width=\linewidth]{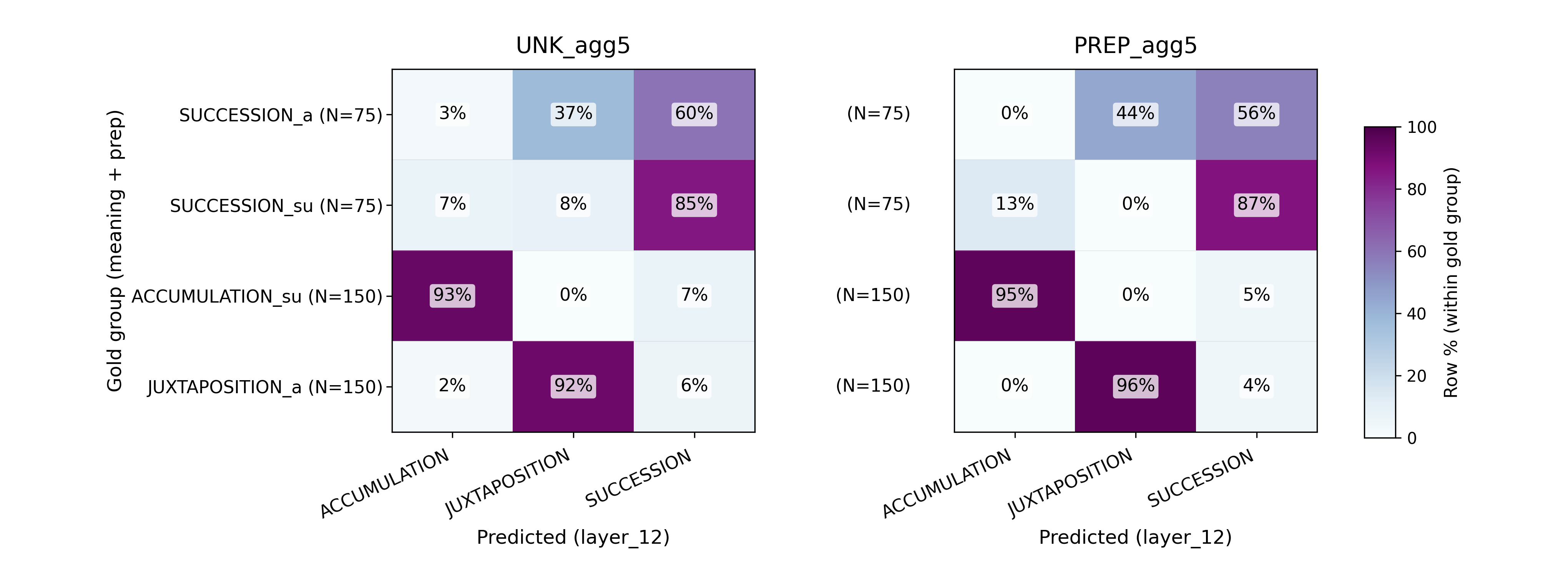}
    \caption{Confusion matrices for the Cxn disambiguation experiment. Results for the Italian BERT model (\texttt{bert-base-italian}) are shown for \texttt{[UNK]} embeddings (left) and \texttt{PREP} embeddings (right). The results from the five random splits are aggregated by summing the prediction errors across the five corresponding probing classifiers.}
    \label{fig:matrix-ex2}
\end{figure}



\end{document}